\documentclass[journal]{IEEEtran}  

\IEEEoverridecommandlockouts                              

\usepackage{graphics} 
\usepackage{epsfig} 
\usepackage{mathptmx} 
\usepackage{times} 
\usepackage{amsmath} 
\usepackage{amssymb}  
\usepackage{xcolor}
\usepackage{textcomp}
\usepackage{lscape}
\usepackage[ruled,vlined]{algorithm2e}

\usepackage{subcaption}
\usepackage{lipsum}
\usepackage{cite}
\usepackage{dblfloatfix}

\makeatletter
\def\BState{\State\hskip-\ALG@thistlm}
\makeatother

\title{\LARGE \bf
Cycloidal Trajectory Realization on Staircase based on  Neural Network Temporal Quantized Lagrange Dynamics (NNTQLD) with Ant Colony Optimization for a 9-Link Bipedal Robot
}

\author{Gaurav Bhardwaj$^{{1},{*}}$, Utkarsh A. Mishra$^{2}$, N. Sukavanam$^{3}$ and  R. Balasubramanian$^{1}$
\thanks{$^{1}$Gaurav Bhardwaj and R.Balasubramanian are with the Computer Science and Engineering Department, IIT Roorkee
        {\tt\small gbhardwaj@cs.iitr.ac.in}, {\tt\small balarfcs@iitr.ac.in}}%
\thanks{$^{2}$Utkarsh A. Mishra is with the Mechanical and Industrial Engineering Department, IIT Roorkee
        {\tt\small umishra@me.iitr.ac.in}}%
\thanks{$^{3}$N. Sukavanam is with the Mathematics Department, IIT Roorkee
        {\tt\small nsukvfma@iitr.ac.in}}%
\thanks{$^{*}$ Corresponding Author}%
\thanks{$^{\dag}$ Funded by Council of Scientific and Industrial Research (CSIR), New Delhi under Grant No- 09/143(0903)/2017-EMR-I .}%
}

\begin{document}

\maketitle
\thispagestyle{empty}
\pagestyle{empty}


\begin{abstract}
In this paper, a novel optimal technique for joint angles trajectory tracking control with energy optimization for a biped robot with toe foot is proposed. For the task of climbing stairs by a 9-link biped model, a cycloid trajectory for swing phase is proposed in such a way that the cycloid variables depend on the staircase dimensions. Zero Moment Point(ZMP) criteria is taken for satisfying stability constraint. This paper mainly can be divided into 3 steps: 1) Planning stable cycloid trajectory for initial step and subsequent step for climbing upstairs and Inverse Kinematics using an unsupervised artificial neural network with knot shifting procedure for jerk minimization. 2) Modeling Dynamics for Toe foot biped model using Lagrange Dynamics along with contact modeling using spring-damper system followed by developing Neural Network Temporal Quantized Lagrange Dynamics which takes inverse kinematics output from neural network as its inputs. 3) Using Ant Colony Optimization to tune PD (Proportional Derivative) controller parameters and torso angle with the objective to minimize joint space trajectory errors and total energy consumed. Three cases with variable staircase dimensions have been taken and a brief comparison is done to verify the effectiveness of our proposed work  Generated patterns have been simulated in MATLAB \textregistered.

\end{abstract}

\section{INTRODUCTION}

The functional planning of legged robots is often derived by animals evolved to excel at the required tasks. However, while imitating such naturally occurring features seen in nature can be very powerful, robots may need to perform several complicated motor tasks that their living counterparts do not. In this context, the most advanced robots which aim towards achieving human-like motion are bipedal robots and humanoids. Different motion patterns of humanoid and biped robots have recently been an active field of research. Walking, running, jogging, ascending and descending stairs and slopes are among the most important ones and can consequentially demonstrate the adaptability of such robots to the surrounding environment. Also, despite wheeled robots, biped robots have a better ability to move on uneven grounds or other complex environments. However, there are some challenges with such systems along with the advantages accompanying it. The dynamics of these robots are complex and inherently unstable. Furthermore, numerous gait patterns have been considered for such types of robots, however, due to the high number of Degrees of Freedom (DOF) of such robots, these methods are not generalizable and each one focuses on realizing a specific gait.

\section{RELATED WORK}
Many researchers proposed the trajectory generation and control algorithm for the biped robots on flat surfaces \cite{c1,c2,c3}. Moreover, studies about the ascending or descending stairs by these robots have been a subject of active research \cite{c4,c5,c6}. Kajita et al. \cite{c2} suggested preview control of Zero Moment Point with a spiral staircase. Park et al. have presented a trajectory generation method and control approach for a biped robot to climb stairs \cite{c5}. They developed an off-line path planning method using the Virtual Height Inverted Pendulum Model (VHIPM) method. Shih and Chio \cite{c7} studied stairs walking for biped robot taking static models. Jeon et al. \cite{c8} proposed a genetic algorithm-based optimal trajectory generation method to walk upstairs. Morisawa et al. \cite{c9} pattern generation technique for biped walking constraint on a parametric surface. Sato et al. \cite{c10} proposed a virtual slope method for biped robot climbing on stairs. Gutmann \cite{c11} proposed a stereo vision technique for a humanoid robot to climb stairs. Significant methods presented in the literature are designed based on a predefined path tracking. However, when the environment changes, it is necessary to redesign the trajectory. Nevertheless, humans do not walk based on a predefined trajectory. Humans predict their next steps based on several environmental and geometric constraints. 

In order to achieve effective knowledge and formulate proper tracking measures, robot dynamics and kinematics play a crucial role. For specifically, in such tracking problems, Inverse Kinematics has been an active field of concern. Artificial neural networks(ANN) played a significant role in robots and manipulators in the past decades. When talking about inverse kinematics, artificial neural networks give more satisfying results compared to geometric, iterative, or analytic methods, the reason being the existence of multiple solutions in case of inverse kinematics. \cite{c12,c13,c14,c15,c16,c17} used a supervised neural network for inverse kinematics solution obtaining training data from the relationship between joint coordinates and end effector Cartesian coordinates. Panwar \cite{c18} applied the ANN method for bipedal stable walking without obtaining training data in an unsupervised manner. For biped robot dynamics, either we can approximate our model by taking simplified cases like Inverted Pendulum Model \cite{c19,c20,c21,c22} for the purpose of trajectory generation. On the other hand, we can take distributed mass to develop dynamic equations as well.

Also, because of the unilateral contact between a humanoid robot foot and the terrain surface, the walking patterns should fulfill certain feasibility constraints as well. These are fundamental constraints that ensure stability and hence make them walk on flat or rough terrains without falling down. Vukobratovic \cite{c23} in 1972 proposed a measuring index for stability called Zero Moment Point. Foot Rotation Indicator(FRI) \cite{c24} is another technique that determines the tendency of feet to rotate. Hirukawa \cite{c25} proposed the Contact Wrench sum and Contact Wrench Cone approached to work as an indicator of stability in biped robots. ZMP is one of the simplest and most used techniques.

\section{CONTRIBUTION OF PRESENTED WORK}

In this paper, a novel cycloidal realization is performed to accomplish stair climbing by a toe-foot biped robot model. Furthermore, to complete the trajectory tracking objectives, ANN-based inverse kinematics and temporally quantized dynamics algorithm was formulated to structure the control profiles with ant colony optimization to tune controller gains and torso pitch angle and hence providing evidence to shed some light on the energy consumption and power requirements as well. The paper incorporates the following contributions:

The cycloidal trajectory proposed in the work is dynamic in nature and hence can easily adapt to the changing stair geometry and for various robot dimensions. In the case of staircases, the optimal rise/run ratio is about $11/7 = 1.57$ \cite{c26}, which is one of the reasons for taking cycloid as reference trajectory for the swing phase because for half of cycloid, horizontal to vertical distance ratio is $\pi/2 = 1.57$ which is an unnoticed fact. This is in complete congruence with the fact dynamic planning strategy humans adapt. The desired trajectory is composed of multiple segments such that the robot plans, catches, and tracks a cycloidal trajectory. Parameters of the cycloid are based on the configuration of the staircase and the algorithm for calculation of ankle trajectory is developed which is applicable for varying staircase dimensions which make it computationally faster compared to polynomial based techniques. A circular arc trajectory is considered for the hip.

A novel online trajectory tracking procedure is developed where an ANN-based real-time unsupervised inverse kinematics is performed and the processed output is served as the desired joint space variables except torso pitch angle for a trajectory tracking algorithm using Ant colony Optimization. The tracking algorithm is based on a novel temporally quantized classical Euler-Lagrangian dynamics with a distributed mass model approach which enables the trajectory tracker to choose controller gains and torso pitch angle in order to satisfy extra constraints while following the desired trajectory to the maximum extent with minimum energy consumption. The approach starts with quantizing time by taking a suitable time step throughout the pre-calculated ankle trajectory for climbing staircase step, then for every two consecutive ankle position $i$ and $(i+1)$, where $i:0,1,..n:n$ denotes total number of time instants, UIKNN provides joint angle variables which are then fed into trajectory tracker to get controller gain values and torso pitch angle with objective to track joint space trajectories with minimum instantaneous power resulting in lesser energy consumption. 

This paper is organized as follows: Section \ref{rmd} describes the considered toe foot, robot model. Section \ref{tp} is dedicated to staircase trajectory planning for robot followed by ANN-based inverse kinematics solution in section \ref{ik}. Section \ref{cm} elaborated our contact model and then Dynamic Modeling for the robot in Section \ref{dm} which elaborated both Lagrange Dynamics and TQLD approach. Section \ref{zmpf} describes the ZMP stability criteria. Section \ref{cd} is dedicated to controller design with ant colony optimization. Finally, in section \ref{rad}, the obtained results are analyzed and the paper concludes the whole procedure in section \ref{cncl}.

\section{ROBOT MODEL DESCRIPTION} \label{rmd}

\begin{figure}[thpb]
      \centering
      \includegraphics[width=\linewidth]{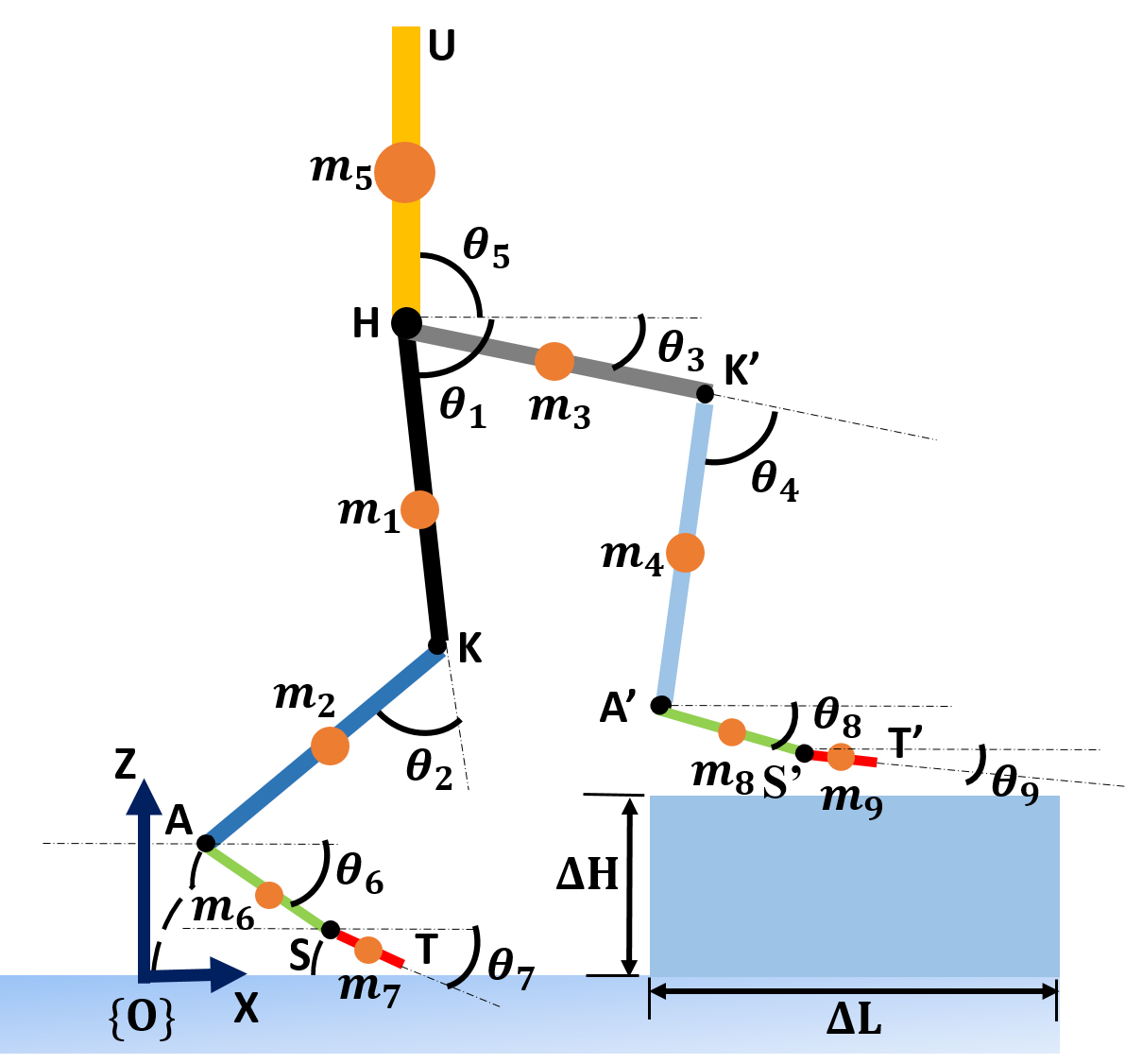}
      \caption{Toe-Foot Robot Model Description and Notations}
      \label{modelpic}
   \end{figure}

The model considered for the work is a planar biped with a toe-foot joint. The motion is completely constrained to the sagittal plane. Each leg consists of 4 links and 4 joints, all of which are revolute in nature as shown in Fig. \ref{modelpic}. These joints are referred as the Hip(H), Knee(K), Ankle(A), Sole(S) along with the Toe tip(T). The attributes of the model can be visualized from Fig. \ref{modelpic} and is shown in Table \ref{tablemodel}.

\begin{table}[h]
\caption{Joint Positions and Attributes of Each Link}
\label{tablemodel}
\begin{center}
\begin{tabular}{|c|c|c|c|c|c|}
\hline
\multicolumn{2}{|c}{Joint Name} & \multicolumn{4}{|c|}{Positions}\\
\hline
\multicolumn{2}{|c}{Hip(H)} & \multicolumn{4}{|c|}{H (both Swing \& Stance Leg)}\\
\hline
\multicolumn{2}{|c}{Knee(K)} & \multicolumn{4}{|c|}{K (Swing Leg), K$'$ (Stance Leg)}\\
\hline
\multicolumn{2}{|c}{Ankle(A)} & \multicolumn{4}{|c|}{A (Swing Leg), A$'$ (Stance Leg)}\\
\hline
\multicolumn{2}{|c}{Sole(S)} & \multicolumn{4}{|c|}{S (Swing Leg), S$'$ (Stance Leg)}\\
\hline
\multicolumn{2}{|c}{Toe(T)} & \multicolumn{4}{|c|}{T (Swing Leg), T$'$ (Stance Leg)}\\
\hline
\multicolumn{6}{|c|}{ } \\
\hline
Link No. & Link & Length & Value(cm) & Mass & Value(Kg)\\
\hline
1 & HK & $l_1$ & 40 & $m_1$ & 6\\
\hline
2 & KA & $l_2$ & 40 & $m_2$ & 4\\
\hline
3 & HK$'$ & $l_3$ & 40 & $m_3$ & 6\\
\hline
4 & K$'$A$'$ & $l_4$ & 40 & $m_4$ & 4\\
\hline
5 & UH & $l_5$ & 30  & $m_5$ & 30\\
\hline
6 & AS & $l_6$ & 12  & $m_6$ & 0.70\\
\hline
7 & ST & $l_7$ & 5 & $m_7$ & 0.15\\
\hline
8 & A$'$S$'$ & $l_8$ & 12 & $m_8$ & 0.70\\
\hline
9 & S$'$T$'$ & $l_9$ & 5 & $m_9$ & 0.15\\
\hline
\end{tabular}
\end{center}
\end{table}
 
 Thus, the total length of the leg is $(l_1+l_2)$ and that of foot is $(l_5+l_6)$. For a gait, the model hip is treated as the base and the ankle of the active leg i considered as the end-effector and this corresponds to a 2-link manipulator. The movement is divided into 2 phases, namely the Double Support Phase (DSP) and the Single Support Phase(SSP). The former one corresponds to the initial condition when both the feet are making contact with the ground whereas the latter represents the swing of the active leg with passive leg's foot remaining in contact with the ground.
 
The presented work explores the possibilities of stable stair climbing for the model shown with known variable step length and height. The complete task objective is as follows:
\begin{itemize}
\item[1.] Consider a standing position, the model starts to climb the first step of the staircase, attains a velocity.
\item[2.] Second task objective is to enable the model to climb alternate steps in such a way to minimize the actuator effort.
\end{itemize}

\section{TRAJECTORY PLANNING} \label{tp}

\subsection{For Swing Leg}

The biped is considered to be planar as visualized from the sagittal plane only. The torso movement is not considered. A complete motion is achieved during the time interval $(t_0 = 0,t_f )$, which is divided into majorly four phases.
\begin{itemize}
\item $t = 0 \text{ to } t_2$: Double Support Phase
\item Single Support Phase - Pick the Cycloid
\item $t = t_2 \text{ to } t_{bc}$: Single Support Phase - Catch the Cycloid
\item $t = t_{bc} \text{ to } t_f$: Single Support Phase - Track the Cycloid
\end{itemize}

Each phase is described as follows:

\subsubsection{Double Support Phase}

The feet of the swing leg denoted by points A, S and T are on the ground with coordinates $(0,0)$, $(l_6 , 0)$ and $(l_6 + l_7 , 0)$ respectively. The feet is considered to move as a two-link manipulator with AS and ST as the two arms about the fixed toe, T.

Thus, during this phase, the ankle coordinates are given by the following equations.
\begin{equation}
x_A (t) = l_6 (1-cos\theta_{6}) + l_7 (1-cos\theta_{7})
\end{equation}
\begin{equation}
z_A (t) = l_6 sin\theta_{6} + l_7 sin \theta_{7}
\end{equation}

For the duration, the sole AS rotates about S, from angle $0$ at $t = 0$ to angle $\theta_a$ at time $t = t_1$ with toe being stable on ground. For the time duration, $(t_1 , t_2 )$ the AS-ST links behave as 2 link manipulator about T where AS reverses its motion and ST rotates from angle $0$ at $t = t_1$ to angle $\theta_b$ at time $t = t_2$ . The parameters $\theta_a$ and $\theta_b$ are modeling parameters for stable gait generation. Thus, for polynomial parametrization of the motion,
\begin{equation}
   \theta_{6} (t)= 
\begin{cases}
    \theta_a (3 \frac{t^2}{t^2_1} - 2 \frac{t^3}{t^3_1}),& \text{if } 0 \leq t \leq t_1\\
    \theta_a (-4 +12 \frac{t}{\Delta t_{DSP}} - 9 \frac{t^2}{\Delta t_{DSP}^2} +2 \frac{t^3}{\Delta t_{DSP}^3}), & \text{if } t_1 \leq t \leq t_2
\end{cases}
\end{equation}
\begin{equation}
   \theta_{7} (t)= 
\begin{cases}
    0 ,& \text{if } 0 \leq t \leq t_1\\
    \theta_b (-5 +12 \frac{t}{\Delta t_{DSP}} - 9 \frac{t^2}{\Delta t_{DSP}^2} +2 \frac{t^3}{\Delta t_{DSP}^3}),              & \text{if } t_1 \leq t \leq t_2
\end{cases}
\end{equation}

where, $\Delta t_{DSP} = t_2-t_1$.

\begin{figure}[thpb]
\centering
\includegraphics[width=0.8\linewidth]{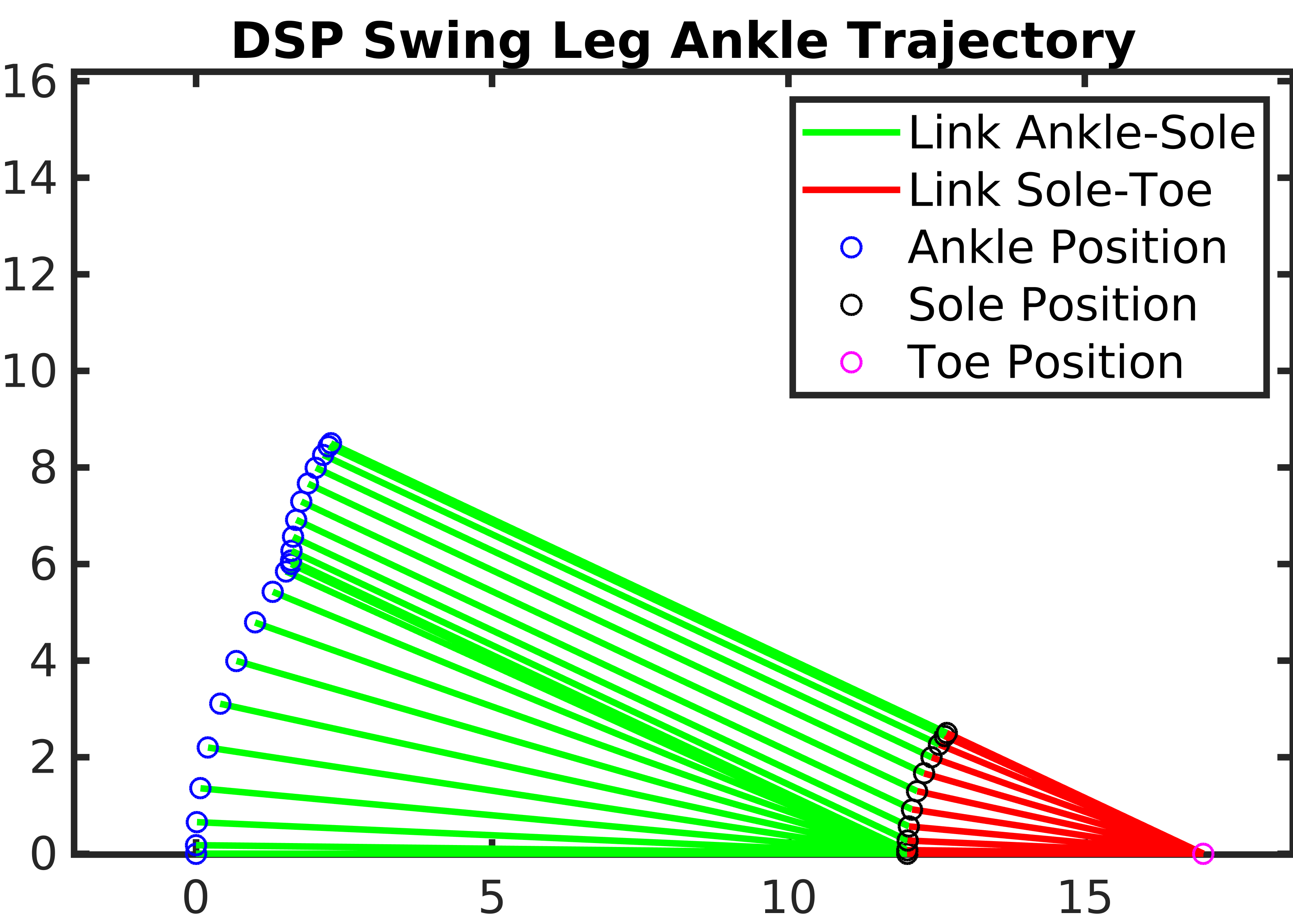}
\caption{Swing leg's Foot Trajectory during DSP}
\label{Result:DSP}
\end{figure}

\subsubsection{Single Support Phase - Pick the Cycloid}

Cycloidal trajectory planning enables null accelerations at the beginning and the end of the gait and hence, the dynamic forces on the links are not significant for those instants. Furthermore, the smooth change in accelerations results to suitable behavior of the dynamic forces contributing to stable gaits. The trajectory parameters used for the study is based on the robot's geometrical as well as the task objective discussed in the model description section. 

Let us consider a cycloid of constant radius $r$ and a time parametrized $\theta (t)$, on which the Cartesian coordinates of a point is given by,
\begin{equation}
x_{cycloid}(t) = r (\theta_c(t) - \sin{\theta_c(t)})
\end{equation}
\begin{equation}
z_{cycloid}(t) = r (1 - \cos{\theta_c(t)})
\end{equation}
The maximum height that can be achieved in such a motion is at $\theta = \pi$ which is at a horizontal distance of $\pi r$. Now, to climb any step of height ($\Delta z$) and at certain distance ($\Delta x$), $r$ is considered to be equal to $\Delta z/2$ and cycloid is formed with the height of the ankle $z_A (t_2)$, at the end of DSP, as its origin height and a variable $x_{c0}$ coordinate depending upon the step parameters. The starting of the cycloid is governed by the value of $\theta_c$ at the beginning. After the DSP ends, the remaining distance to be covered by the trajectory happens to be $\Delta x_c = (\Delta x - x_A(t_2))$. The value of $\theta_c$ in the beginning, $\theta_{c0}$, is chosen such that it satisfies,

\begin{equation}
r (\theta_{c0} - \sin{\theta_{c0}}) = \pi r - \Delta x_c,
\end{equation}

if $\pi r \geq \Delta x_c$ and $0$ otherwise. (7) shows the dependence of the value of $x_{c0}$ on $r$ and $\Delta x_c$. As the DSP is bounded by the workspace and feasibility constraints of a 2-link mechanism, it becomes difficult for it to follow the above described cycloidal trajectory just after it ends and hence, there should be a bridge to catch the cycloid after the DSP ends.

\begin{figure}[thpb]
\centering
\includegraphics[width=0.8\linewidth]{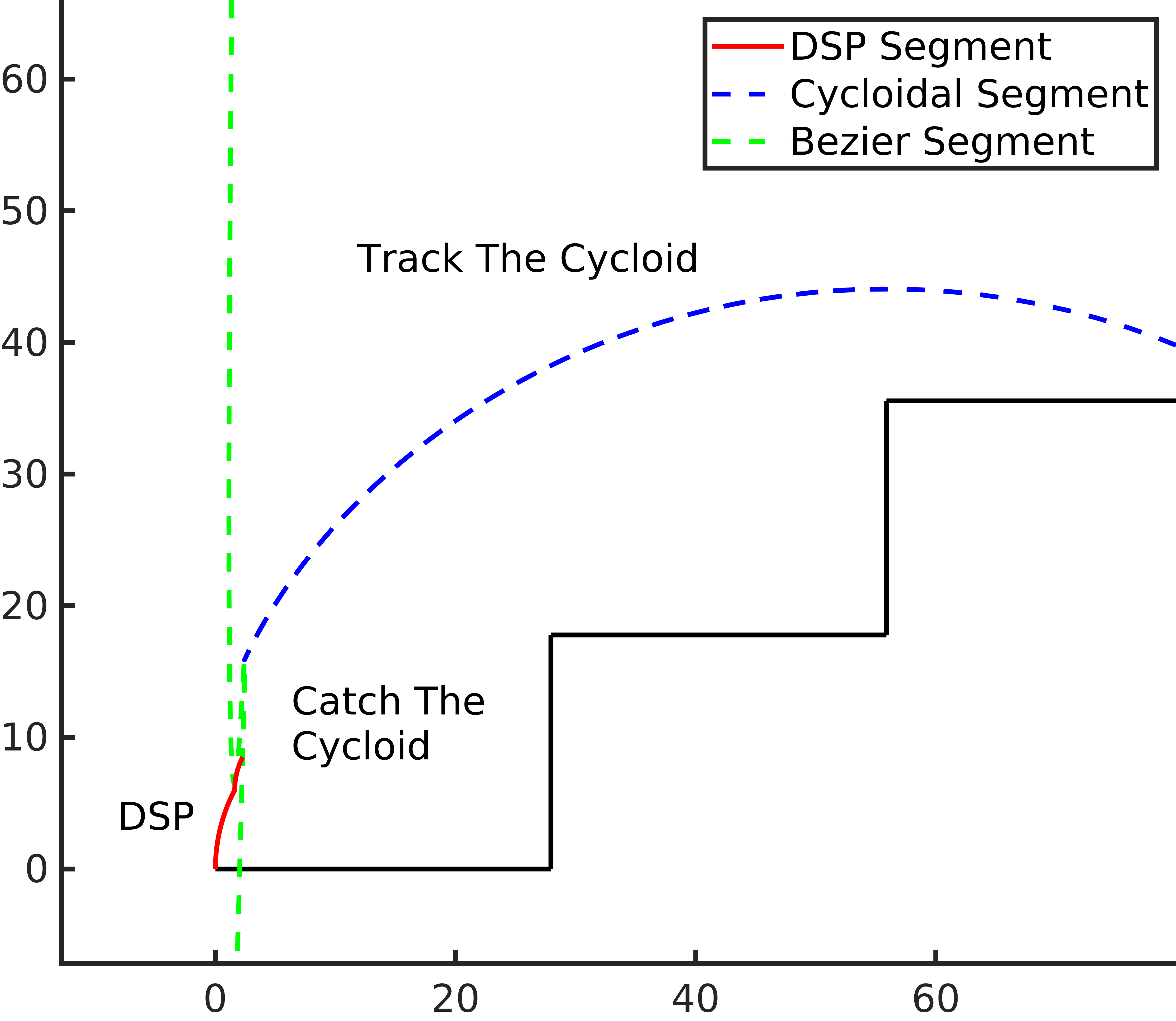}
\caption{Various Segments of Swing Leg's Ankle Trajectory Planning for Subsequent Upstairs Climbing}
\label{Result:iupsegment}
\end{figure}

\subsubsection{Single Support Phase - Catch the Cycloid}
This bridge is modeled as a bezier curve. The trajectory was formulated based on 4 control points, namely $(x^{b}_{c_i},z^{b}_{c_i})$ for $i=1$ to $4$, which starts from the end of DSP i.e. $(x^{b}_{c1},z^{b}_{c1}) = (x_A (t_2),z_A (t_2))$ being the first control point. The second control point was chosen based on the cycloid formulation, with the value of $\theta_{c0}$.
\begin{equation}
   x^{b}_{c2}= 
\begin{cases}
    x_A (t_2),& \text{if } \pi r \geq \Delta x_c\\
    (\Delta x - \pi r),              & \text{otherwise}
\end{cases}
\end{equation}
\begin{equation}
   z^{b}_{c2}= 
\begin{cases}
    z_A (t_2)+r (1-\cos{\theta_{c0}}),& \text{if } \pi r \geq \Delta x_c\\
    z_A (t_2),              & \text{otherwise}
\end{cases}
\end{equation}
The remaining control points were selected from within the cycloid such that the bezier curve blends completely with the cycloidal trajectory. Considering that the bezier curve catches the cycloid in time $t_{bc}$, the other two control points were chosen to be at an angular displacement of $\theta^{b}_{c3} = \theta_{c0}+\Delta \theta_{c0}$ and $\theta^{b}_{c4} = \theta_{c0}+2\Delta \theta_{c0}$ respectively, such that the coordinates $(x^{b}_{c3},z^{b}_{c3})$ and $(x^{b}_{c4},z^{b}_{c4})$ are obtained from the cycloid equation given in (5) and (6). Finally, the bezier curve is determined to be,
\begin{equation}
x^{b}(t) = (1-s)^3 x^{b}_{c1} + (1-s)^2 s x^{b}_{c2} + (1-s) s^2 x^{b}_{c3} + s^3 x^{b}_{c4}
\end{equation}
\begin{equation}
z^{b}(t) = (1-s)^3 z^{b}_{c1} + (1-s)^2 s z^{b}_{c2} + (1-s) s^2 z^{b}_{c3} + s^3 z^{b}_{c4}
\end{equation}
where $s = \frac{t - t_2}{t_{bc}-t_2}$ is the normalized time in the duration for which the ankle traverses the bezier curve trajectory. Furthermore, to maintain the continuity the velocity at the blend of the bezier and cycloid curves should be same, hence 
\begin{equation}
\dot{\theta}_{c}(t=t_{bc}) = \frac{\dot{z}^{b}(t=t_{bc})}{r \sin{\theta^{b}_{c4}}}
\end{equation}
Also, to make the complete utilization of cycloidal trajectory, $\ddot{z}_c(\theta_c = \pi) = g$, where $g$ is the acceleration due to gravity. This deduces to $\dot{\theta}_c(t=t_3) = \sqrt{g/r}$. Finally, the complete cycloid definition can be completed by defining $\theta_c(t)$ as,
\begin{equation}
\theta_c(t) = a_{c0} + a_{c1} t + a_{c2} t^2 + a_{c3} t^3,
\end{equation}
where the coefficients $a_{ci}'s$ are given by,
\begin{equation}
\begin{bmatrix}
\theta^{b}_{c4} \\
\dot{\theta}_{c}(t=t_{bc}) \\
\pi \\
\dot{\theta}_c(t=t_3)
\end{bmatrix} = 
\begin{bmatrix}
1 & t_{bc} & t_{bc}^2 & t_{bc}^3 \\
0 & 1 & 2 t_{bc} & 3 t_{bc}^2 \\
1 & t_{3} & t_{3}^2 & t_{3}^3 \\
0 & 1 & 2 t_{3} & 3 t_{3}^2
\end{bmatrix}
\begin{bmatrix}
a_{c0}\\
a_{c1}\\
a_{c2}\\
a_{c3}
\end{bmatrix}
\end{equation}

\subsubsection{Movement of Sole and Toe}

The trajectory of the sole (S) and toe (T) during the complete motion $(0 \leq t \leq t_3)$ were chosen based on the geometry of the robot and the stair-step with which it is interacting. It was assured that no collision with the surface occurs at any point. The $x$ and $z$ coordinates of S and T are given by the following equations.

\begin{equation}
   x_T (t)= 
\begin{cases}
    l_6 + l_7,& \text{if } 0 \leq t \leq t_2\\
    x_A(t) + (l_6+l_7) \cos{\theta_6(t)}, & \text{if } t_2 \leq t \leq t_3
\end{cases} 
\end{equation}
\begin{equation}
  z_T (t)= 
\begin{cases}
    0 ,& \text{if } 0 \leq t \leq t_2\\
    z_A(t) + (l_6+l_7) \sin{\theta_6(t)}, & \text{if } t_2 \leq t \leq t_3
\end{cases}
\end{equation}
\begin{equation}
   x_S (t)= 
\begin{cases}
    l_6 ,& \text{if } 0 \leq t \leq t_1\\
    l_6 + l_7 - l_7 \cos{\theta_{7}(t)} ,& \text{if } t_1 \leq t \leq t_2\\
    x_A(t) + l_3 \cos{\theta_6(t)}, & \text{if } t_2 \leq t \leq t_3
\end{cases}
\end{equation}
\begin{equation}
   z_S (t)= 
\begin{cases}
    0 ,& \text{if } 0 \leq t \leq t_1\\
    l_7 \sin{\theta_{7}(t)} ,& \text{if } t_1 \leq t \leq t_2\\
    z_A(t) + l_6 \sin{\theta_6(t)}, & \text{if } t_2 \leq t \leq t_3
\end{cases}
\end{equation}

Here,
\begin{equation}
\theta_3(t) = a_{t30} + a_{t31} t + a_{t32} t^2 + a_{t33} t^3,
\end{equation}
for $t = (t_2,t_3)$, where the coefficients $a_{t3i}'s$ are given by,
\begin{equation}
\begin{bmatrix}
\theta_{b} \\
0\\
\theta_{b} \\
0
\end{bmatrix} = 
\begin{bmatrix}
1 & t_{2} & t_{2}^2 & t_{2}^3 \\
0 & 1 & 2 t_{2} & 3 t_{2}^2 \\
1 & t_{3} & t_{3}^2 & t_{3}^3 \\
0 & 1 & 2 t_{3} & 3 t_{3}^2
\end{bmatrix}
\begin{bmatrix}
a_{t30}\\
a_{t31}\\
a_{t32}\\
a_{t33}
\end{bmatrix}
\end{equation}

Once, the swing leg lands with it's toe (T) touching the ground, the next DSP phase starts and the ankle follows the exact reverse trajectory of that discussed in the first part of this segment i.e. first as a 2-link manipulator with fixed axis at T and then a single link manipulator with fixed axis at sole (S).

\subsection{For Hip Motion}

The motion of the hip was formulated as a COG trajectory with the ZMP equation of a one-mass COG model in sagittal plane. During the single support phase, the analytical solution of ZMP trajectory is considered as the COG trajectory. Additionally, the ZMP motion is planned as a 3-degree polynomial parametrization with C-2 continuity. For such a formulation, the ZMP trajectory in saggital plane is as follows.
\begin{equation}
p_x(t) = x_C(t) - \frac{z_C}{g} \ddot{x}_C(t)
\end{equation}
\begin{equation}
p_x(t) = a_{z0} + a_{z1} t + a_{z2} t^2 + a_{z3} t^3
\end{equation}
where $(p_x(t),0)$ and $(x_C,z_C)$ represent the ZMP and COG position respectively. The analytical solution for the ZMP equations (21) and (22), and the COG trajectories is obtained as follows,
\begin{equation}
\begin{split}
x_C(t) = C_1 e^{\omega t} + C_2 e^{-\omega t} + a_{z0} + a_{z1} t + a_{z2} t^2 + a_{z3} t^3 \\
+ \frac{z_C}{g}(6 a_{z3} t + 2 a_{z2})
\end{split}
\end{equation}
\begin{equation}
\omega = \sqrt{\frac{g}{z_{Ci}}}
\end{equation}
where, $z_{Ci}$ is the initial centroidal height. Various researches have used COG trajectory for the z-direction in the form of a virtual slope given by $z_C(t) = k x_C(t) + z_{Ci}$ where the value of $k$ is determined  based on the slope of the stairs step need to be climbed \cite{c26}. But, such an approximation is not practical for determining the hip motion. Hence, the hip motion is modeled as a circular arc with the virtual slope trajectory as its chord length. This inverted circular arc traverses between the initial hip position $(x_H^{initial},z_H^{initial})$ to the final hip position $(x_H^{final},z_H^{final})$ in a more practical manner. Let $R_{H}$ and $\theta_{H}$ denote the segment corresponding to the desired arc.
\begin{equation}
\begin{split}
R_H \sin(\theta_H) = x_H^{final} - x_H^{initial} \\
R_H (1-\cos(\theta_H)) = z_H^{final} - z_H^{initial} \\
z_C(t) = z_{Ci} + \sqrt{R_H^2 - (x_C(t) - x_H^{initial})^2 }
\end{split}
\end{equation}
Furthermore, the value of coefficients $a_{zi}'s$, $C_1$ and $C_2$ in equation (23) is determined using the boundary conditions of ZMP and COG respectively for the time-interval $(0, t_3)$. These boundary conditions are given in the following equations.
\begin{equation}
\begin{split}
\begin{bmatrix}
x_{ZMP}^{initial} \\
0\\
x_{ZMP}^{final} \\
0
\end{bmatrix} = 
\begin{bmatrix}
1 & 0 & 0 & 0 \\
0 & 1 & 0 & 0 \\
1 & t_{3} & t_{3}^2 & t_{3}^3 \\
0 & 1 & 2 t_{3} & 3 t_{3}^2
\end{bmatrix}
\begin{bmatrix}
a_{z0}\\
a_{z1}\\
a_{z2}\\
a_{z3}
\end{bmatrix} \\
x_C(t=0) = x_{COG}^{initial}, \; x_C(t=t_3) = x_{COG}^{final}
\end{split}
\end{equation}

\subsection{For Stance leg}

The stance leg follows the hip trajectory while keeping the ankle fixed. Furthermore, the sole (S') and toe (T') of stance leg remain at the same point. As the swing leg starts DSP for the next phase, the DSP for stance leg commences. The planning for this motion is similar, although, the planning next place for the next intermediate step.

\section{INVERSE KINEMATICS} \label{ik}

\subsection{Inverse Kinematics formulation}
For forward kinematics we have used equation from \cite{c13} based on D-H Procedure.The previous trajectory planning segment clearly defines the trajectory of the hip (H), ankle (A), sole (S) and toe (T). Considering hip (H) as the base and ankle (A) as the end-effector, the forward kinematics equation is obtained as a function of the joint angles $\theta_1(t)$ and $\theta_2(t)$.
\begin{equation}
x_A(t) = x_H(t) + (l_1 \cos{\theta_1(t)} + l_2 \cos(\theta_1(t)+\theta_2(t)))
\end{equation}
\begin{equation}
z_A(t) = z_H(t) - (l_1 \sin{\theta_1(t)} + l_2 \sin(\theta_1(t)+\theta_2(t)))
\end{equation}
The equations (26) and (27) apply for both the swing leg and the stance leg with joint angles $\theta_1(t)$,$\theta_2(t)$ and $\theta_5(t)$,$\theta_6(t)$ respectively.

\subsection{Knot shifting procedure based Unsupervised Inverse Kinematics Neural Network (UIKNN) approach}
\subsubsection{UIKNN}
An ANN based approach was used to solve the inverse kinematics problem from the forward kinematic equations (27) and (28). A feed forward network was modeled and trained in real time for every time instant. The hyperparameters of the network are given in Table \ref{tableiknn}. 
\begin{table}[h]
\caption{Hyperparameters for Feed Forward Network}
\label{tableiknn}
\begin{center}
\begin{tabular}{|c|c|}
\hline
Parameter & Value\\
\hline
Input Neurons & 2\\
\hline
Output Neurons & 2\\
\hline
Hidden layer & 1\\
\hline
Hidden layer nodes & 10\\
\hline
Activation function & Sigmoid ${1}/{(1+e^{-x})}$\\
\hline
Learning Rate & $10^{-4}$\\
\hline
Maximum Iterations & 5000\\
\hline
\end{tabular}
\end{center}
\end{table}
For a desired ankle position $(x_A^{input},z_A^{input})$, the joint angles $\theta_1^{network}$ and $\theta_2^{network}$ were approximated. These approximated joint angles were substituted in the (27) and (28) to obtain the network estimated ankle position $(x_A^{network},z_A^{network})$. Finally, the loss (error) function, $E_{network}$, was calculated as the squared error between the desired and approximated ankle positions.
\begin{equation}
E_{network} = (x_A^{input}-x_A^{network})^2 + (z_A^{input}-z_A^{network})^2
\end{equation}
The loss function was optimized using a gradient descent based on the partial differentiation of the error with respect to the layer weights. Finally, the weights$(W)$ were updated according to the calculated gradients $(\delta)$, for each time step i.e.
\begin{equation}
W_{n+1}(t) = W_n(t) - \alpha \delta ,
\end{equation}
where $n$ represents the iteration (epoch) number. The weight updates were stop once the error goes down below the threshold equal to $10^{-6}$.

\begin{algorithm}[thpb]
\SetAlgoLined
\KwResult{Unsupervised Inverse Kinematics Neural Network}
$\textit{Determine } \text{cooridnates of}\textit{ Hip(H), Ankle(A)}$\;
$W_{ij}  \gets \textit{Hidden Layer Weights}$\;
$W_{jk}  \gets \textit{Output Layer Weights}$\;
$W_{ij},\;\;W_{jk}  \gets \textit{Random Initializing}$\;
$Act() \gets \textit{Sigmoid Activation function}$\;
$\alpha \gets \textbf{learning rate}$\;
 \While{$t \leq t_{final}$}{
 $input \gets (A)[x, y]$ \;
 $output[\theta_1, \theta_2] \gets W_{jk}(Act(W_{ij}*input))$\;
 $(A)_{modified} \gets function((H),\theta_1, \theta_2) \textbf{ from (27) (28)}$\;
 $\text{Error} (E) \gets (A)_{modified} - (A) \textbf{ from (29)}$\;
 
 \eIf{$(E) \leq 10^{-6}$}{
 Stop\;
 }{
 $\delta_{ij},\delta_{jk}, \gets {\partial E}/{W_{ij}}, {\partial E}/{W_{jk}}$\;
 $W_{ij} \gets W_{ij} - \alpha \delta_{ij}$\;
 $W_{jk} \gets W_{jk} - \alpha \delta_{jk}$\;
 }
 }
 \caption{UIKNN Algorithm}
 \label{iknn}
\end{algorithm}

   \subsubsection{Knot Shifting Procedure}
 In case of bipedal robot or for a general manipulator also, joint space trajectory demands for jerk free 
trajectory so that basic equations of motion are valid which are applicable in case of uniform acceleration. First two equations of motion with non zero jerk are given by: 
\begin{equation}
\label{eqm1} 
v = u + at + 1/2 jt^2
\end{equation}
and
\begin{equation}
\label{eqm2} 
s = ut + 1/2 at^2 + 1/6 jt^3
\end{equation}
where $s,v,u,a,j,t$ denotes displacement, final velocity, initial velocity, acceleration, jerk and time respectively. To neglect the jerk term we can minimize the $t$ which we are doing by temporal quantization with $\Delta t=0.01$ and we must also keep value of $j$ as minimum as possible for which we are applying knot shifting procedure. 

 \begin{figure}[thpb]
      \centering
      \includegraphics[width=\linewidth]{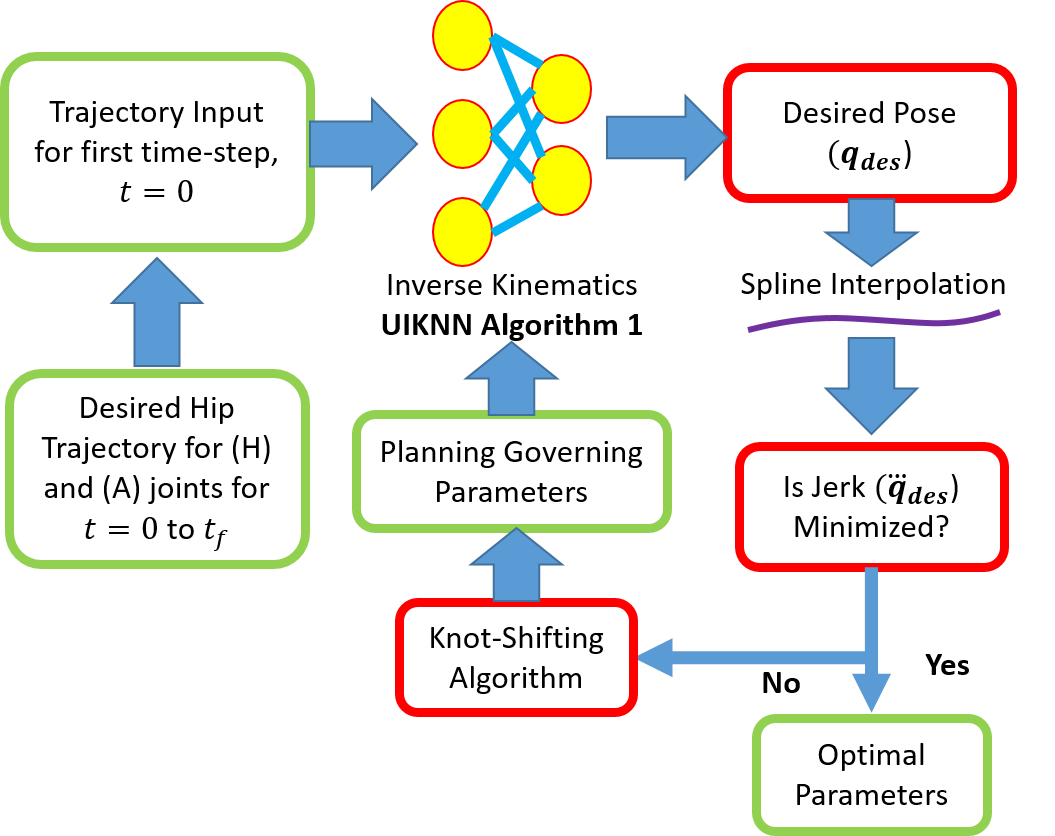}
      \caption{Knot Shifting IKNN Algorithmic Flow}
      \label{knot}
   \end{figure}
   
The trajectory planning section is governed by various parameters which can be changed in order to achieve optimal solution based on a suitable objective function. Furthermore, analysis over a complete trajectory seems futile and optimization at knot-points are enough to achieve the desired optimal solution. With these considerations and an objective function based on minimization of the absolute jerk faced in joint space, knot shifting procedure was implemented, as in Fig. \ref{knot}, such that a trajectory with relatively minimum jerk can be obtained. The aim of such an approach is to choose the governing parameters based on some real life applicable objectives and serves as an add-on to the trajectory planning methodology proposed in the study. Thus, suitable values of the parameters after implementing knot shifting are given in Section IX in Table \ref{tableparam}.

\section{CONTACT MODEL FOR GROUND REACTIONS} \label{cm}

In comparison to flat foot, toe foot biped model gives more realization to practical human walk mainly during transition from double support phase to single support phase as shown in previous sections for trajectory planning. As we have constraint our model in sagittal plane, so for contact modeling we have considered 6 contact points, 3 for each foot; one at toe end, one at sole and one at ankle point.

We have used virtual spring damper model , for modeling the contact forces with parameters $k_s$, $k_d$ and $\mu$ which represent the value of spring constant, damping constant and coefficient of friction respectively. These parameters has been taken apriori before the simulation.
If $\mathbf{J}^T$ represents the Jacobian transpose and $n_c$ denotes the number of contact points (6 in our case), $\mathbf{F}$ denotes the contact forces which have 2 components for $j^{th}$ contact point:
\begin{equation}
F_{n_j} = -k_s \Delta z - k_d \Delta\dot{z} \;\;\;\; \text{if} \;\;\; \Delta z \leq 0
\end{equation}
\begin{equation}
F_{x_j} = \mu F_{n_j}
\end{equation} 
Then extra term is added along with external joint torques given by:
\begin{equation}
\label{contact}
\mathbf{B} = \mathbf{J}^T\mathbf{F}
\end{equation}
where $\mathbf{J}^T$ is $9 \times 12$ matrix and $\mathbf{F}$ is $12 \times 1$ matrix.  
where  $\Delta z$ is the difference between ground level and contact point.

\section{DYNAMICS MODELING} \label{dm}
In previous sections we have planned a staircase trajectory for our Toe Foot Biped robot model and obtained inverse kinematics solution for the same using unsupervised feed forward artificial neural network in real time.

In this section, we are proposing a novel approach for dynamic model where we are combining traditional Lagrange's dynamics with temporal quantization of joint angles based on Inverse kinematics output using artificial neural network . \cite{c27} has proposed spatially quantized dynamics for biped robot model but considered Linear Inverted Pendulum Model (LIPM) which is approximation of real dynamics model and applied spatial quantization to hip trajectory. So to get a improved dynamic model, first, we will calculate Lagrange dynamics for our 9 link toe foot model and then using that dynamics equations, we will model our final Neural Network Temporal Quantized Lagrange Dynamics (NNTQLD) model.  

\subsection{Lagrange Formulation for Dynamics}
Euler-Lagrangian or simply Lagrangian formation is based on differentiation of energy terms with respect to system variables and time. It provides equations for getting relationship between position, velocity, acceleration and external force/torque for joint. As we are dealing with only revolute joints, we only need external torque for that joint.  For our biped model we have 9 system variables denoted by joint angle vector $\mathbf{q}=[q_1, q_2, q_3, q_4, q_5, q_6, q_7, q_8, q_9]^T$ where $q_i:\theta_i,i=1...9$ denotes respective joint angle as shown in Fig. \ref{modelpic}, $\mathbf{\dot{q}}$ denotes joint angle velocity vector, $\mathbf{\ddot{q}}$ represents joint acceleration vector and $\pmb{\tau} = [\tau_1, \tau_2, \tau_3, \tau_4, \tau_5, \tau_6, \tau_7, \tau_8, \tau_9]^T$ represents external torque vector. Then total kinetic energy $K$ of our model is function of $\dot{q}_i$ and total potential energy $P$ is function of $q_i$.

If $x_{com_i}$ and $z_{com_i}$ denotes the center of mass position of link $i$ for $i=1...9$, then for a particular link, by taking whole mass of link concentrated at center, its kinetic energy is given by:
\begin{equation}
k_i = \frac{1}{2}m_i.v_{com_i}^2 + \frac{1}{2}I_i.\dot{q}_i^2
\end{equation}
where $m_i$ and $I_i$ denotes the mass and moment of inertia respectively and $v_{com_i}$ denotes center of mass velocity for link $i$ given by:
\begin{equation}
v_{com_i}^2 = \dot{x}_{com_i}^2 + \dot{z}_{com_i}^2
\end{equation}
Similarly potential energy for a particular link is given by:
\begin{equation}
p_i = m .g. z_{com_i}
\end{equation} 
where $g$ denotes gravity constant. Center of mass equations of each link for our model are calculated as per our model.

Then total kinetic and potential energy of our model are given by the following equation:
\begin{equation}
K = \sum_{i=1}^9 k_i \;\;\; \text{and} \;\;\; P = \sum_{i=1}^9 p_i  
\end{equation} 
Then, Lagrangian can be defined by following equation:
\begin{equation}
L = K - P
\end{equation}
Then we can get the following generalized second order ordinary differential equation:
\begin{equation}
\frac{d}{dt}(\frac{\partial L}{\partial \dot{q_i}}) - \frac{\partial L}{\partial q_i}=\tau_i + B
\end{equation}
where $\tau_i$ denotes external torque applied for link $i$.

By solving above equations, we will get final robot equation in following form:
\begin{equation}
\label{eq1} 
\mathbf{M(q)\ddot{q}} + \mathbf{C(q,\dot{q})} + \mathbf{G(q)} = \pmb{\tau} + \mathbf{B} 
\end{equation}
where $\mathbf{M(q)}$ is $9 \times 9$ inertia matrix, $\mathbf{C(q,\dot{q})}$ is $9 \times 1$ coriolis matrix, $\mathbf{G(q)}$ is $9 \times 1$ gravity matrix and $\mathbf{B}$ is the contact model matrix. Robot dynamics has 2 beneficial properties:
\begin{itemize}
\item $\mathbf{M(q)}$ is symmetric, bounded and positive definite.
\item $\mathbf{\dot{M}(q)}\dot{q} - 2\mathbf{C(q,\dot{q})}$ is skew symmetric.
\end{itemize} 
 
\subsection{Temporal Quantized Lagrange Dynamics (TQLD)}
While calculating inverse kinematics, we are calculating $\mathbf{q}$ vector value at each time instant by taking time as uniformly quantized parameter given by:
\begin{equation}
t_k = \Delta t.k \;\;\; (k = 0,1,2...),
\end{equation}   
where $\Delta t$ is constant time step. At every time step, state vector $\mathbf{[q,\dot{q}]}$ is updated. For each time instant desired joint angles vector (pose vector) $\mathbf{q}_{des}$ is given by neural network based inverse kinematics and desired joint angles velocity vector (pose velocity vector) $\mathbf{\dot{q}}_{des}$ is calculated by finite difference method. So for a particular instant $k$, equation of motion can be written as:
\begin{equation}
\label{eq2} 
\mathbf{\dot{q}}_{k+1} =  \mathbf{\dot{q}}_k + \mathbf{\ddot{q}}_{k} \Delta t \;\; ,
\end{equation}
 
and
\begin{equation}
\label{eq3} 
\mathbf{q}_{k+1} =  \mathbf{q_k} + \mathbf{\dot{q}}_k (\Delta \mathbf{t} ) + \frac{1}{2}\mathbf{\ddot{q}}_{k} (\Delta \mathbf{t})^2 
\end{equation} 
Using equation (\ref{eq1}) for particular time instant $k$ we can have:
\begin{equation}
\label{eq4} 
\mathbf{\ddot{q}}_{k} = \mathbf{M^{-1}(q)}_k[\pmb{\tau}_k +\mathbf{B}_k - \mathbf{C(q,\dot{q})}_k \mathbf{\dot{q}}_{k} - \mathbf{G(q)}_k] 
\end{equation}
Using equation (\ref{eq4}) in (\ref{eq2}) and (\ref{eq3}), we have updated equations as:
\begin{equation}
\label{eq5}
\begin{split}
\mathbf{\dot{q}}_{k+1} =  \mathbf{\dot{q}}_k + \mathbf{M^{-1}(q)}_k[\pmb{\tau}_k +\mathbf{B}_k - \mathbf{C(q,\dot{q})}_k - \mathbf{G(q)}_k].(\Delta \mathbf{t})  
\end{split}
\end{equation}
\begin{equation}
\label{eq6} 
\begin{split}
\mathbf{q}_{k+1} =  \mathbf{q_k} + \mathbf{\dot{q}}_k (\Delta \mathbf{t}) + \frac{1}{2} \mathbf{M^{-1}(q)}_k[\pmb{\tau}_k +\mathbf{B}_k - \mathbf{C(q,\dot{q})}_k - \\ \mathbf{G(q)}_k].(\Delta \mathbf{t})^2 
\end{split}
\end{equation}
Equation (\ref{eq5}) and (\ref{eq6}) combining will give our Temporal Quantized Lagrange Dynamics (TQLD).

\begin{algorithm}[thpb]
\SetAlgoLined
\KwResult{State Transition}
$\pmb{\tau},\pmb{q}_C,\pmb{\dot{q}}_C$ To achieve Next State (N) from Current State (C) \;
$\pmb{\tau}  \gets \textit{Input Torque vector}$\;
$\pmb{q}_C  \gets \textit{Current Pose Vector}$\;
$\pmb{\dot{q}}_C \gets \textit{Current Rate of Change of Pose Vector}$\;
$\pmb{q}_N  \gets \textit{Next Pose Vector}$\;
$\pmb{\dot{q}}_N \gets \textit{Next Rate of Change of Pose Vector}$\;
\While{True}{
$\pmb{\ddot{q}}_C \gets Dynamics(\pmb{\tau},\pmb{q}_C,\pmb{\dot{q}}_C) \textit{ from (46)}$\;
$\pmb{\dot{q}}_N \gets \pmb{\dot{q}}_C + \pmb{\ddot{q}}_C \Delta t \textit{ from (47)}$\;
$\pmb{q}_N \gets \pmb{q_C} +  \pmb{\dot{q}}_C \Delta t + \frac{1}{2}\pmb{\ddot{q}}_C (\Delta t)^2 \textit{ from (48)}$\;
}
\Return $\pmb{q}_N, \pmb{\dot{q}}_N, \pmb{\ddot{q}}_C$\;
 \caption{Temporal Quantized Dynamics}
 \label{tqld}
\end{algorithm}

\section{ZMP STABILITY FORMULATION} \label{zmpf}

The Zero Moment Point (ZMP) based stability is a necessity for performing stable gaits. It is the point where the net moment of all the inertial and gravity forces along axes parallel to ground is equal to zero. The ZMP must lie inside the convex hull formed by the contact points of the biped with the floor and stability margin is defined as the distance of the actual ZMP from the boundaries of the convex hull. The farther the ZMP from all the boundaries, more is the stability margin and it is more feasible to execute the gait. For the SSP, the ZMP must lie within the portion of the stance leg's feet touching the ground whereas, for the DSP, it must be within the supporting polygon formed by all the contact points. The $x_{ZMP}$ was first formulated for a single mass LIPM model for the kinematic trajectory planning and now, it is assured that the model executes the desired trajectory as much as possible maintaining the actual ZMP considerations, for which $x_{ZMP}^{actual}$ is calculated as,

   \begin{figure*}[thpb]
      \centering
      \includegraphics[width=\linewidth]{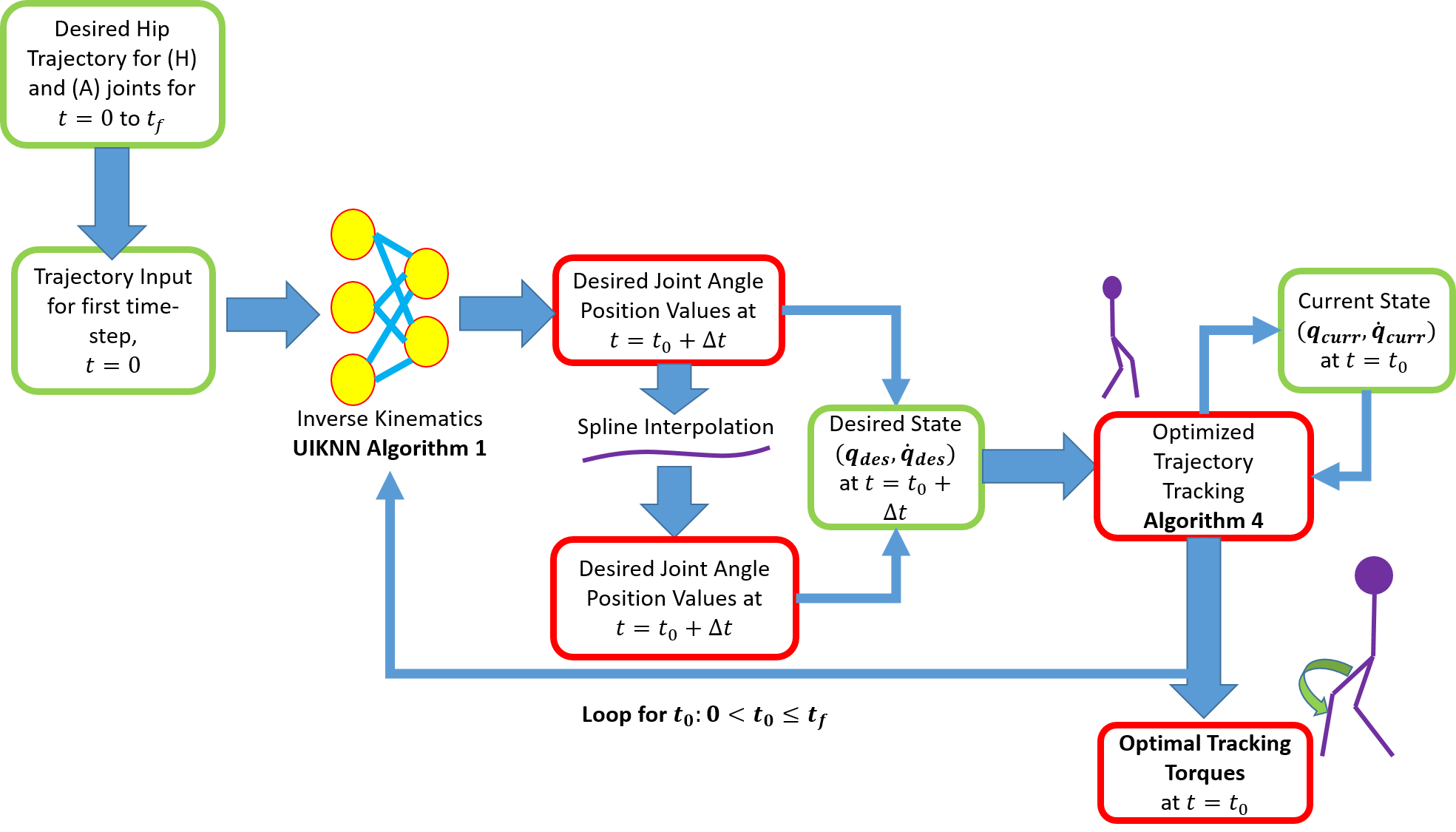}
      \caption{Complete NNTQLD based Trajectory Tracking Optimization}
      \label{NNTQLD}
   \end{figure*}
   
\begin{equation}
x_{ZMP}^{actual} = \frac{\sum^n_{i=1} m_i x_i(\ddot{z}_i + g) - \sum^n_{i=1} m_i \ddot{x}_i z_i}{\sum^n_{i=1} m_i(\ddot{z}_i + g) - k \sum^n_{i=1} m_i \ddot{x}}
\label{zmp}
\end{equation}
where $\ddot{x}_i$ and $\ddot{z}_i$ is calculated in Cartesian space from the  joint angle accelerations given by (\ref{eq4}) in the joint space and $k$ is the virtual slope considered for modeling the CG motion based on one-mass COG (Center of Gravity) model.
\section{Controller Design} \label{cd}
We have considered PD controller to control the joint torques. Expression for the same is given by:
\begin{equation}
    \tau = K_p e + K_d \dot{e}
\end{equation}
where $\tau$ is actual torque supplied by controllers to the joints, $K_p$ and $K_d$ represents proportional and derivative gains. $e_k$ and $\dot{e}_k$ are joint position and joint velocity errors given by following equations:
\begin{equation}
    e = q_d - q
\end{equation}
\begin{equation}
    \dot{e} = \dot{q}_d - \dot{q}
\end{equation}
where $q_d$ and $\dot{q}_d$ denotes $9 \times 1$ desired joint angle vector and desired joint angle velocity vector respectively.
\subsection{Torso Pitch Angle Optimization}
Torso plays a very significant role in case of staircase walk as most of the bipedal robot mass is constituted by it. Torso pitch angle, $q_5$ in our case can highly effect the stability and can prevent the bipedal robot from toppling down specially in case of ascending stairs. So it should be chosen in such a way that it can increase the stability margin as well as providing a more energy efficient gait.   
\subsection{Ant Colony Optimization} \label{ACO}
Ant Colony Optimization (ACO) \cite{c28, c29} works in the same manner as the ants trying to find sources of food. They do so by selecting shortest path among multiple paths to reach their target. for the purpose of communication, pheromones deposition is the key factor whose amount depict the number of ant followed a particular path. It acts as basis of attraction to follow a particular path which finally results in all ants following shortest path. In past, ACO has been used to tune parameters of PID (Proportional-Integral Derivative) and MPC (Model Predictive) Controllers \cite{c30, c31}. We will use ACO to tune PD controller parameters and torso pitch angle satisfying ZMP constraint with objective function for time instant $k$ as follows: 
\begin{equation}
\begin{split}
\text{Minimize}\;\;\; \sum_{\forall \text{joints}}[|\mathbf{q}_k - \mathbf{q}^{des}_k + |\mathbf{\dot{q}}_k - \mathbf{\dot{q}}^{des}_k| +|\mathbf{\tau}_k\mathbf{\dot{q}}_k|]
\end{split}
\end{equation}

along with maximizing ZMP Stability Margin based on equation \ref{zmp} and subjected to calculating $\mathbf{q}_k$ and $\mathbf{\dot{q}}_k$ from equations (\ref{eq5}) and (\ref{eq6}) for the torques $\pmb{\tau}_k$. First two terms in objective function are for minimizing joint space trajectory errors and third term is basically the instantaneous power which is minimized to finally get a more energy efficient gait.

\begin{algorithm}[thpb]
\SetAlgoLined
\KwResult{Satisfy Constraints}
$(\pmb{q}_C,\pmb{\ddot{q}}_C)$\;
$\pmb{q}_C  \gets \textit{Current Pose Vector}$\;
$\pmb{\ddot{q}}_C \gets \textit{Pose Vector Acceleration}$\;
$\mathbf{x,\ddot{x},z,\ddot{z}} \gets \text{Cartesian position and acceleration}$\;
$M \gets \textit{Weight of all masses}$\;
$ZMP_{fmax} \gets \textit{Max feasible ZMP position}$\;
$ZMP_{fmin} \gets \textit{Min feasible ZMP position}$\;
\While{True}{
$\mathbf{x,\ddot{x},z,\ddot{z}} \gets func(\pmb{q}_C,\pmb{\ddot{q}}_C)$\;
$ZMP \gets calculateZMP (M,\mathbf{x,\ddot{x},z,\ddot{z}}) \text{ from (48)}$\;
}
\Return $ZMP_{fmin} \leq ZMP \leq ZMP_{fmax}$\;
 \caption{Satisfying ZMP Constraints}
 \label{constraints}
\end{algorithm}

\begin{algorithm}[thpb]
\SetAlgoLined
\KwResult{Determine Optimal Joint Torques}
$\pmb{\tau}  \gets \textit{Random Initial Torque}$\;
$\pmb{q}_C  \gets \textit{Current Pose Vector}$\;
$\pmb{\dot{q}}_C \gets \textit{Current Pose Velocity Vector}$\;
$\pmb{\ddot{q}}_C \gets \textit{Pose Vector Acceleration}$\;
$\pmb{q}^{des}_N  \gets \textit{Desired Next Pose Vector}$\;
$\pmb{\dot{q}}^{des}_N \gets \textit{Desired dNext Pose Velocity Vector} $\;
\While{True}{
$\pmb{q}_N, \pmb{\dot{q}}_N, \pmb{\ddot{q}}_C \gets StateTransition(\pmb{\tau},\pmb{q}_C,\pmb{\dot{q}}_C)$\;
Minimize $O_1 = |\pmb{q}_N -\pmb{q}^{des}_N|+|\pmb{\dot{q}}_N -\pmb{\dot{q}}^{des}_N| + |\pmb{\tau}_N \pmb{\dot{q}}_N|$ \;
Such that $SatisfyConstraints(\pmb{q}_C,\pmb{\ddot{q}}_C) \text{ is True}$\;
Stop if $min(O_1)$ is reached for $\pmb{\tau}_{optimal}$\;
}
\Return $\pmb{\tau}_{optimal},\pmb{q}^{optimal}_N, \pmb{\dot{q}}^{optimal}_N $\;
 \caption{Ant Colony Optimization}
 \label{track}
\end{algorithm}

\section{RESULT AND DISCUSSIONS} \label{rad}

\begin{figure}[thpb]
      \centering
      \includegraphics[width=0.8\linewidth]{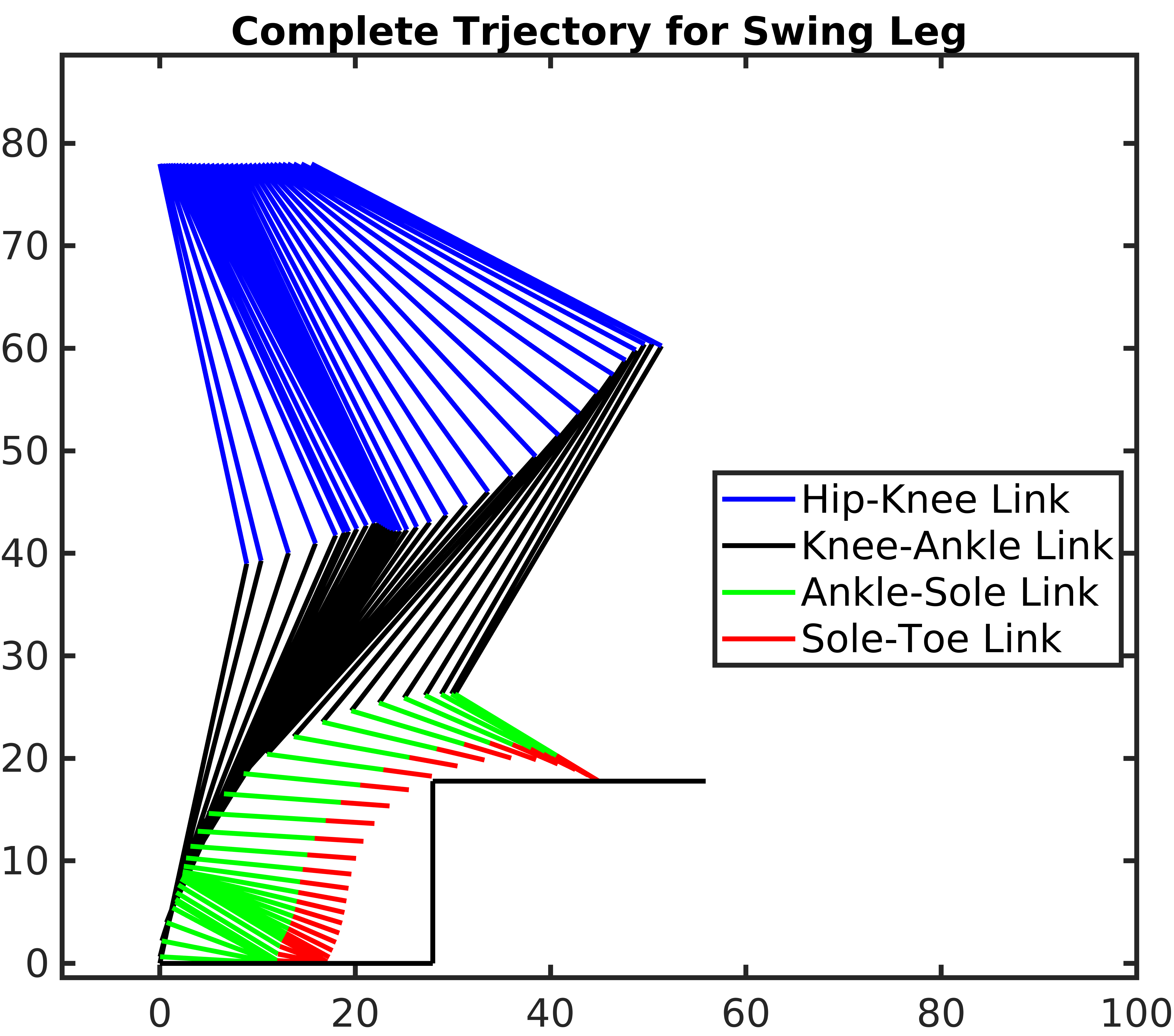}
      \caption{Bipedal Swing Leg Motion for Upstair Climbing}
      \label{Result:upswing}
   \end{figure}
   
\begin{figure}[thpb]
      \centering
      \includegraphics[width=0.8\linewidth]{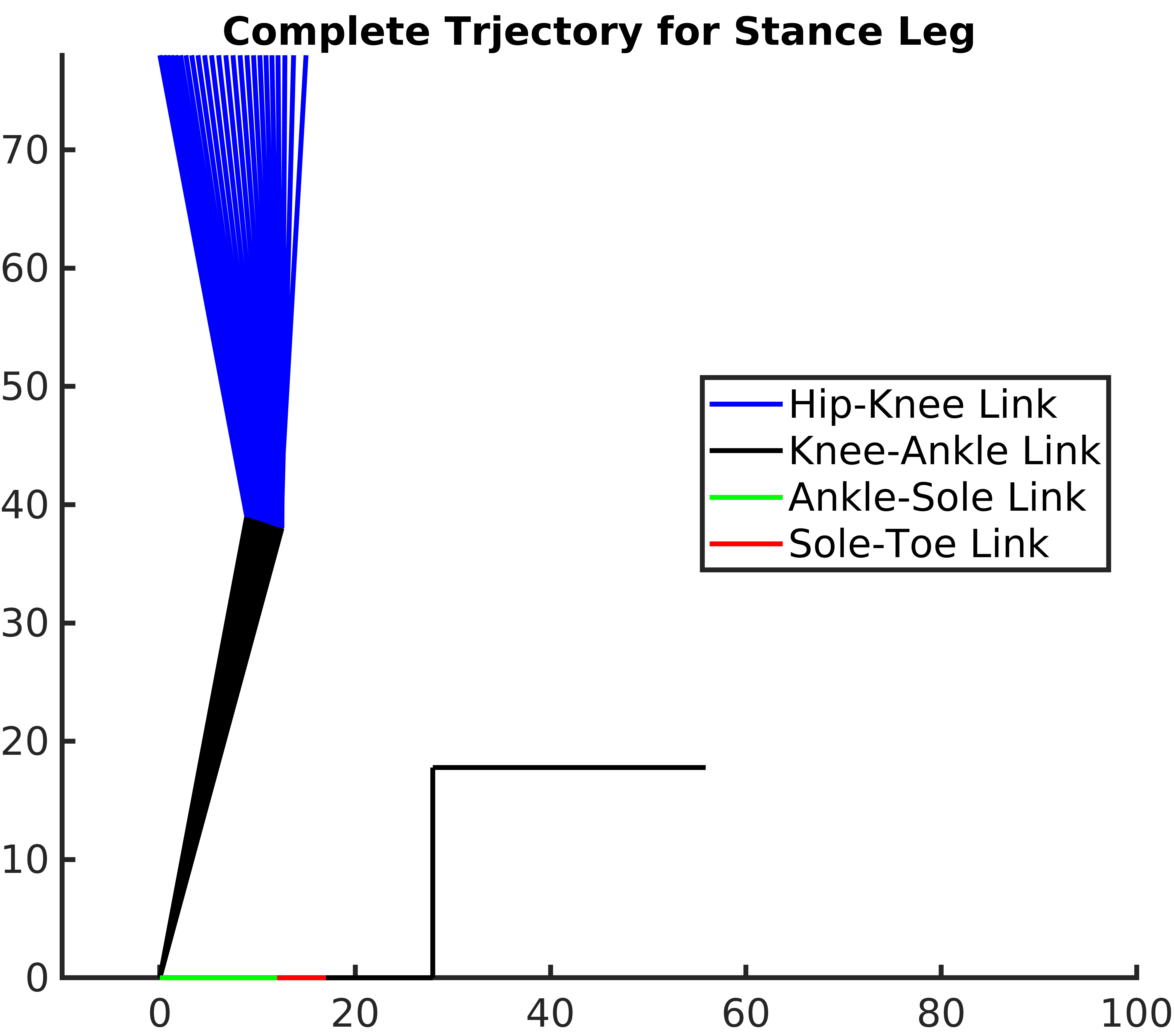}
      \caption{Bipedal Stance Leg Motion for Upstair Climbing}
      \label{Result:upstance}
   \end{figure}

In this section, the simulation results of the considered toe-foot biped model based on the proposed methodology is presented. All the simulations are performed using \copyright MATLAB and CPU computations on an Intel\textregistered i7-7500U CPU@2.70GHz. The simulation consists of a complete upstairs climbing starting from the first step to gradually climbing the subsequent steps.

The trajectory planning starts with the Double Support Phase(DSP) of the active leg, as shown in Fig. \ref{Result:DSP}, and gradually gets continued to the Single Support Phase(SSP) based on the knot shifting algorithm in Fig. \ref{knot}. Based on the knot shifting IKNN optimization the planning governing parameters are obtained as shown in Table \ref{tableparam}.   

\begin{table}[h]
\caption{Optimal Value of Trajectory Governing Parameters}
\label{tableparam}
\begin{center}
\begin{tabular}{|c|c|c|c|}
\hline
Parameter & Value & Parameter & Value\\
\hline
$t_1$ & 0.50 sec & $t_3$ & 3.50 sec\\
\hline
$t^p_1$ & 0.80 sec & $\theta_a$ & $\pi/6$ rad\\
\hline
$t^p_2$ & 1.10 sec & $\theta_b$ & $\pi/6$ rad\\
\hline
$t_2$ & 1.40 sec & $\Delta \theta_{c0}$ & 0.01 rad\\
\hline
$t_{bc}$ & 2.20 sec & $z_{Ci}$ & 0.78 meters\\
\hline
\end{tabular}
\end{center}
\end{table}

\begin{figure}[thpb]
      \centering
      \includegraphics[width=0.8\linewidth]{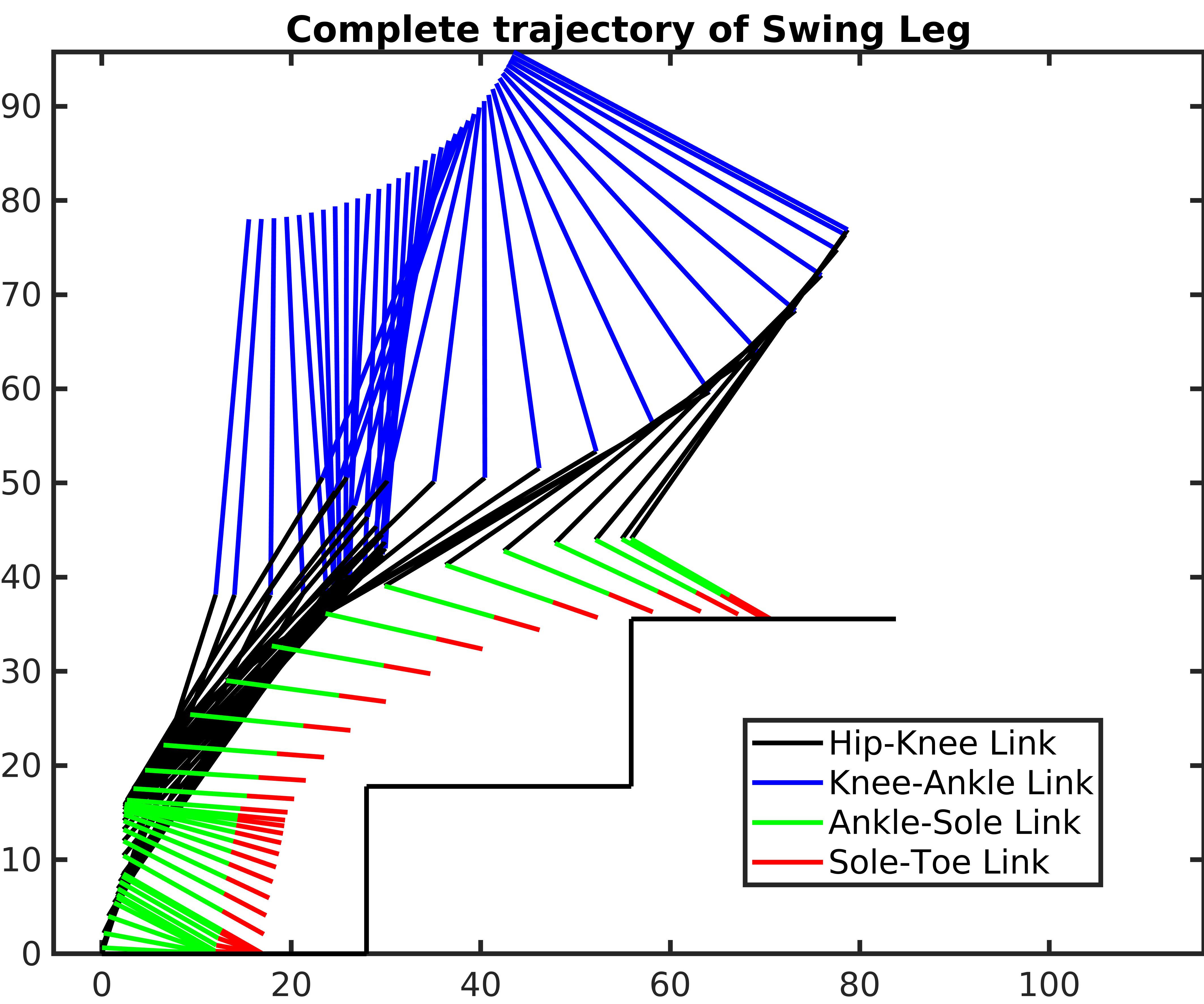}
      \caption{Bipedal Swing Leg Motion for Subsequent Upstair Climbing}
      \label{Result:iupswing}
   \end{figure}
   
\begin{figure}[thpb]
      \centering
      \includegraphics[width=0.8\linewidth]{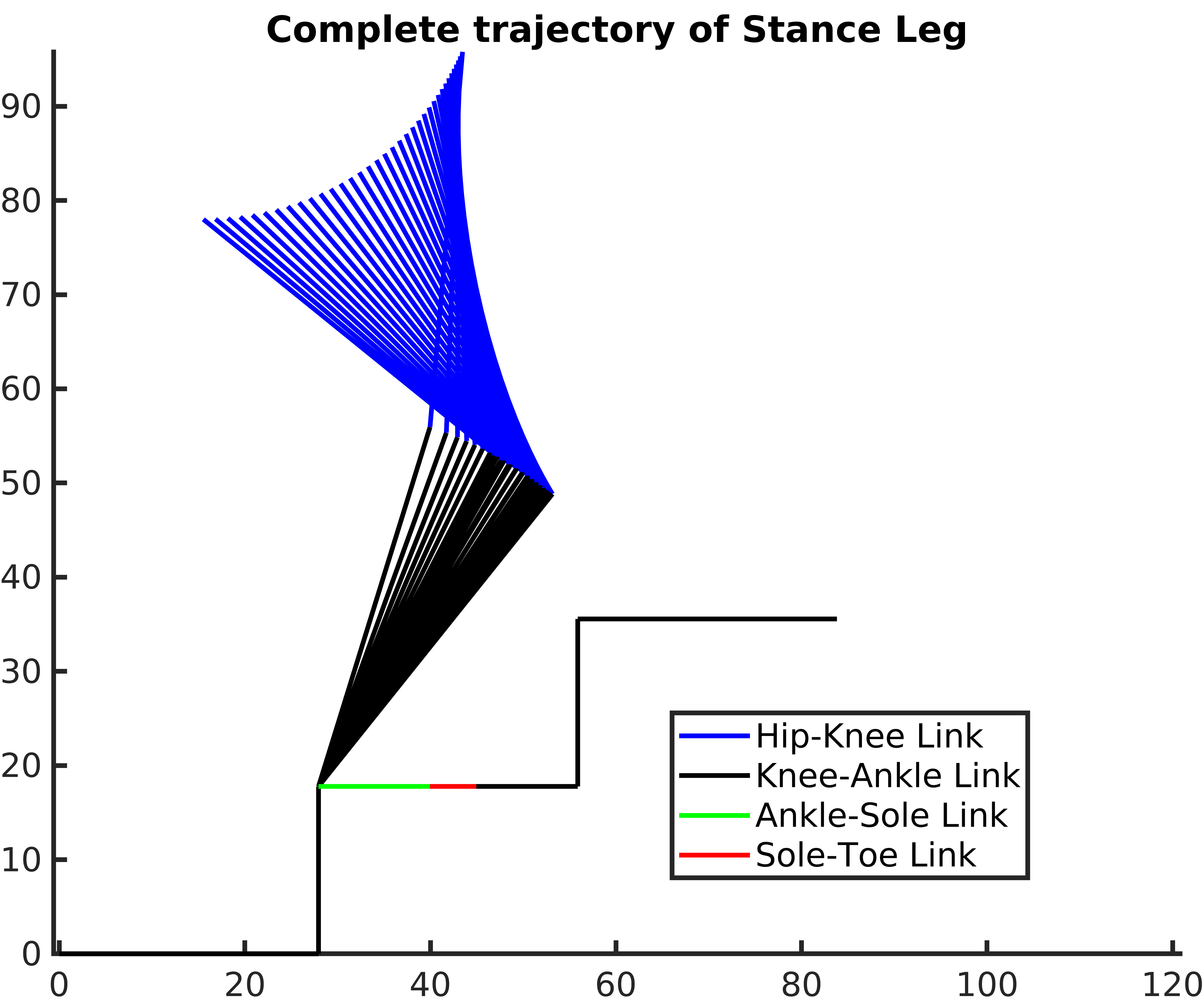}
      \caption{Bipedal Stance Leg Motion for Subsequent Upstair Climbing}
      \label{Result:iupstance}
   \end{figure}

Using the optimized parameter values, the trajectory of Hip(H) and Ankle(A) was constructed and Algorithm \ref{iknn} was used to construct the complete joint angle trajectories. The algorithm would be same in trajectory planning for a single step as well as for the following subsequent steps. The only difference will be the effective length and height of the stairs as upstairs witnesses no change in hip height and a single step is simulated but for the subsequent steps the double step climbing is accompanied by a rise in hip altitude as well. For the sake of simplicity and due to space constraints, the final algorithm in action is shown for both single and double step climbing  whereas the final trajectory values are shown graphically for the double step climbing only.

The only constraint while climbing the first step is that the ZMP-must lie within the span of the stance foot. Thus, it makes climbing the first step relatively easy. Even the movement does not witnesses any change in the hip altitude. The model considered in the study resembles a lower body framework of a mature male of hip height 80 cm. The effect of considering such a model can be visualized from Fig. \ref{Result:upswing} where the swing leg combined with a cyloidal realization based trajectory is shown in action. 

The corresponding action of the passive leg to support the hip movement accompanying with the active leg swing motion is shown in Fig. \ref{Result:upstance}. The ankle of the stance leg is fixed and moves only after the reverse-DSP for the swing leg finishes. This motion continues to the swing motion for the subsequent step climbing where the passive leg now becomes active and vice-versa.

The stance leg after the completion of the reverse-DSP in the single step phase undergoes a similar planning procedure based on double step climbing targets. The key importance of this phase is to ensure that the ZMP reaches the stance leg foot area on the first step during the DSP such that the model can lift the swing leg maintaining proper stability margin. The importance of proper trajectory governing parameter optimization can be realized where a trade off between the duration of DSP and the position of hip, as soon as the DSP commences, is observed. Furthermore, it should also be noted that during this phase, if the hip moves too far then it may not be possible for a 2 link manipulator with base at hip to follow the desired ankle trajectory. Thus the planning has multiple objectives to satisfy based on the ZMP stability considerations as well as the ankle position must be within the feasible workspace of a 2 link manipulator with its base at the hip as well as satisfying joint angle constraints.

\begin{figure}[thpb]
      \centering
      \includegraphics[width=0.8\linewidth]{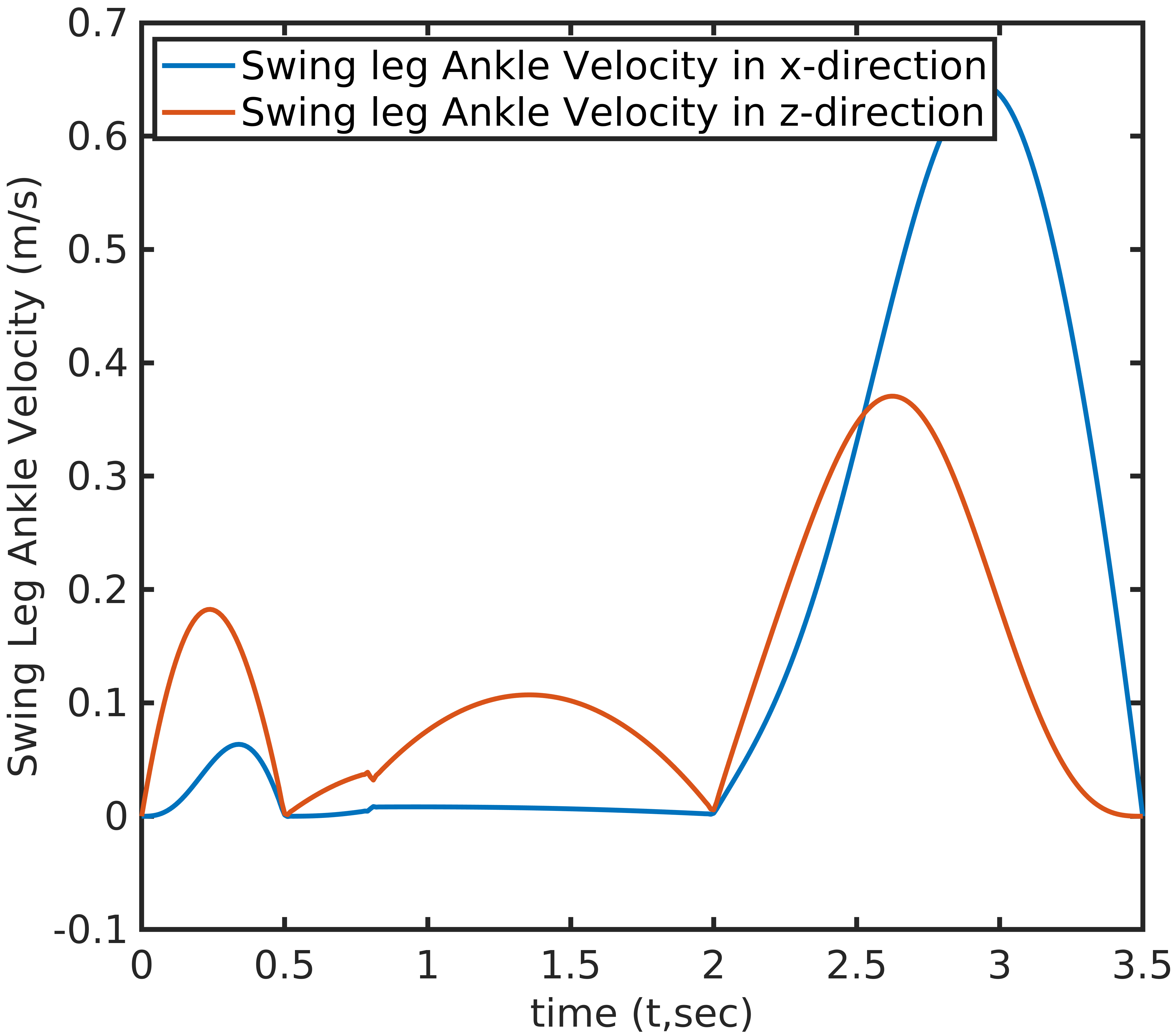}
      \caption{Swing Leg's Ankle x,z-velocity profile for Subsequent Upstair Climbing}
      \label{Result:iuankvel}
   \end{figure}
   
\begin{figure}[thpb]
      \centering
      \includegraphics[width=0.8\linewidth]{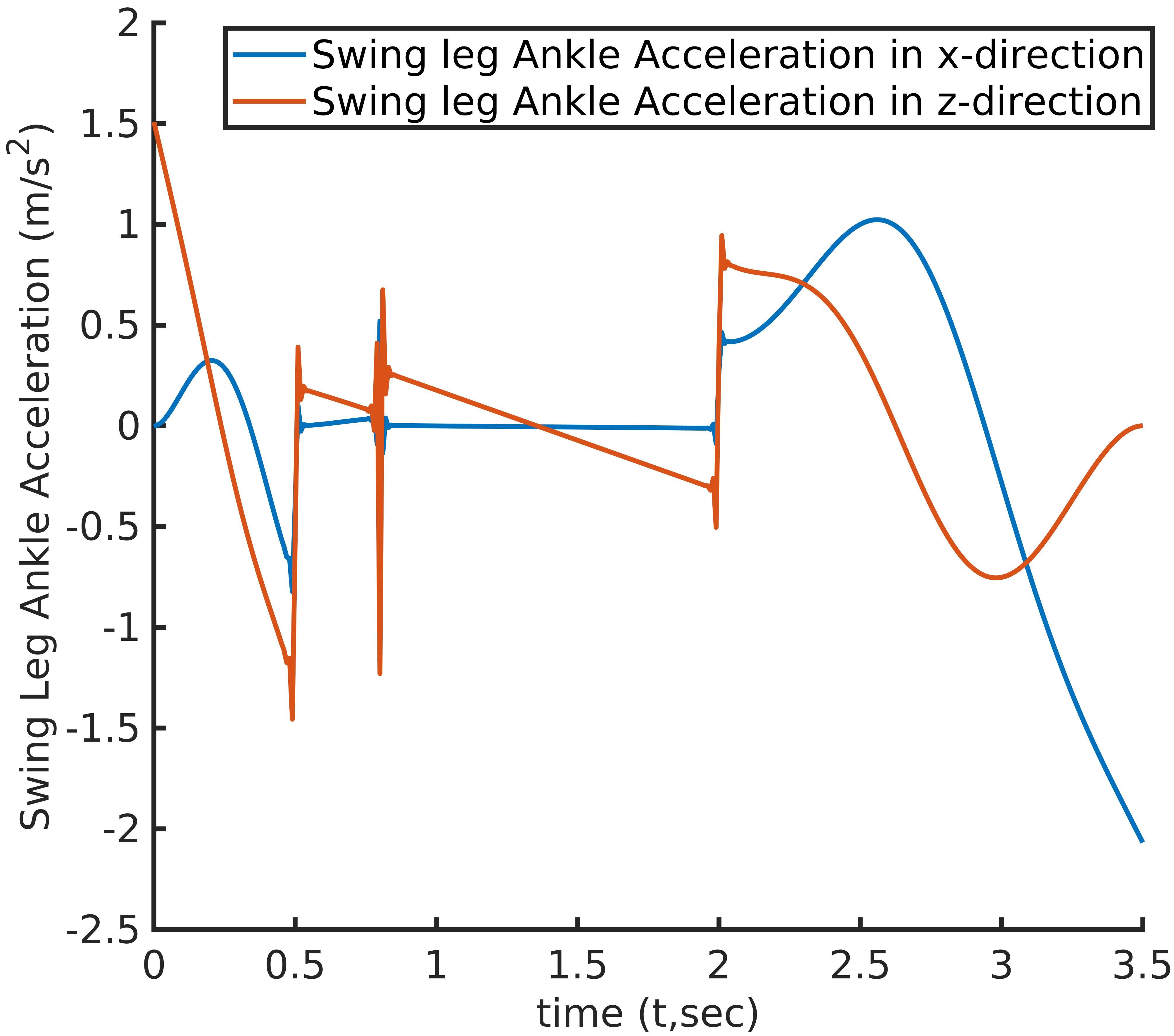}
      \caption{Swing Leg's Ankle x,z-acceleration profile for Subsequent Upstair Climbing}
      \label{Result:iuankacc}
   \end{figure}
   
\begin{table*}[!b]
\caption{Staircase Parameters and Concerning Optimization Results for all three cases}
\label{tableresults}
\begin{center}
\begin{tabular}{|c|c|c|c|}
\hline
Case & 1 & 2 & 3 \\
\hline
Stair Length & 55.88 cm & 50.80 cm & 55.88 cm\\
\hline
Rise-Run Ratio & 0.64 & 0.7000 & 0.72\\
\hline
Max-Acceleration & 42.77 m/s$^2$ & 34.86 m/s$^2$ & 42.75 m/s$^2$\\
\hline
Max-Jerk & 1894.6 m/s$^3$ & 1453.1 m/s$^3$ & 3406.6 m/s$^3$\\
\hline
Number of Ants & 30 & 30 & 30 \\
\hline
Number of Iterations & 100 & 100 & 100 \\
\hline
Evaporation Rate & 0.7 & 0.7 & 0.7 \\
\hline
Torso Pitch Angle($q_5$) & 1.25847 rad & 1.2055 rad & 1.2728 rad \\
\hline
$K_p$ & 23.3027 & 2.6474 & 32.0680 \\
\hline
$K_d$ & 86.2633 & 7.3724 & 86.7533 \\
\hline
Cost & 20.6346 & 14.9627 & 36.8319 \\
\hline
$K_p$ & 23.3027 & 2.6474 & 32.0680 \\
\hline
Total Energy Consumption & $1.0192 \times 10^5$ & $8.7396 \times 10^4$ & $9.6890 \times 10^4$ \\
\hline
Peak Torque (N-m) & 114.5284 & 35.6167 & 60.3183 \\
\hline
\end{tabular}
\end{center}
\end{table*}
   
\begin{figure*}[thpb]
\centering
\begin{subfigure}{.30\textwidth}
  \centering
  \includegraphics[width=\linewidth]{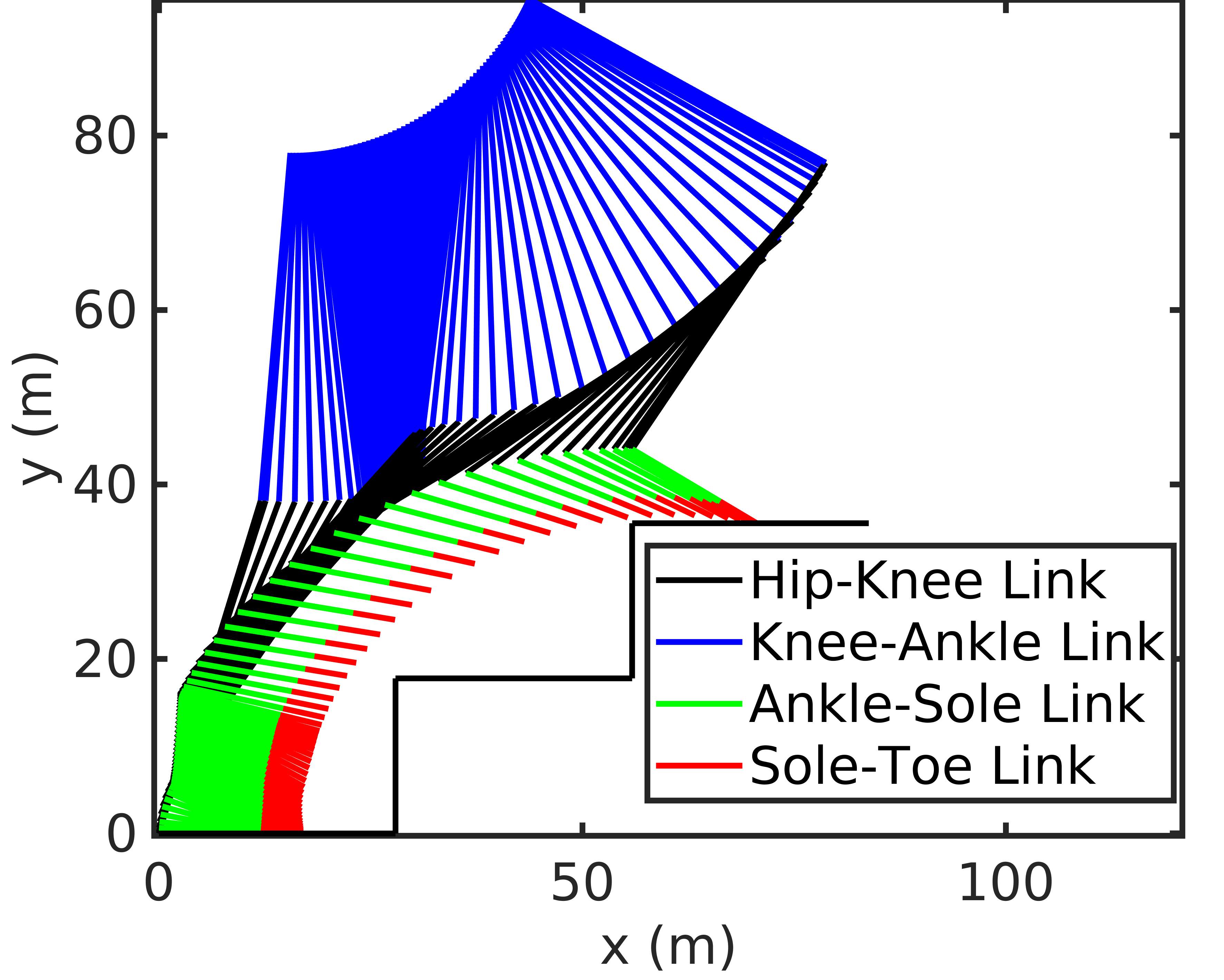}
  \caption{For Case 1}
\label{fig:case1s}
\end{subfigure}
\begin{subfigure}{.30\textwidth}
  \centering
  \includegraphics[width=\linewidth]{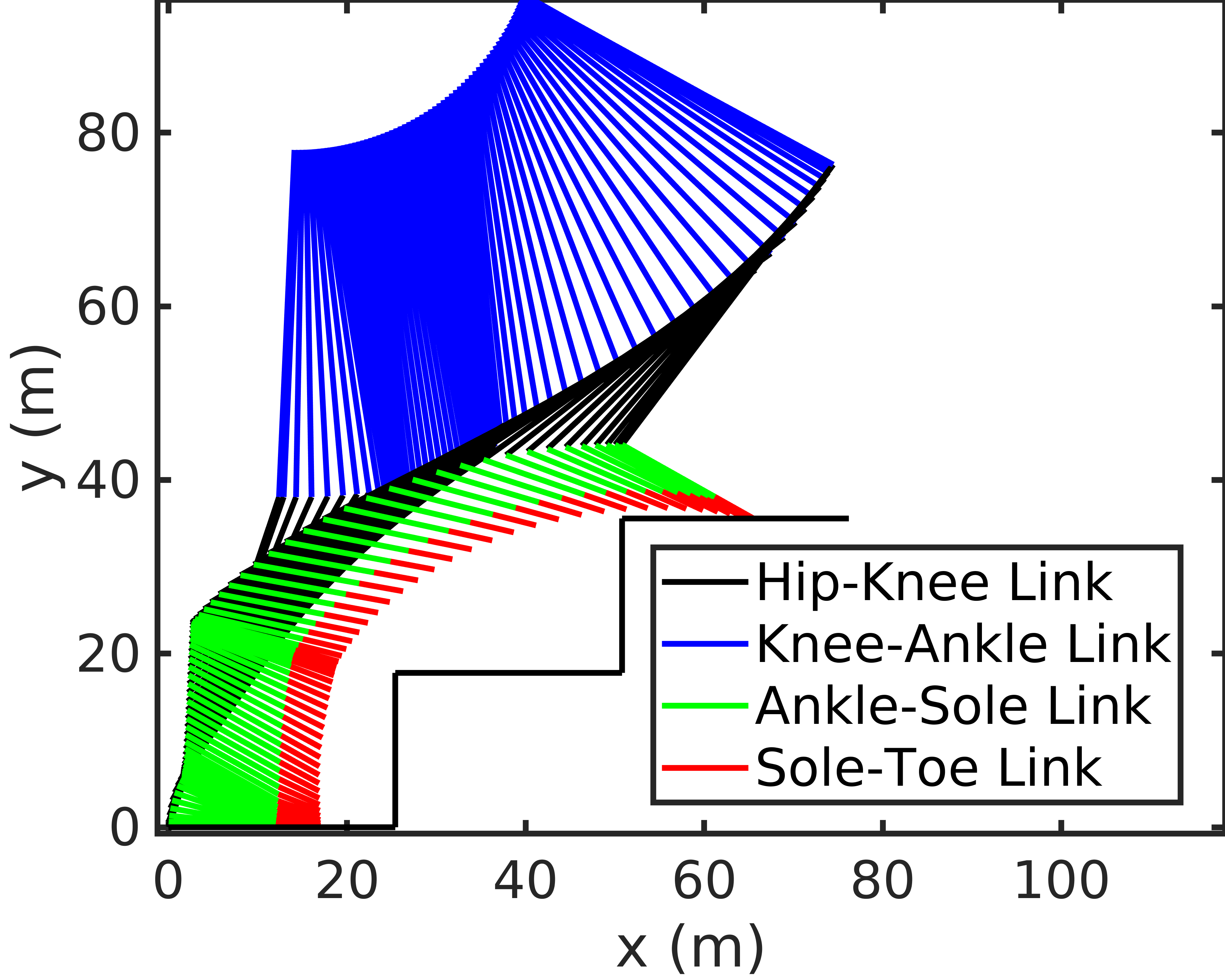}
  \caption{For Case 2}
\label{fig:case2s}
\end{subfigure}
\begin{subfigure}{.30\textwidth}
  \centering
  \includegraphics[width=\linewidth]{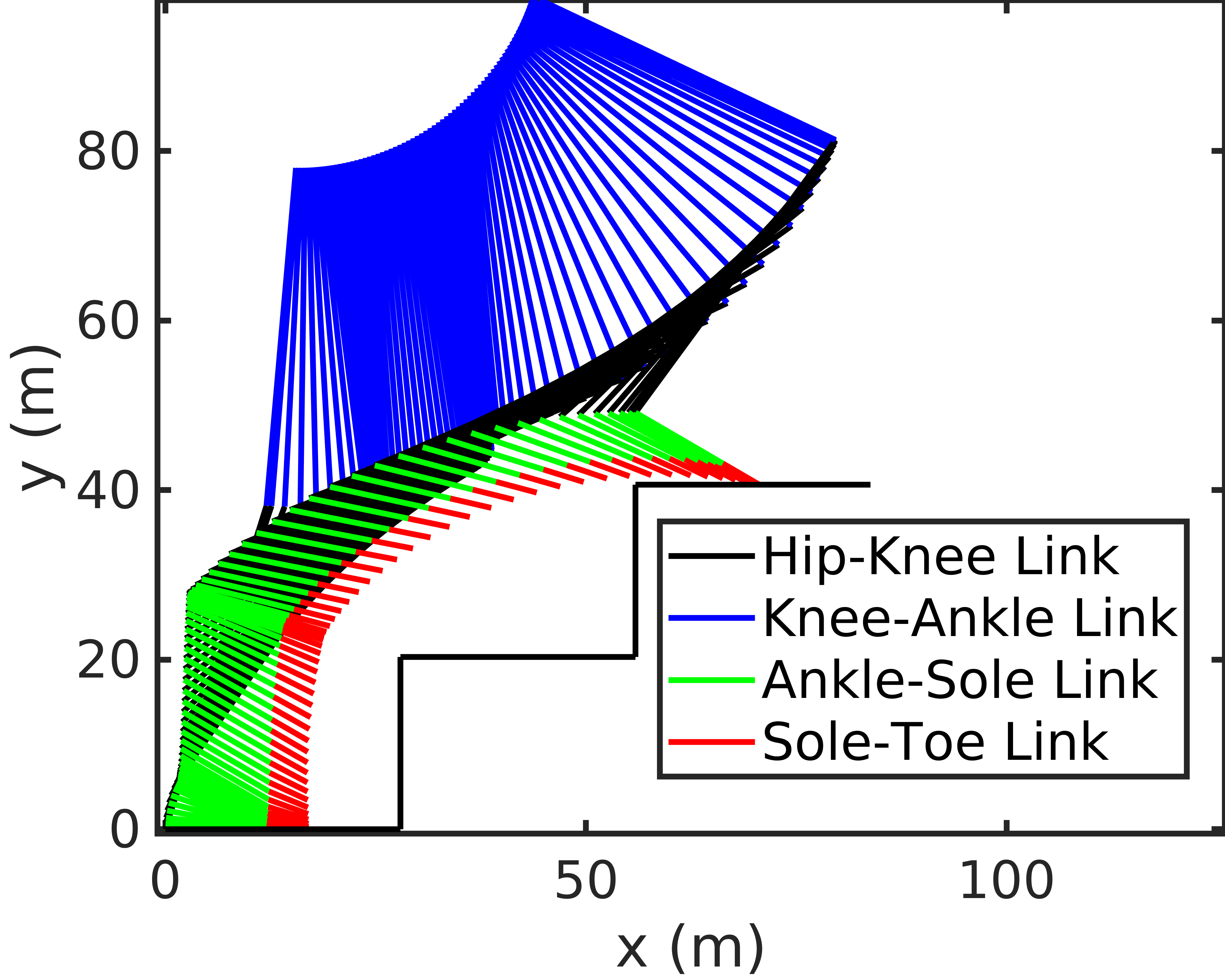}
  \caption{For case 3}
\label{fig:case3s}
\end{subfigure}
\caption{Comparison of Swing Leg Trajectory w.r.t stair dimensions for all three cases}
\label{fig:swingcomp}
\vspace{-2mm}
\end{figure*}

The trajectory for ankle was planned in three segments as shown in Fig. \ref{Result:iupsegment}, namely, the DSP, the Cycoidal segment as well as the bezier bridge segment connecting the former two. The control points of the bezier curve are chosen such that it completely meshes with the DSP as well as the cycloid and supports continuity at the transitions.

\begin{figure}
  \centering
  \includegraphics[width=0.8\linewidth]{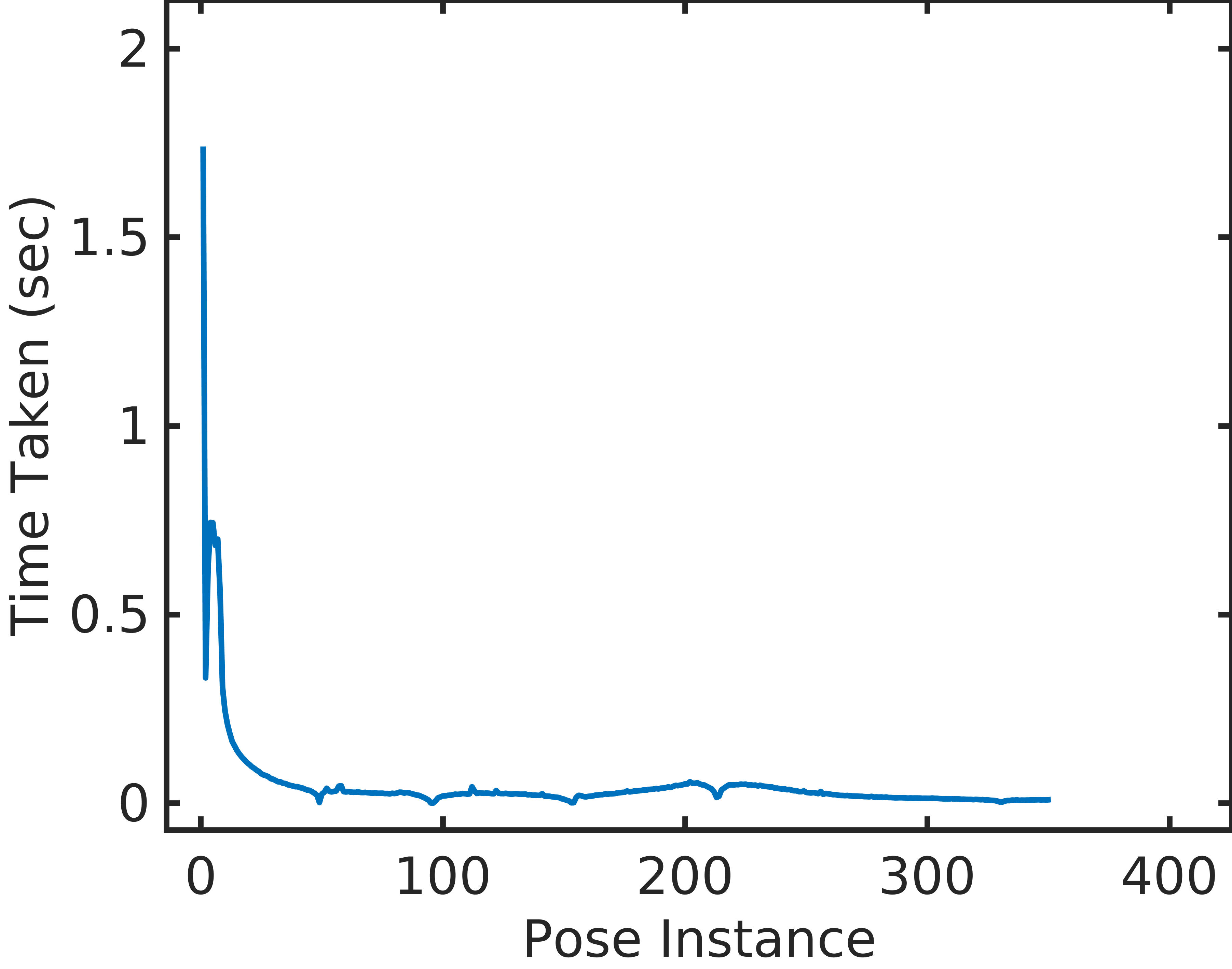}
  \caption{Computational Time per Pose}
\label{fig:ctpp}
\end{figure}

For subsequent steps, we have taken 3 different cases with different Rise/Run ratio as mentioned in Table \ref{tableresults} with given ACO parameters. An external disturbance term $\frac{1}{70} sin(2t)$ is considered. The maximum acceleration and maximum jerk considering all joint angle trajectories are mentioned for every case. Based on the computation results, the acceleration and jerk metric were found to be minimum for Case 2 with 0.700 rise/run ratio and values 34.86 m/s$^2$ and 1453.1 m/s$^3$ respectively. respectively. Figures \ref{fig:case1s}, \ref{fig:case2s} and \ref{fig:case3s} shows the obtained swing leg trajectories for subsequent steps for all the three cases. It is worth to be mentioned that a slight change in Rise/Run ratio may lead to a substantial change in jerk profile which is quite evident from Table \ref{tableresults}, that is, when we increase this ratio from 0.64 to 0.70, maximum jerk tends to decrease from 1894.6 m/s$^3$ to 1453.1 m/s$^3$ and further if we again increase Rise/Run ratio slightly from 0.70 to 0.72, maximum jerk obtained again increases drastically from 1453.1 m/s$^3$ to 3406.6 m/s$^3$. .Figure \ref{fig:ctpp} shows Case 1 UIKNN computation time for every discrete pose instant.At first instant, when error is very high, computation time is quite high but after first instant on wards, when neural network weights are trained, computation time decreases drastically and becomes suitable enough for real time tracking after few pose instants only which makes this approach better in comparison to traditional methods as well as supervised neural network methods.  

\begin{figure*}[thpb]
\centering
\begin{subfigure}{.30\textwidth}
  \centering
  \includegraphics[width=\linewidth]{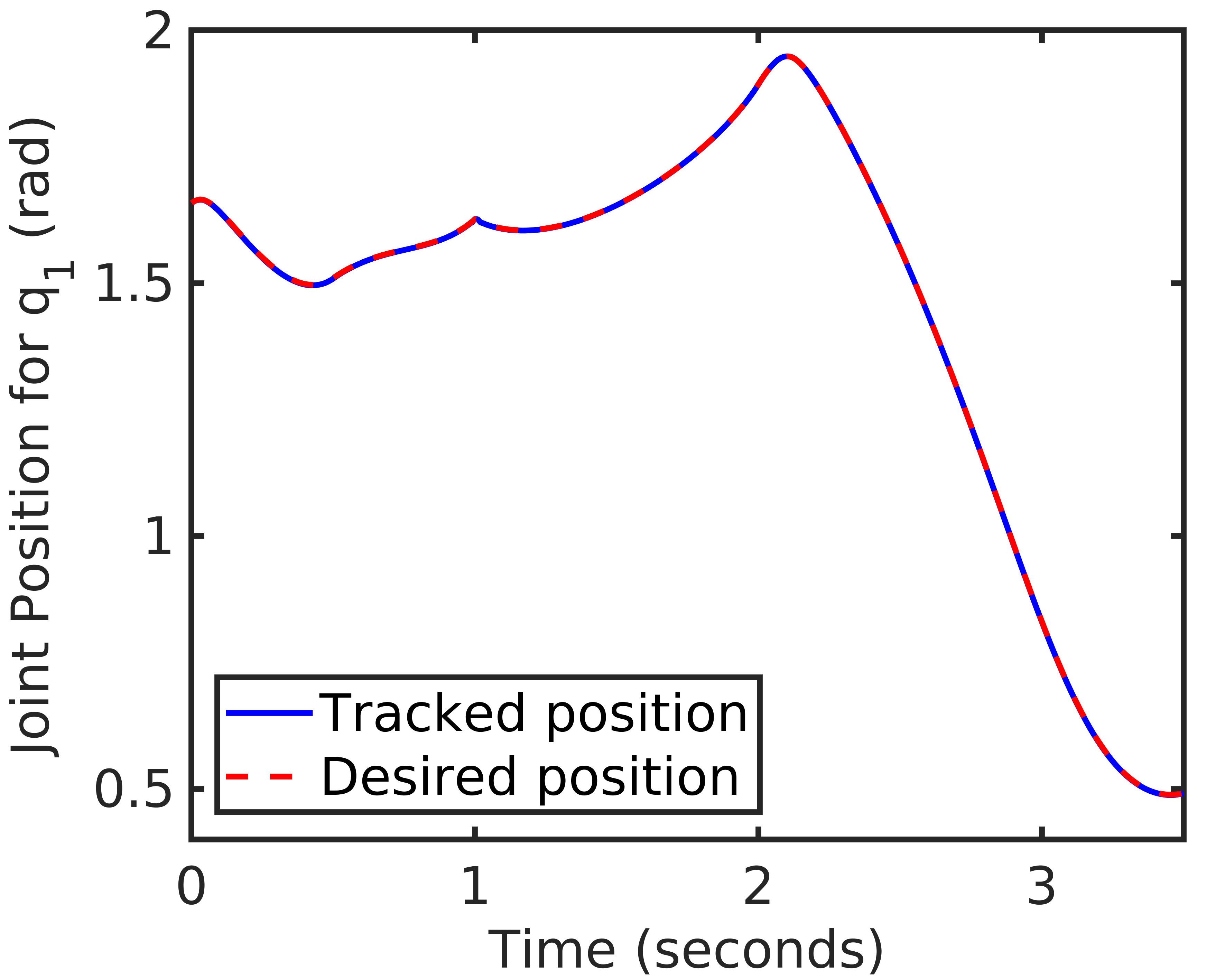}
  \caption{For Case 1}
\label{fig:case1s}
\end{subfigure}
\begin{subfigure}{.30\textwidth}
  \centering
  \includegraphics[width=\linewidth]{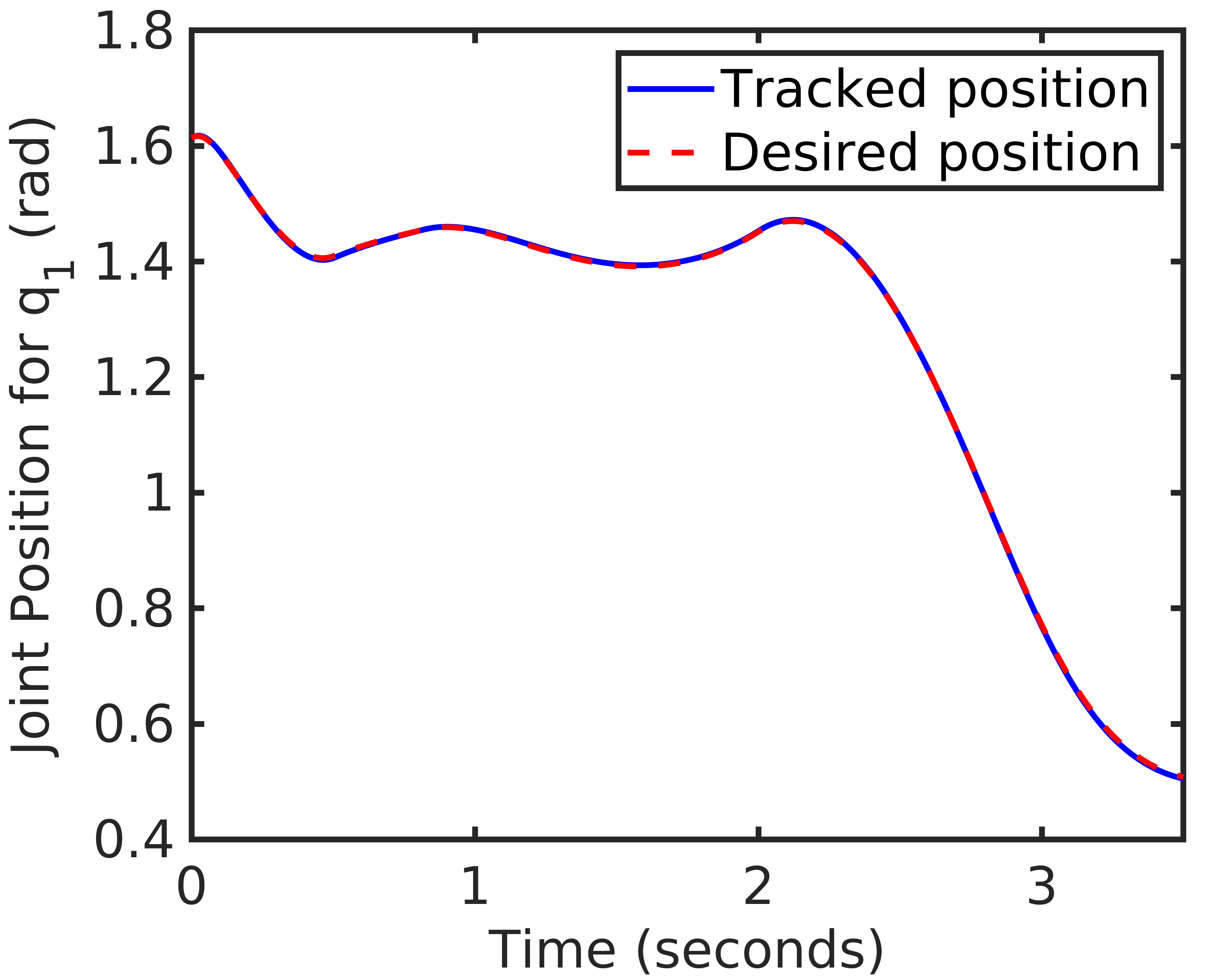}
  \caption{For Case 2}
\label{fig:case2s}
\end{subfigure}
\begin{subfigure}{.30\textwidth}
  \centering
  \includegraphics[width=\linewidth]{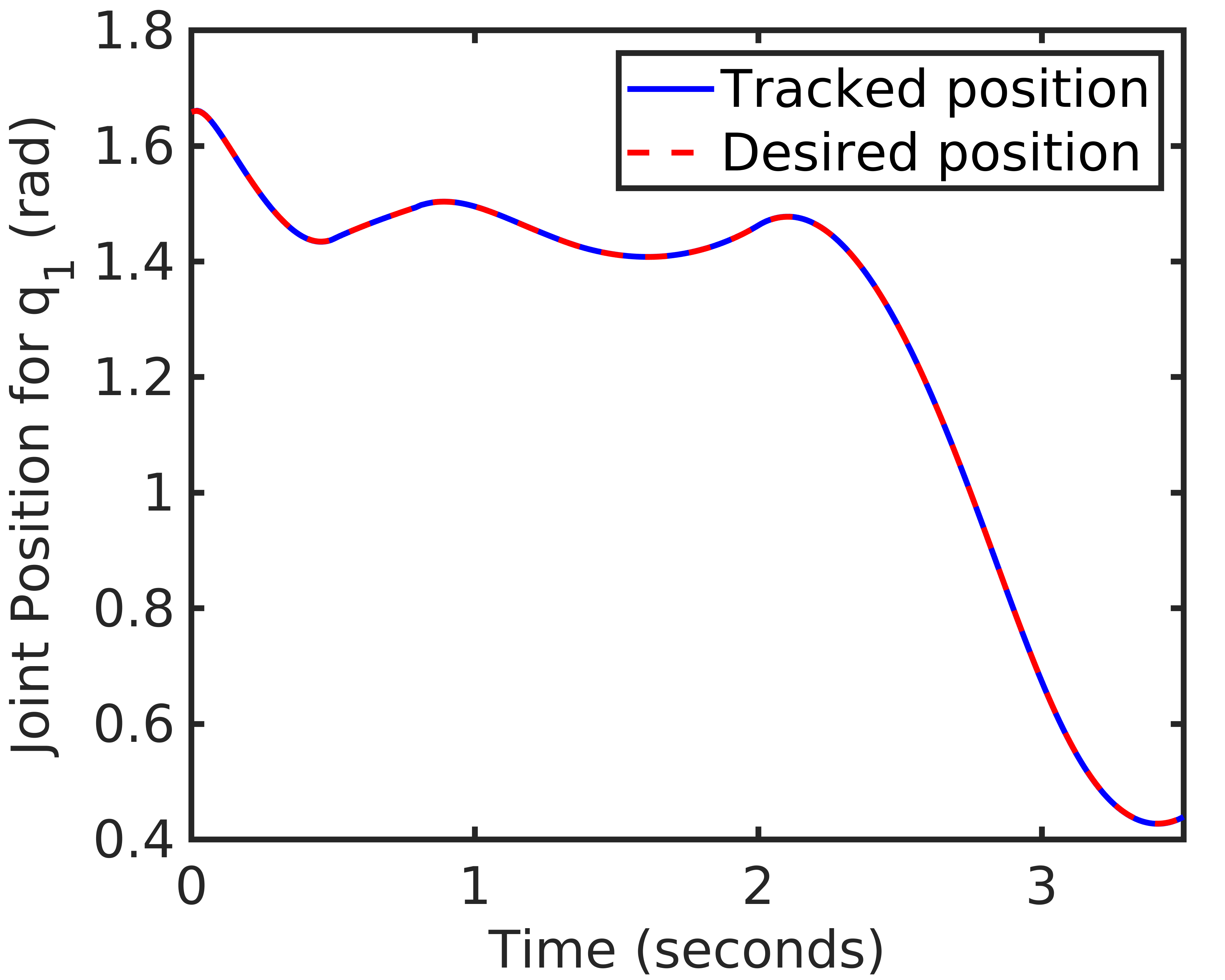}
  \caption{For case 3}
\label{fig:case3s}
\end{subfigure}
\caption{Tracking Performance $q_1$}
\label{fig:q1tp}
\vspace{-2mm}
\end{figure*}

\begin{figure*}[thpb]
\centering
\begin{subfigure}{.30\textwidth}
  \centering
  \includegraphics[width=\linewidth]{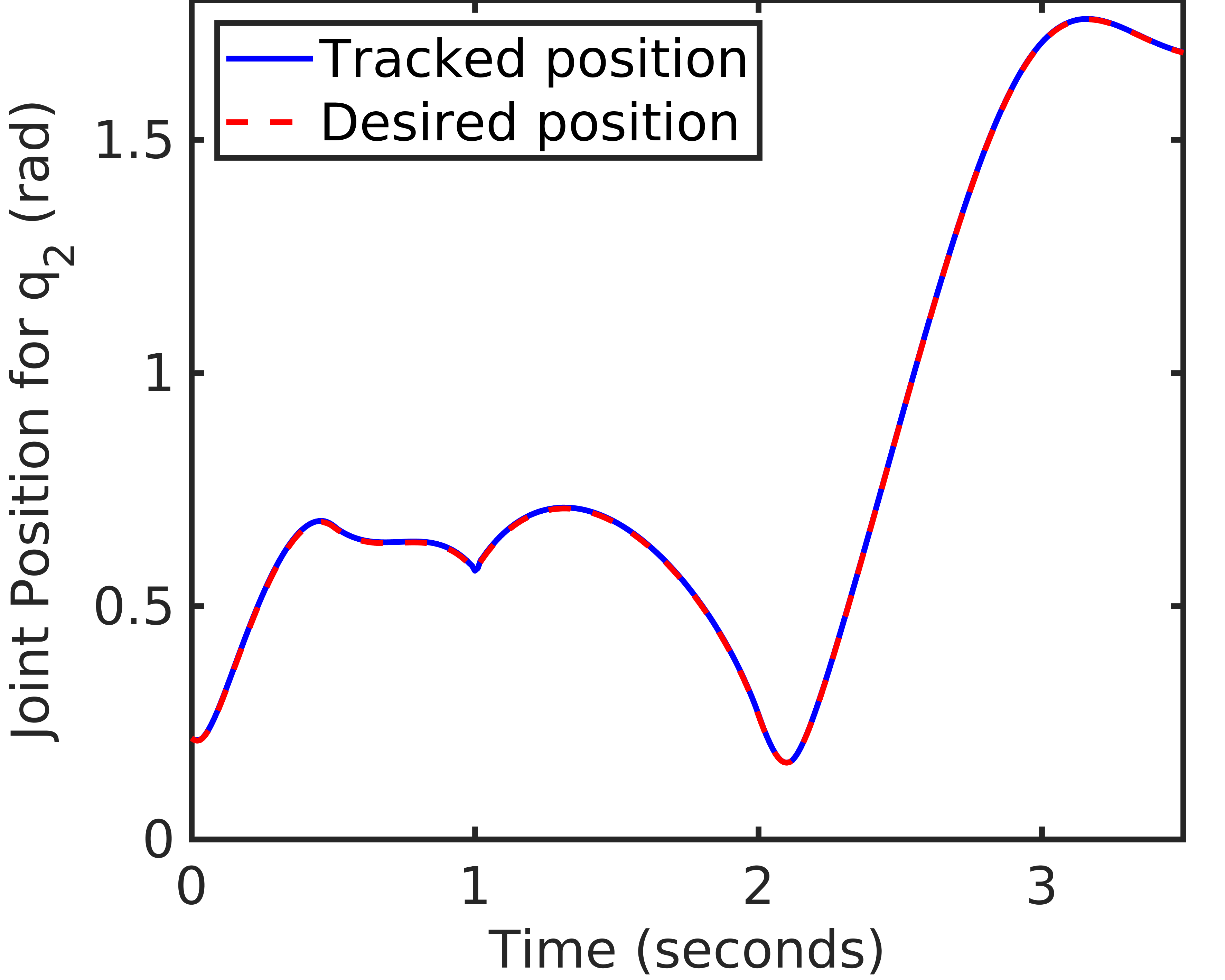}
  \caption{For Case 1}
\label{fig:case1s}
\end{subfigure}
\begin{subfigure}{.30\textwidth}
  \centering
  \includegraphics[width=\linewidth]{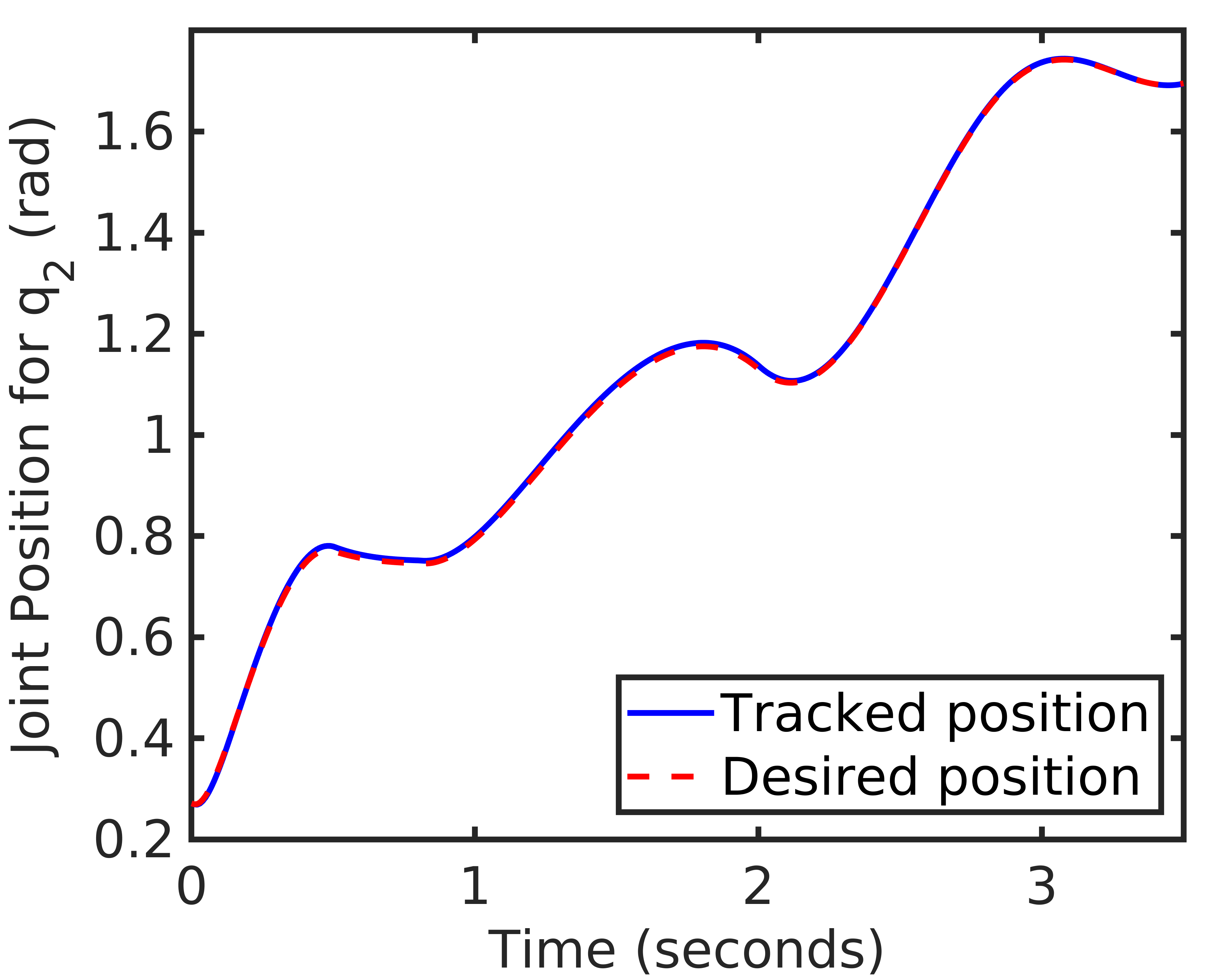}
  \caption{For Case 2}
\label{fig:case2s}
\end{subfigure}
\begin{subfigure}{.30\textwidth}
  \centering
  \includegraphics[width=\linewidth]{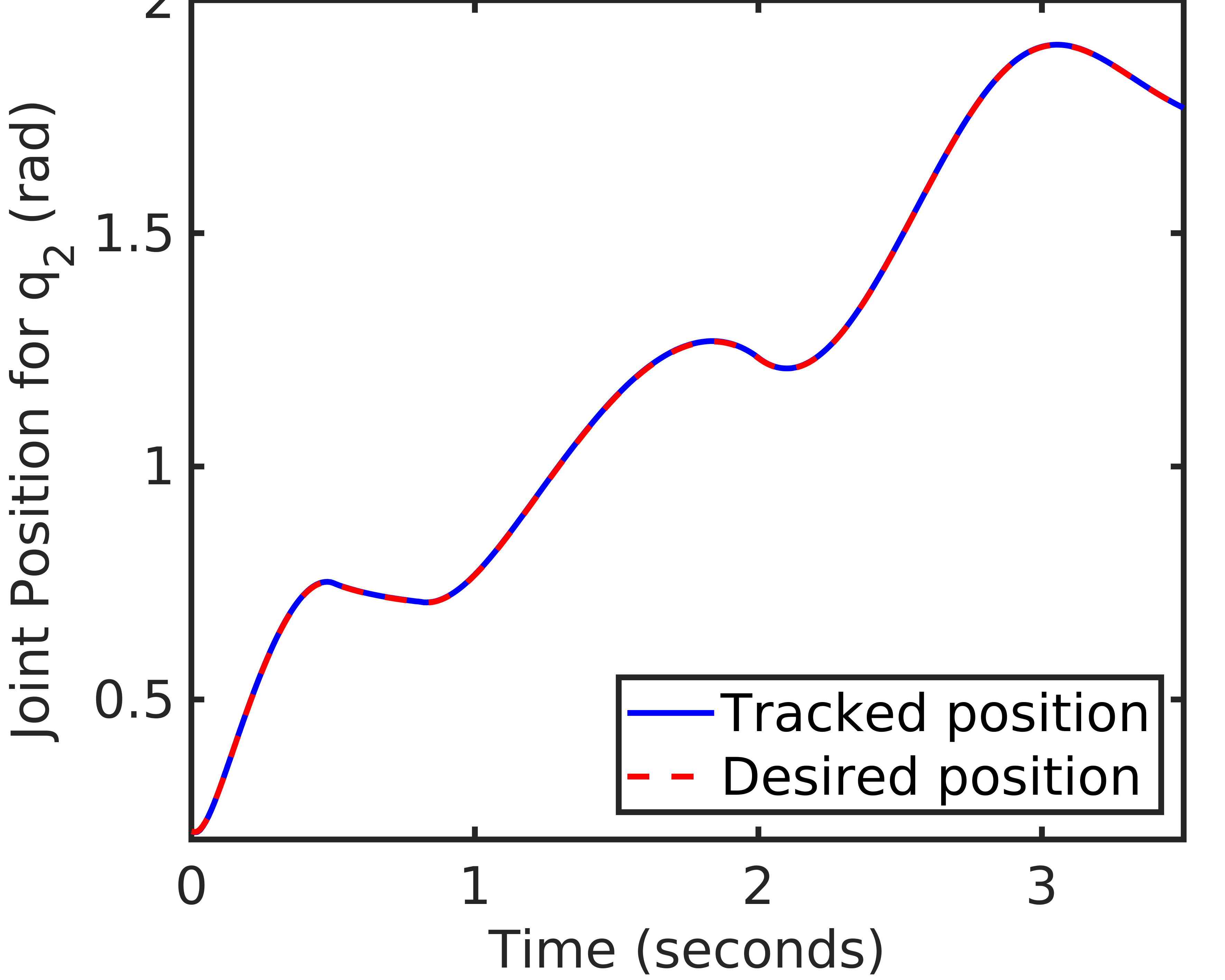}
  \caption{For case 3}
\label{fig:case3s}
\end{subfigure}
\caption{Tracking Performance $q_2$}
\label{fig:q2tp}
\vspace{-2mm}
\end{figure*}

\begin{figure*}[thpb]
\centering
\begin{subfigure}{.30\textwidth}
  \centering
  \includegraphics[width=\linewidth]{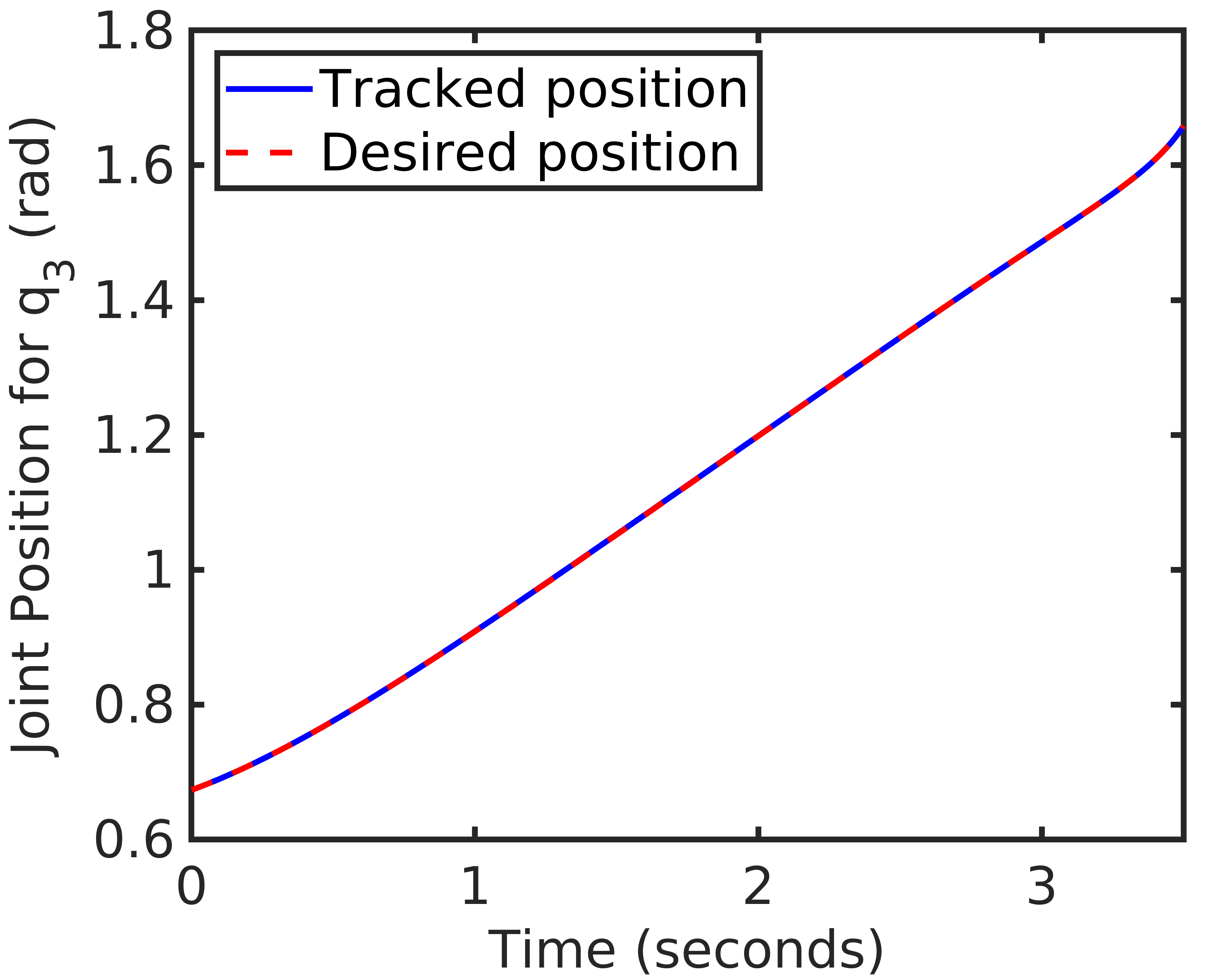}
  \caption{For Case 1}
\label{fig:case1s}
\end{subfigure}
\begin{subfigure}{.30\textwidth}
  \centering
  \includegraphics[width=\linewidth]{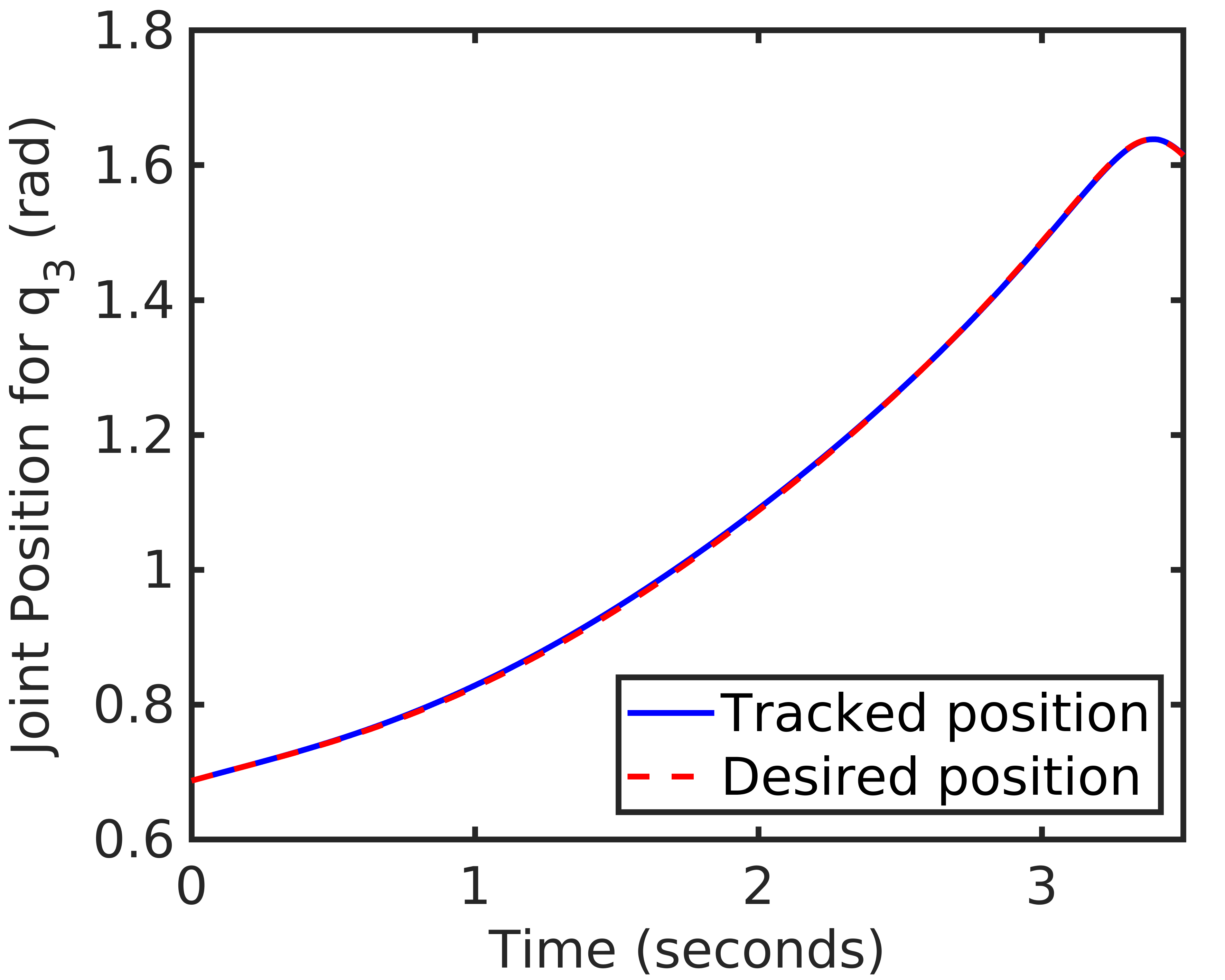}
  \caption{For Case 2}
\label{fig:case2s}
\end{subfigure}
\begin{subfigure}{.30\textwidth}
  \centering
  \includegraphics[width=\linewidth]{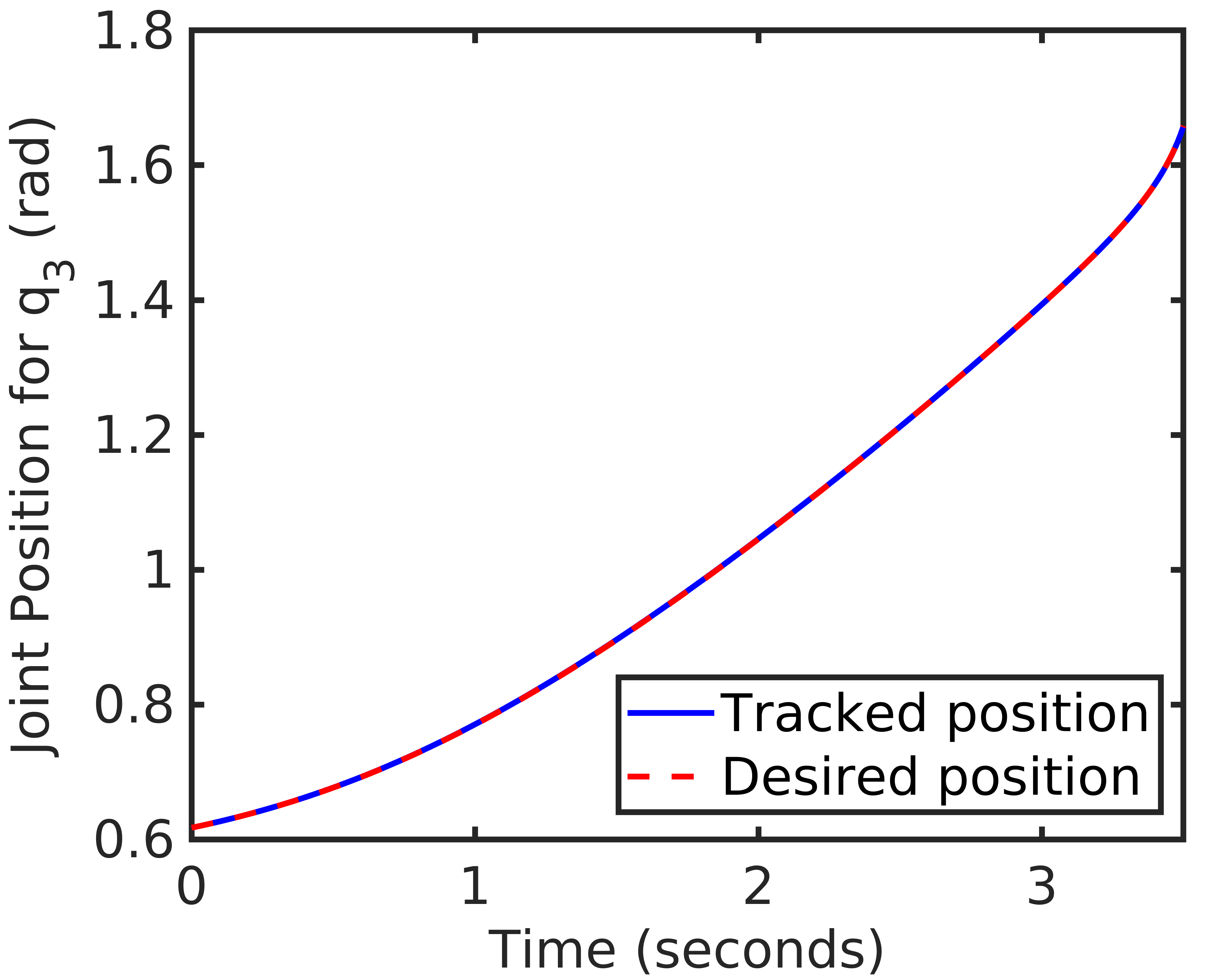}
  \caption{For case 3}
\label{fig:case3s}
\end{subfigure}
\caption{Tracking Performance $q_3$}
\label{fig:q3tp}
\vspace{-2mm}
\end{figure*}

\begin{figure*}[thpb]
\centering
\begin{subfigure}{.30\textwidth}
  \centering
  \includegraphics[width=\linewidth]{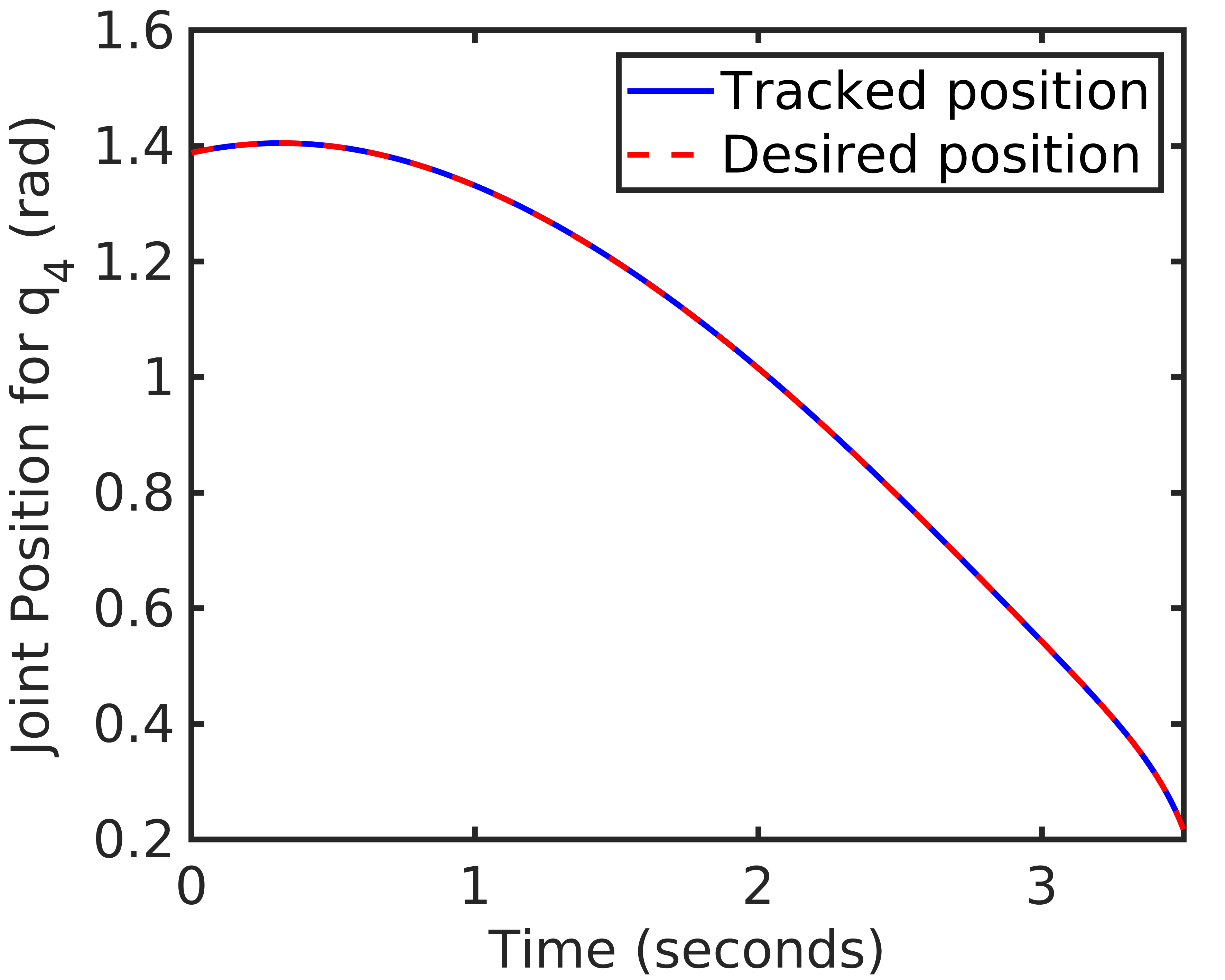}
  \caption{For Case 1}
\label{fig:case1s}
\end{subfigure}
\begin{subfigure}{.30\textwidth}
  \centering
  \includegraphics[width=\linewidth]{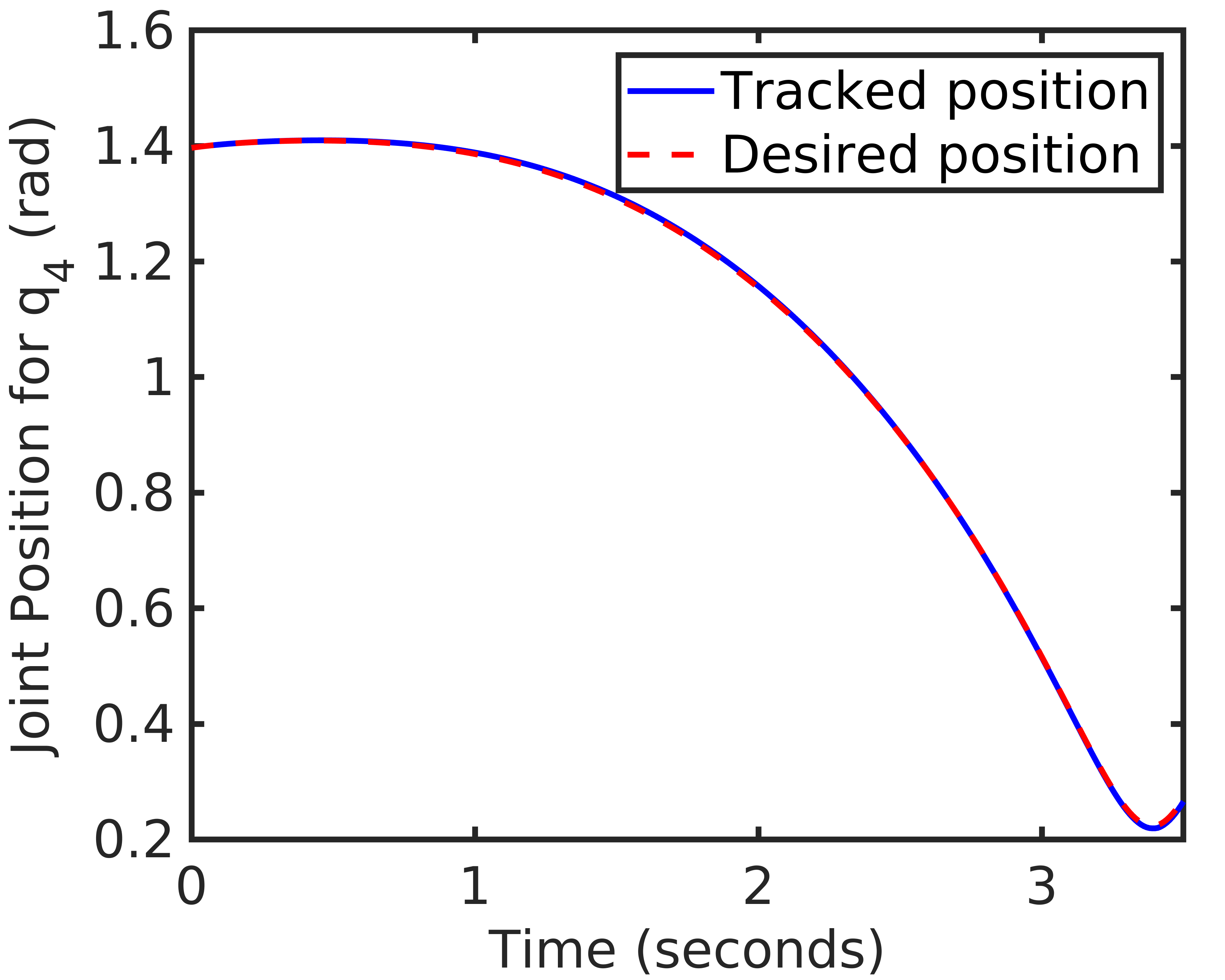}
  \caption{For Case 2}
\label{fig:case2s}
\end{subfigure}
\begin{subfigure}{.30\textwidth}
  \centering
  \includegraphics[width=\linewidth]{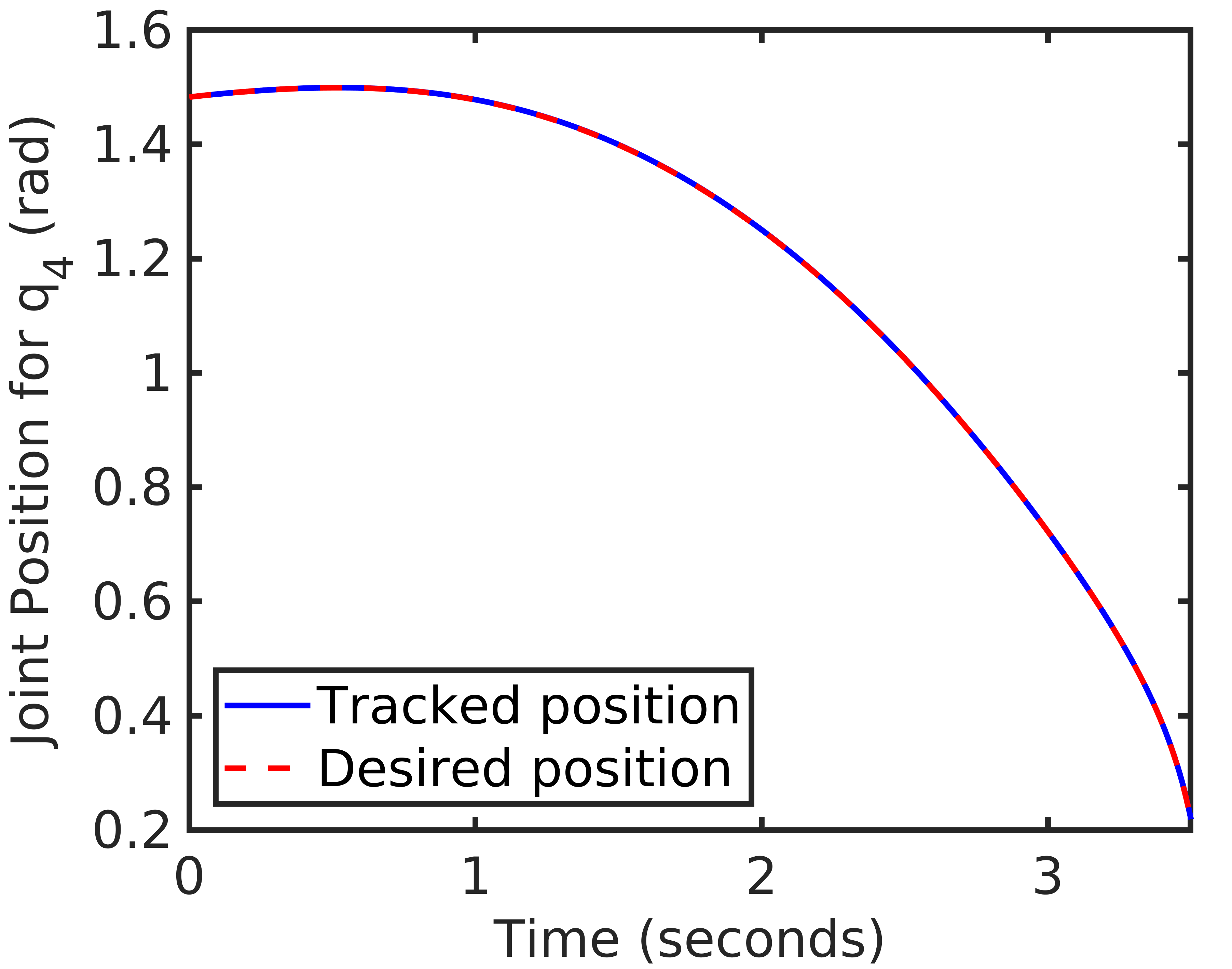}
  \caption{For case 3}
\label{fig:case3s}
\end{subfigure}
\caption{Tracking Performance $q_4$}
\label{fig:q4tp}
\vspace{-2mm}
\end{figure*}

\begin{figure*}[thpb]
\centering
\begin{subfigure}{.30\textwidth}
  \centering
  \includegraphics[width=\linewidth]{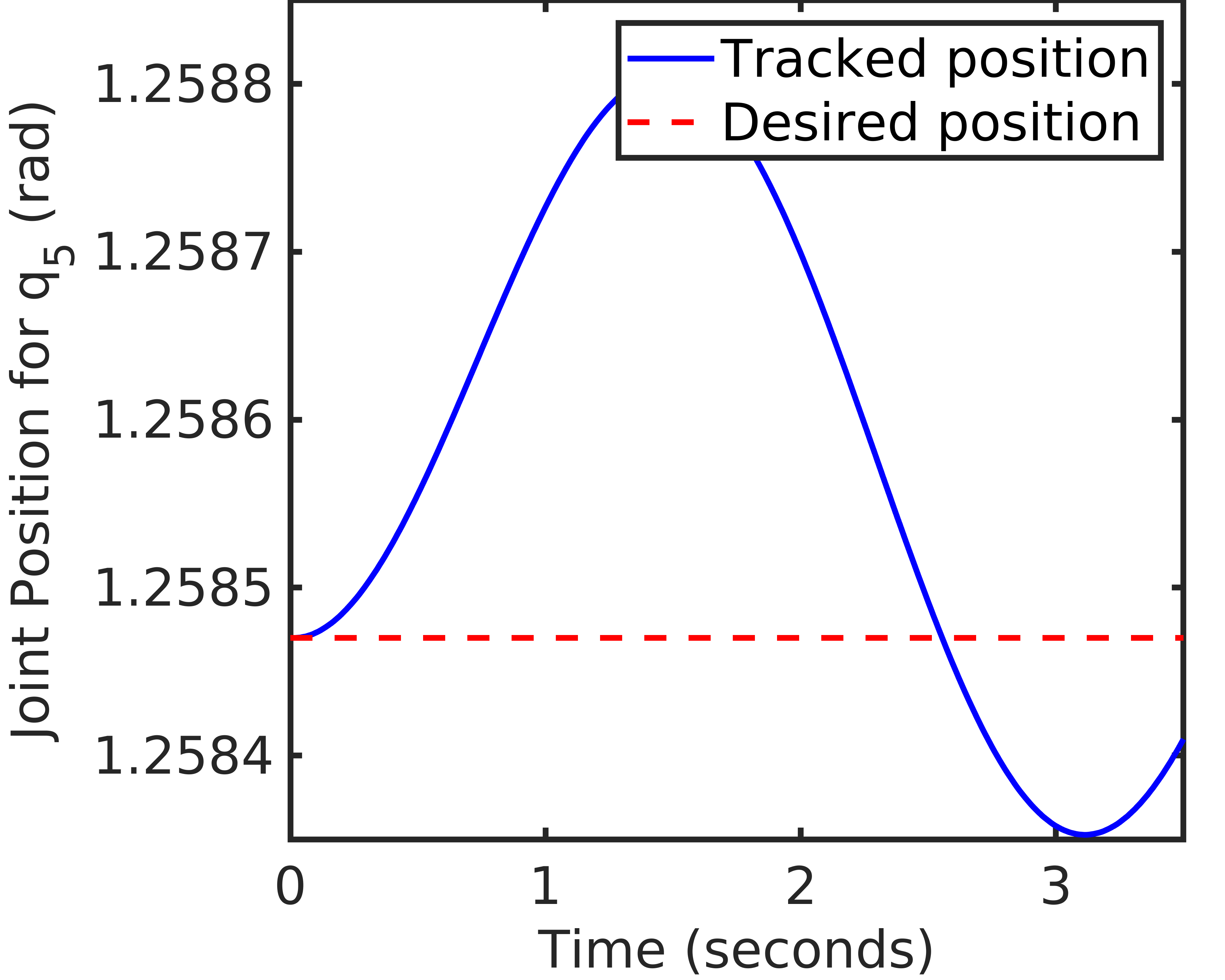}
  \caption{For Case 1}
\label{fig:case1s}
\end{subfigure}
\begin{subfigure}{.30\textwidth}
  \centering
  \includegraphics[width=\linewidth]{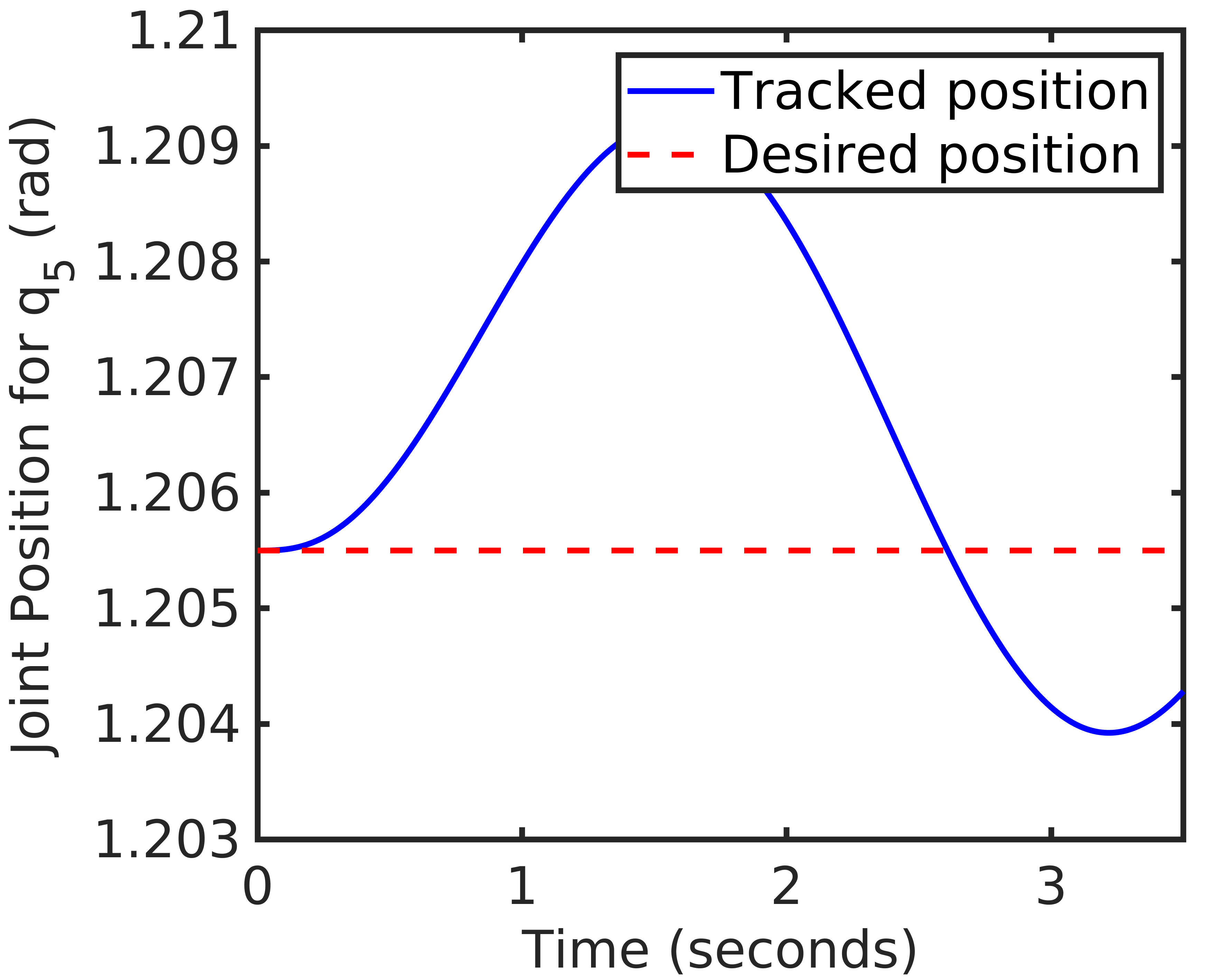}
  \caption{For Case 2}
\label{fig:case2s}
\end{subfigure}
\begin{subfigure}{.30\textwidth}
  \centering
  \includegraphics[width=\linewidth]{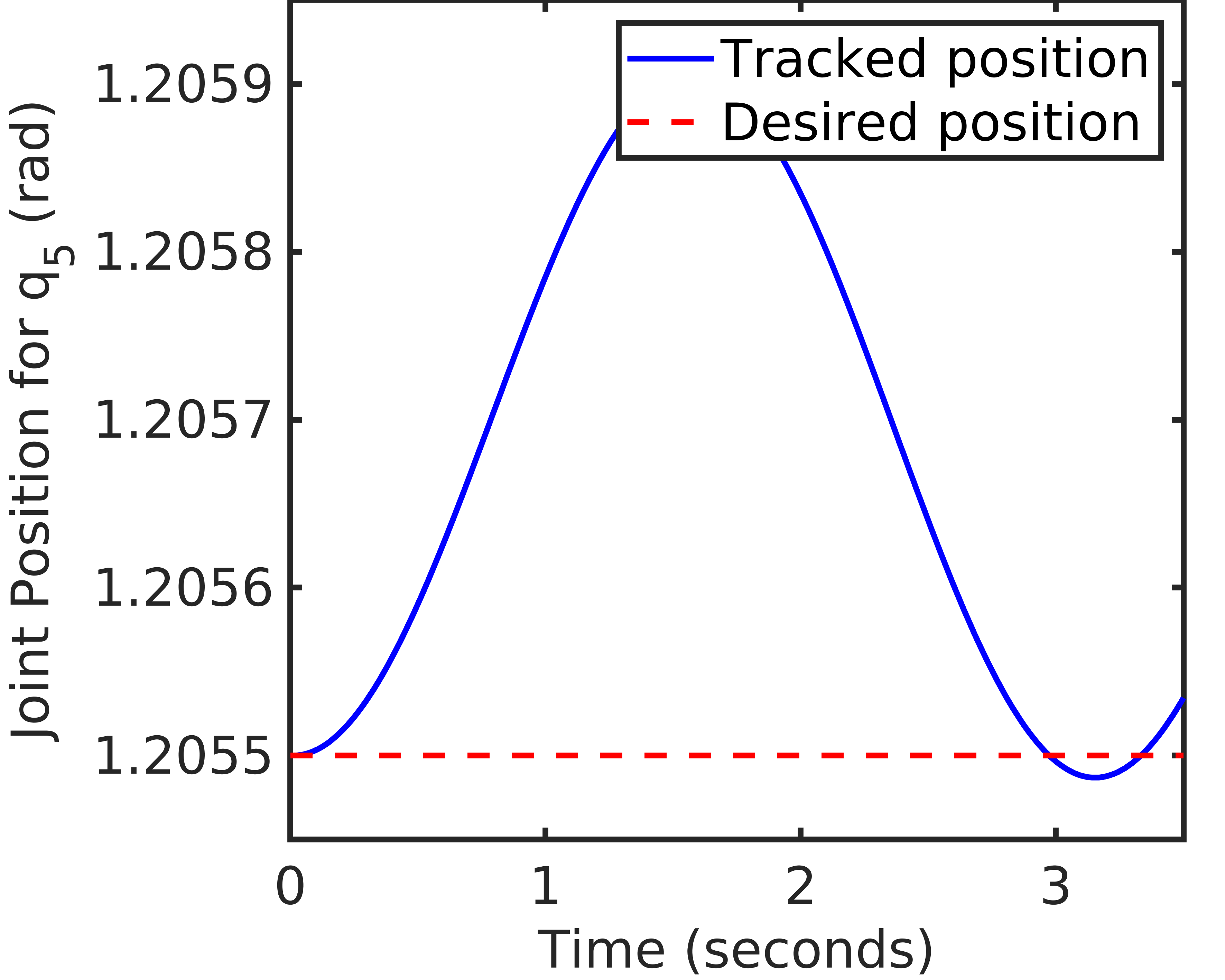}
  \caption{For case 3}
\label{fig:case3s}
\end{subfigure}
\caption{Tracking Performance $q_5$}
\label{fig:q5tp}
\vspace{-2mm}
\end{figure*}

\begin{figure*}[thpb]
\centering
\begin{subfigure}{.30\textwidth}
  \centering
  \includegraphics[width=\linewidth]{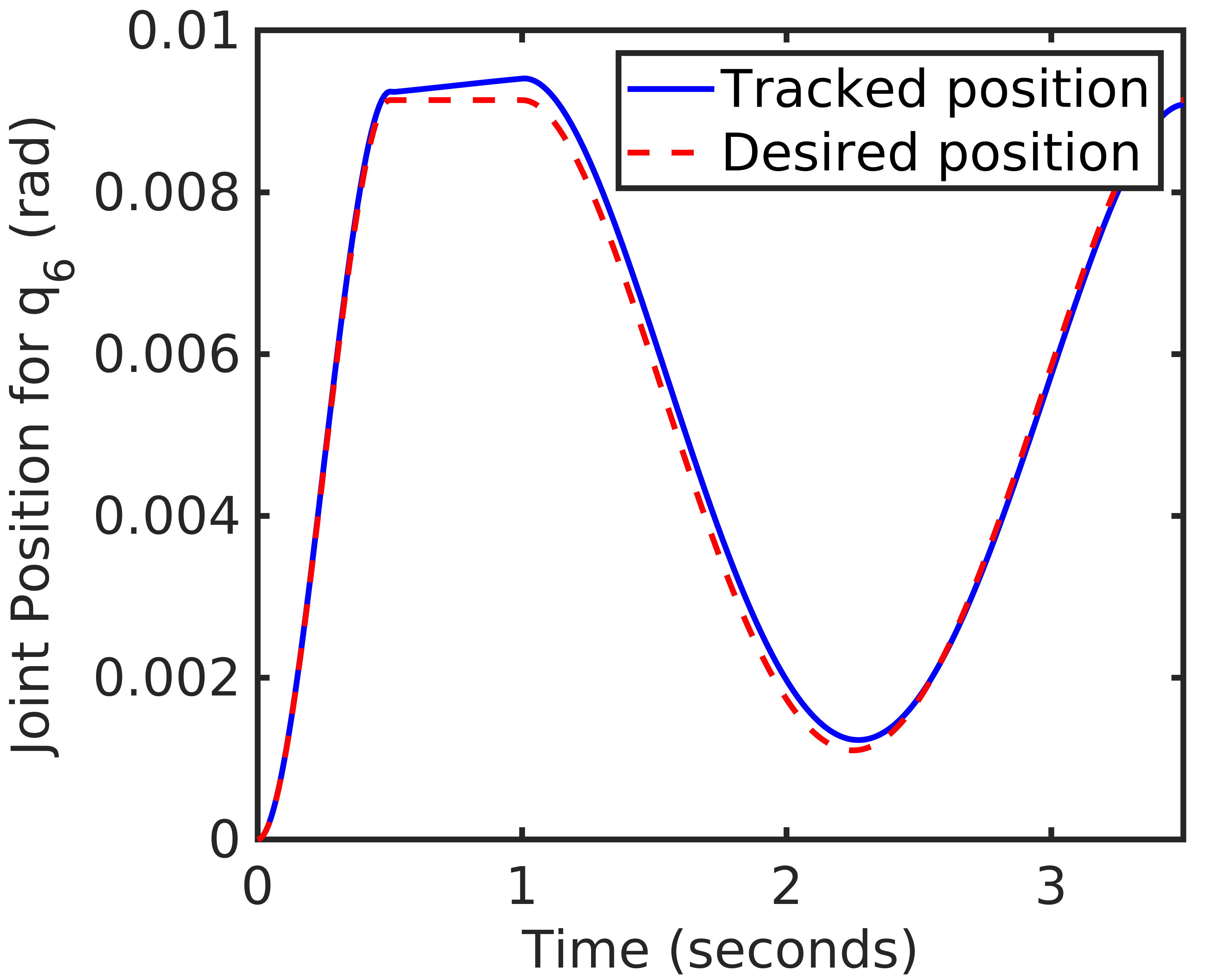}
  \caption{For Case 1}
\label{fig:case1s}
\end{subfigure}
\begin{subfigure}{.30\textwidth}
  \centering
  \includegraphics[width=\linewidth]{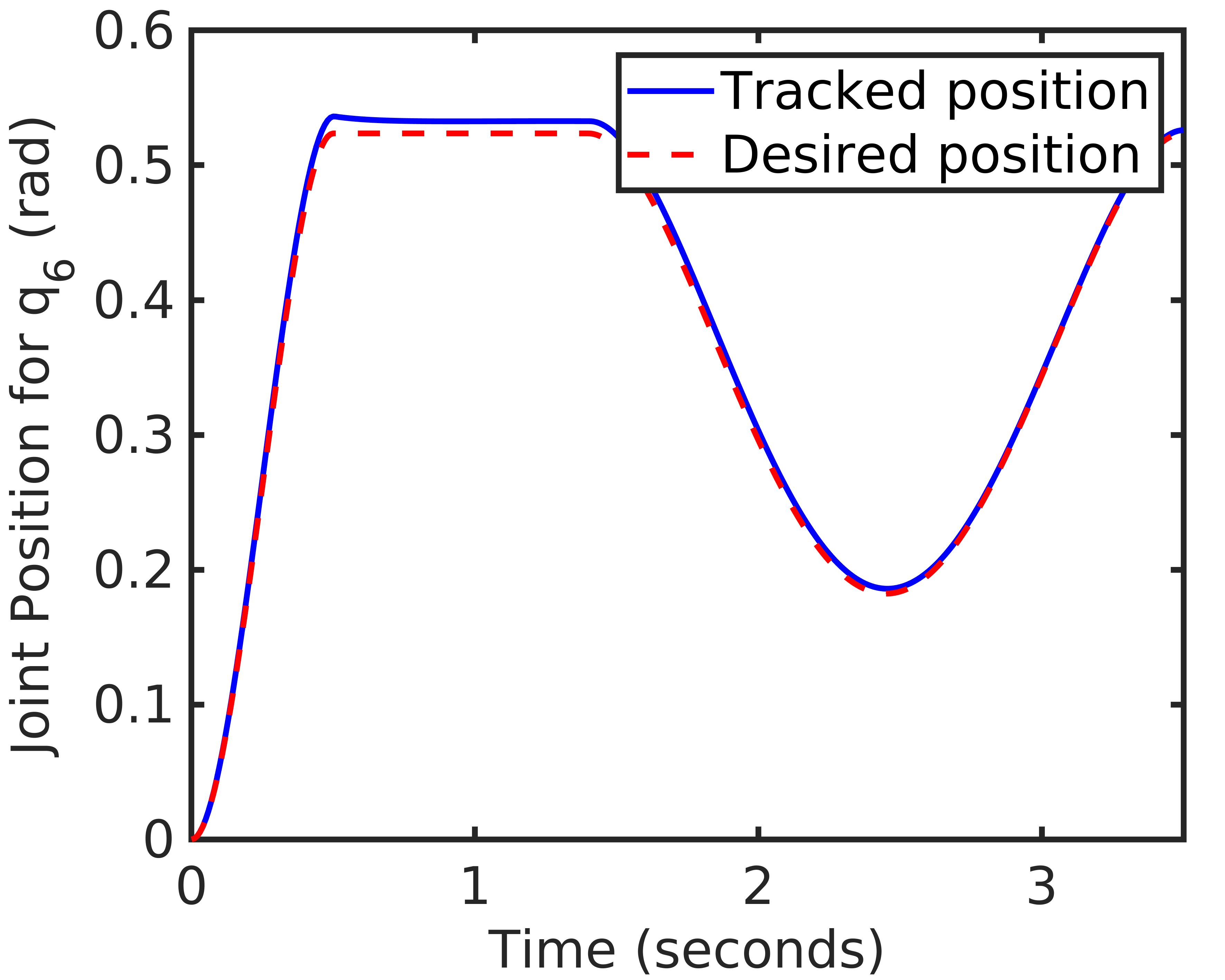}
  \caption{For Case 2}
\label{fig:case2s}
\end{subfigure}
\begin{subfigure}{.30\textwidth}
  \centering
  \includegraphics[width=\linewidth]{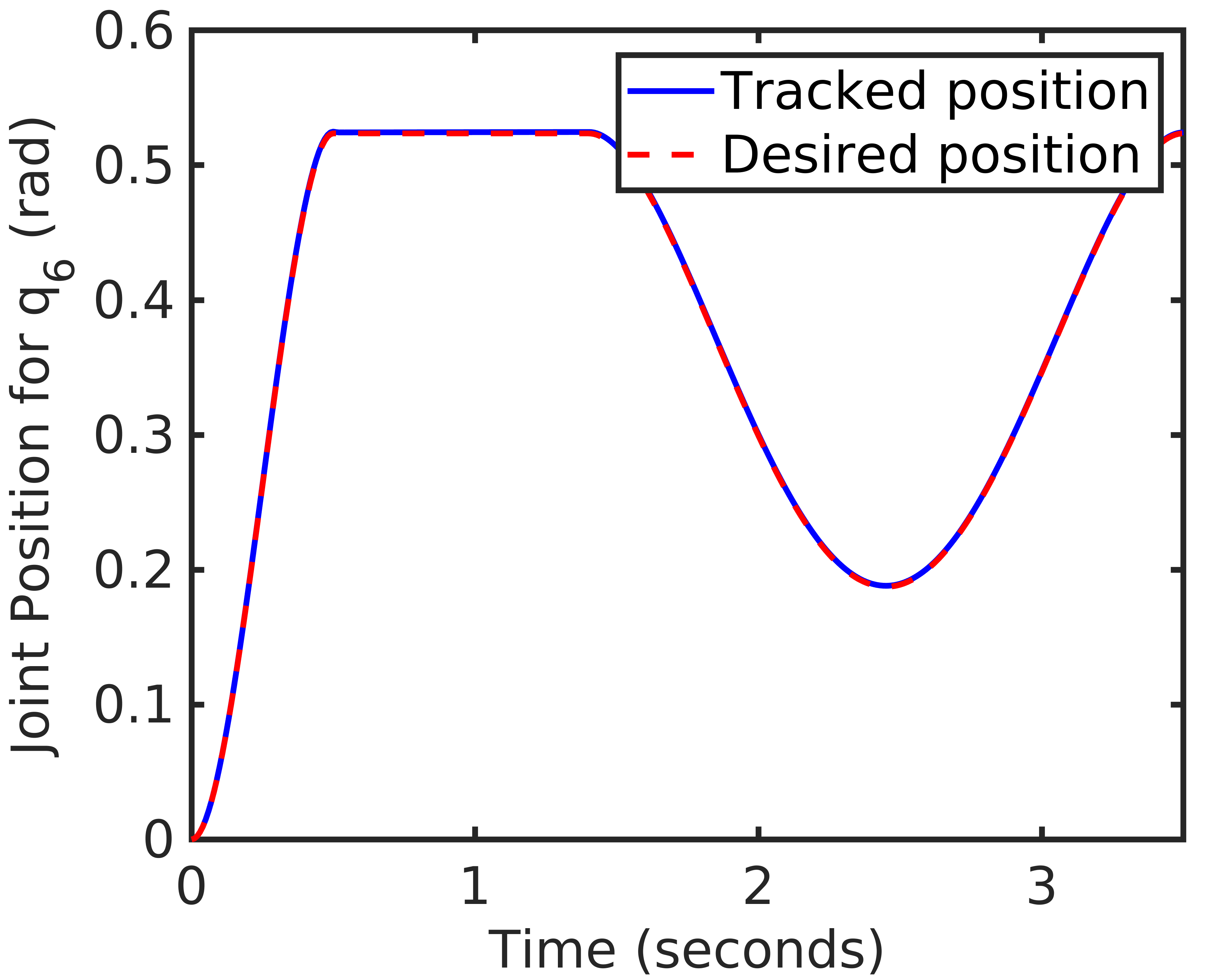}
  \caption{For case 3}
\label{fig:case3s}
\end{subfigure}
\caption{Tracking Performance $q_6$}
\label{fig:q6tp}
\vspace{-2mm}
\end{figure*}

\begin{figure*}[thpb]
\centering
\begin{subfigure}{.30\textwidth}
  \centering
  \includegraphics[width=\linewidth]{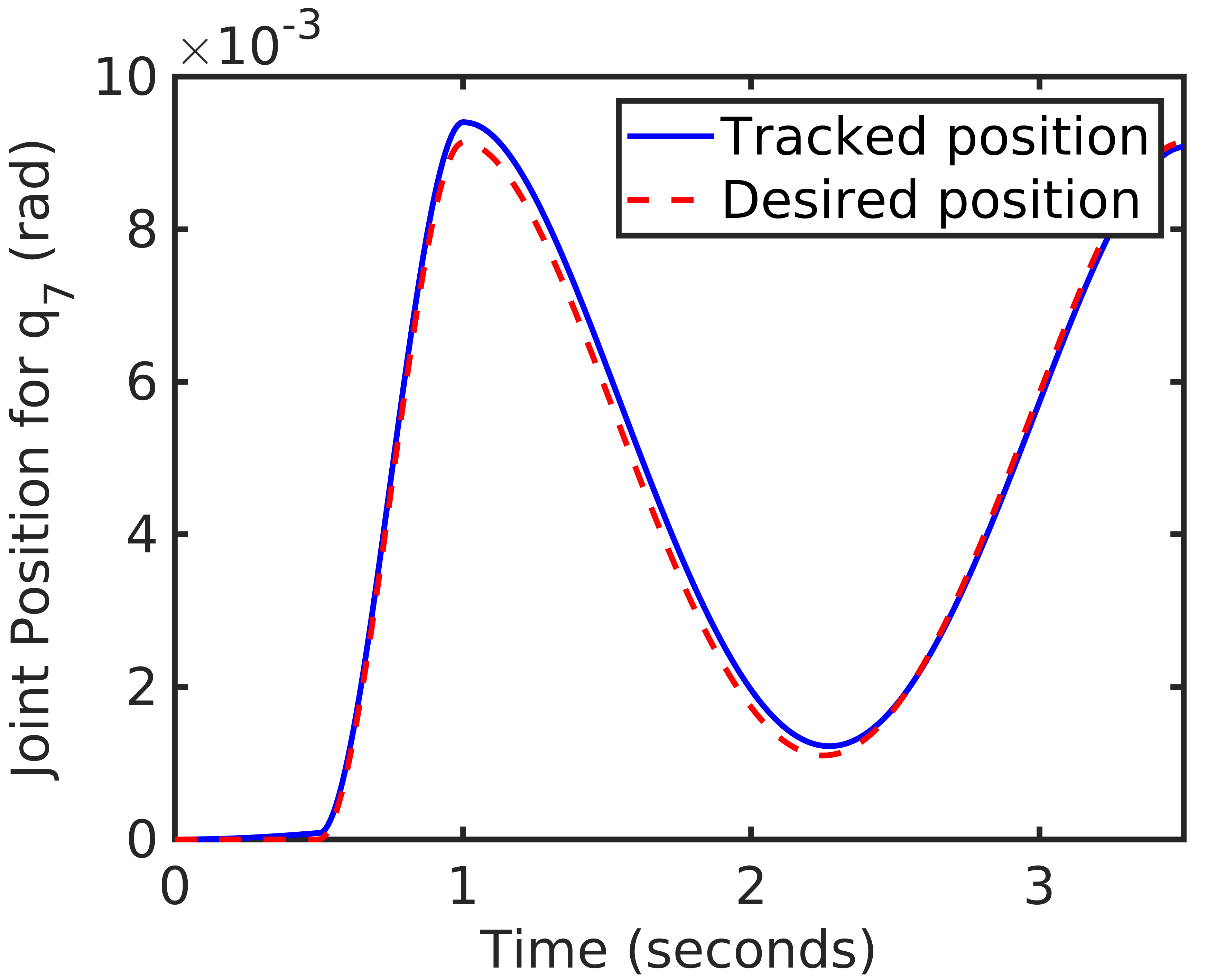}
  \caption{For Case 1}
\label{fig:case1s}
\end{subfigure}
\begin{subfigure}{.30\textwidth}
  \centering
  \includegraphics[width=\linewidth]{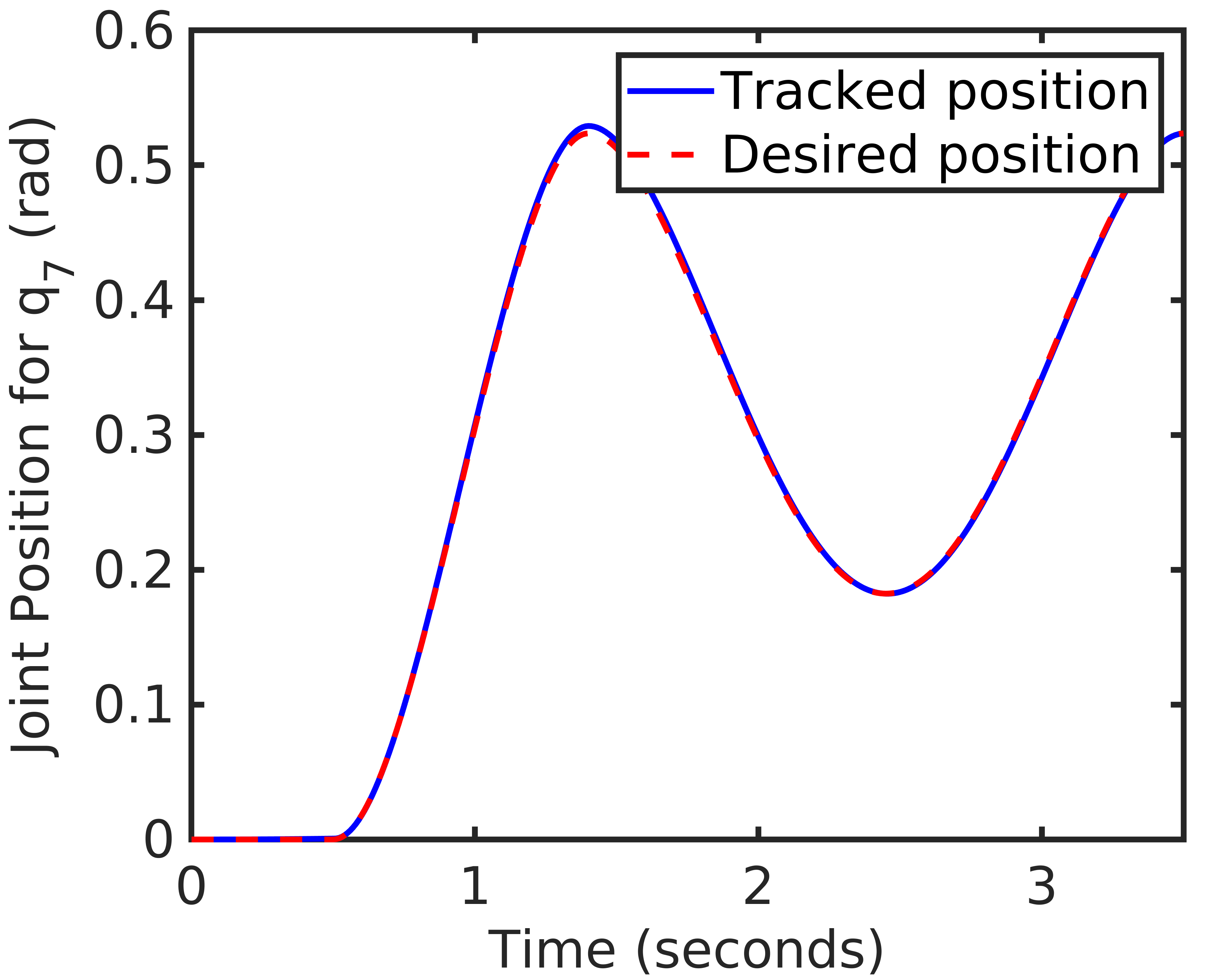}
  \caption{For Case 2}
\label{fig:case2s}
\end{subfigure}
\begin{subfigure}{.30\textwidth}
  \centering
  \includegraphics[width=\linewidth]{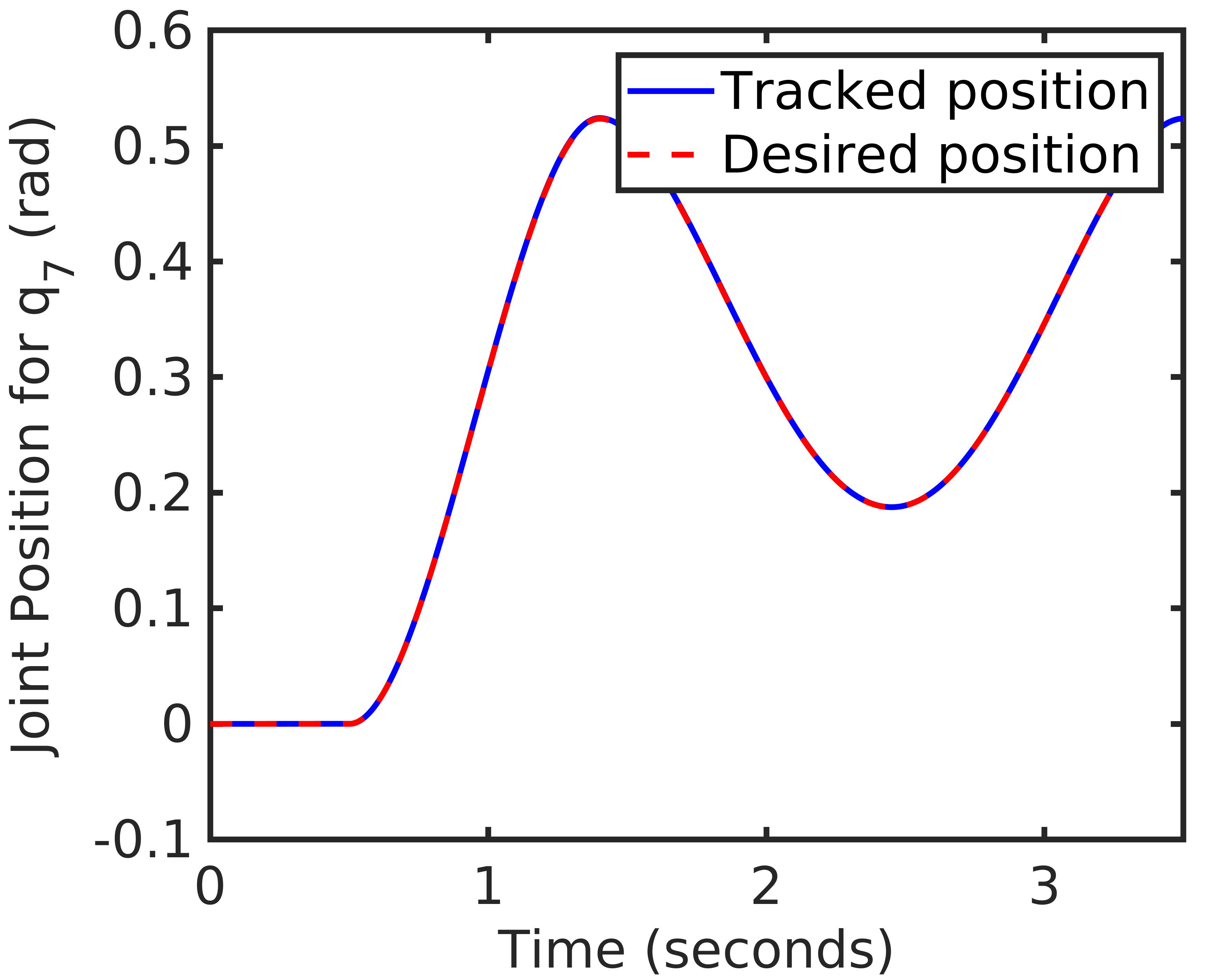}
  \caption{For case 3}
\label{fig:case3s}
\end{subfigure}
\caption{Tracking Performance $q_7$}
\label{fig:q7tp}
\vspace{-2mm}
\end{figure*}

\begin{figure*}[thpb]
\centering
\begin{subfigure}{.30\textwidth}
  \centering
  \includegraphics[width=\linewidth]{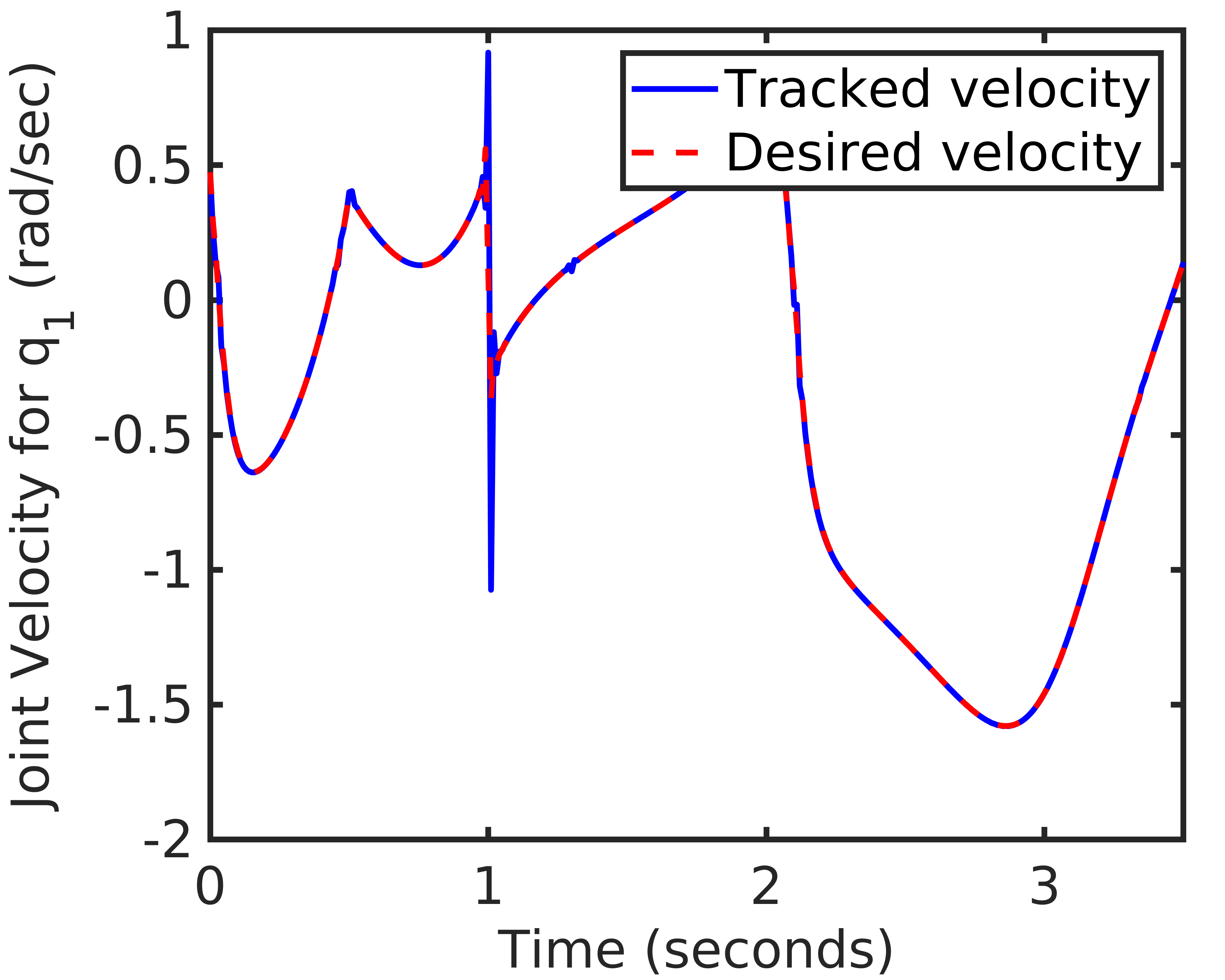}
  \caption{For Case 1}
\label{fig:case1s}
\end{subfigure}
\begin{subfigure}{.30\textwidth}
  \centering
  \includegraphics[width=\linewidth]{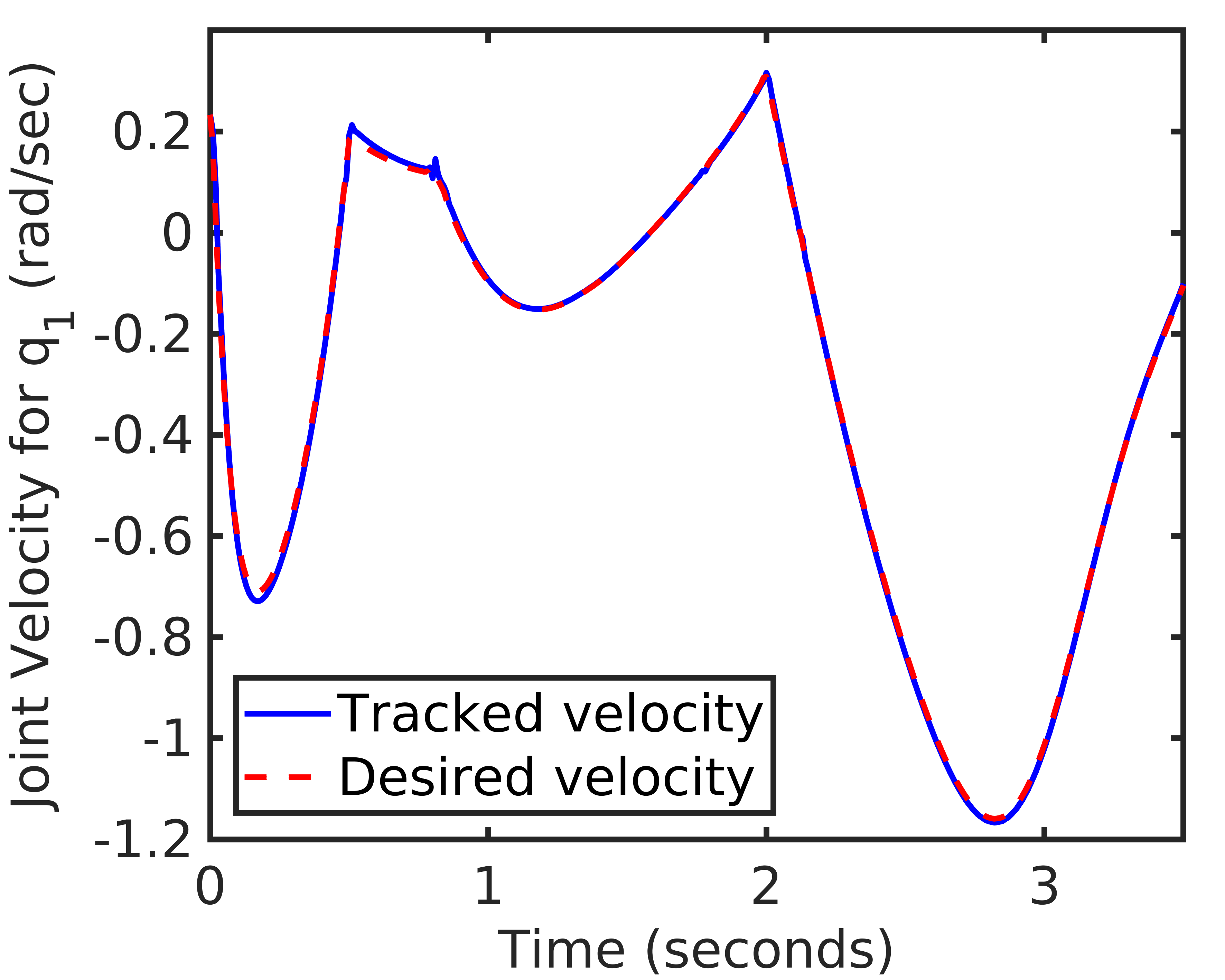}
  \caption{For Case 2}
\label{fig:case2s}
\end{subfigure}
\begin{subfigure}{.30\textwidth}
  \centering
  \includegraphics[width=\linewidth]{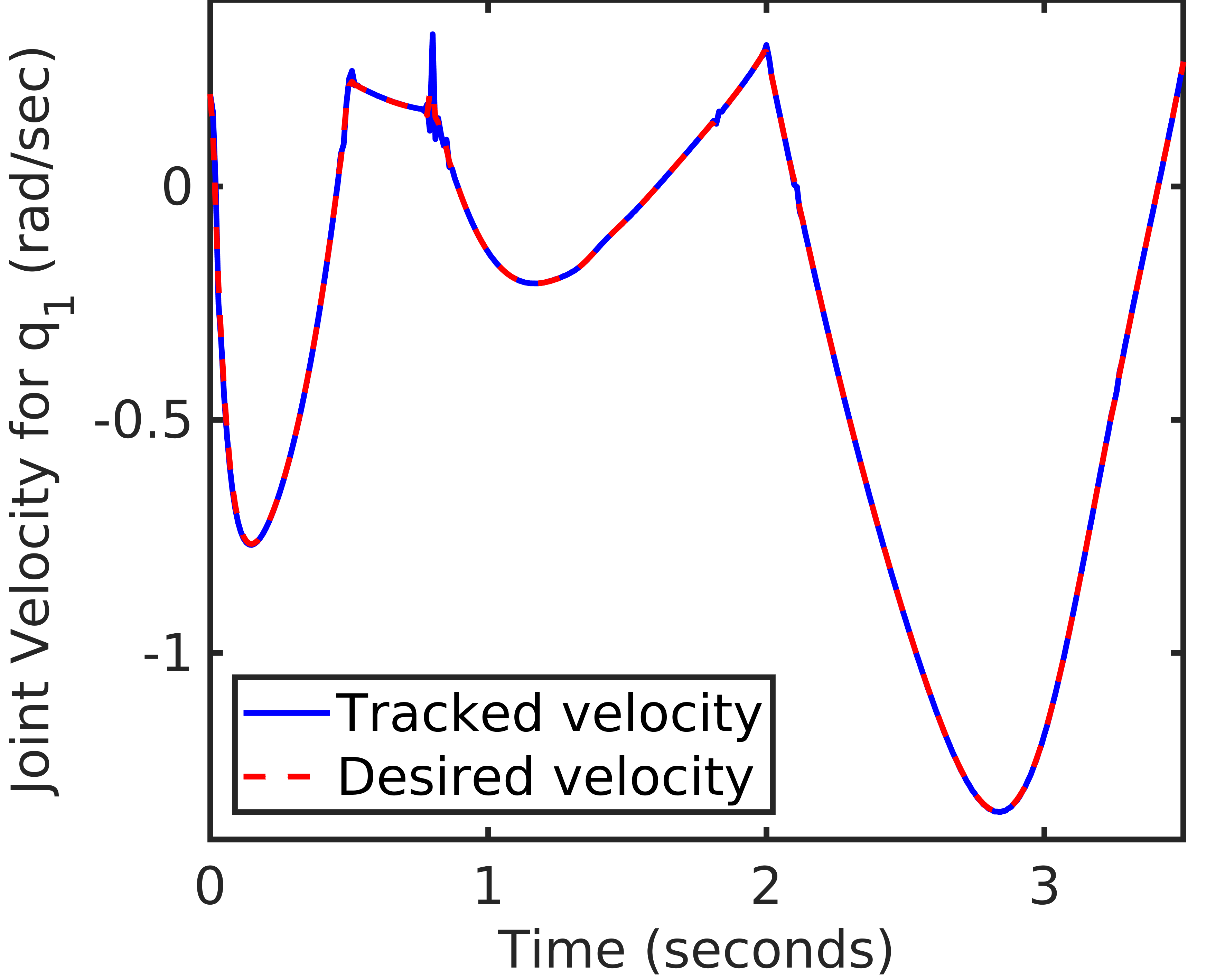}
  \caption{For case 3}
\label{fig:case3s}
\end{subfigure}
\caption{Tracking Performance $\dot{q}_1$}
\label{fig:dq1tp}
\vspace{-2mm}
\end{figure*}

\begin{figure*}[thpb]
\centering
\begin{subfigure}{.30\textwidth}
  \centering
  \includegraphics[width=\linewidth]{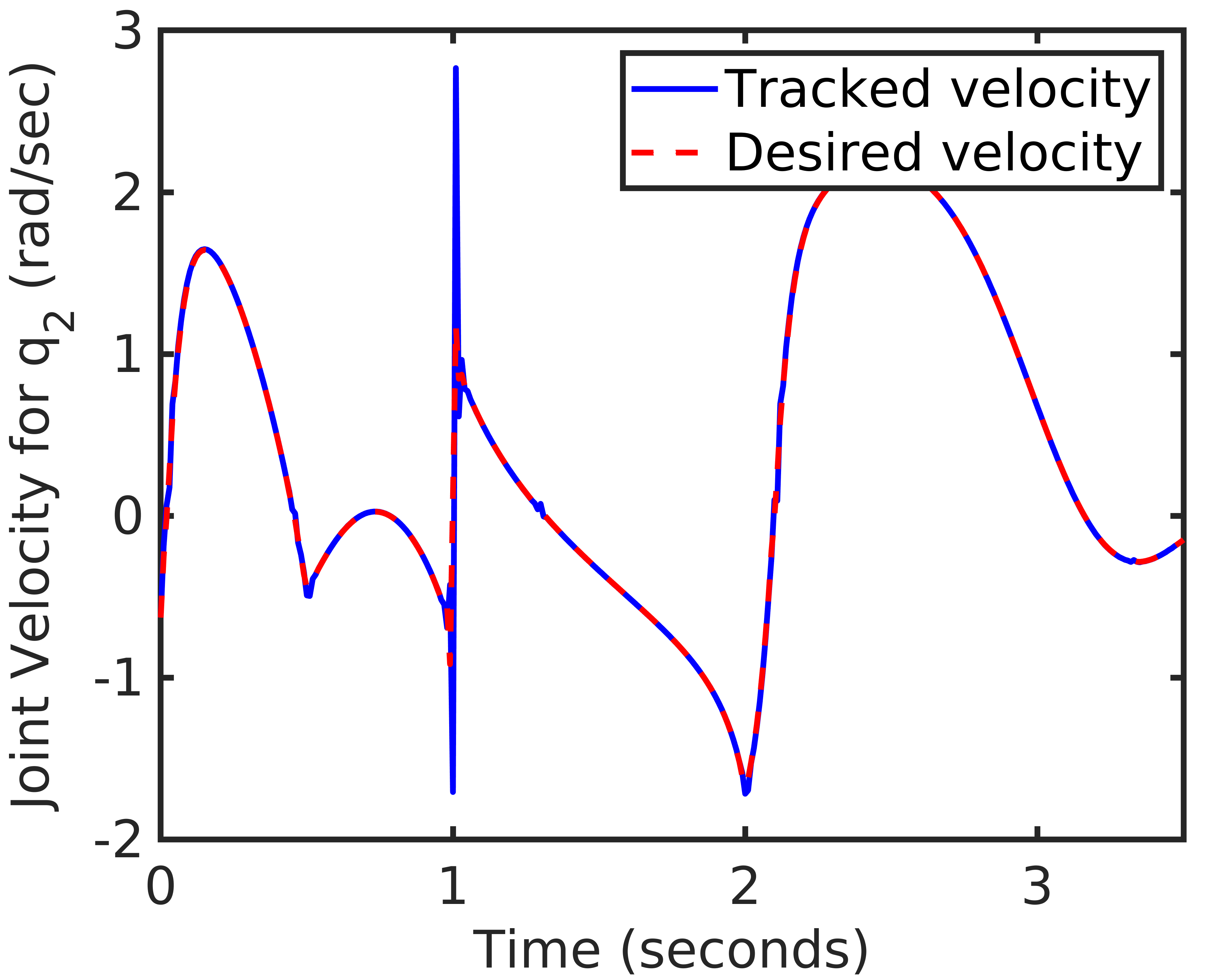}
  \caption{For Case 1}
\label{fig:case1s}
\end{subfigure}
\begin{subfigure}{.30\textwidth}
  \centering
  \includegraphics[width=\linewidth]{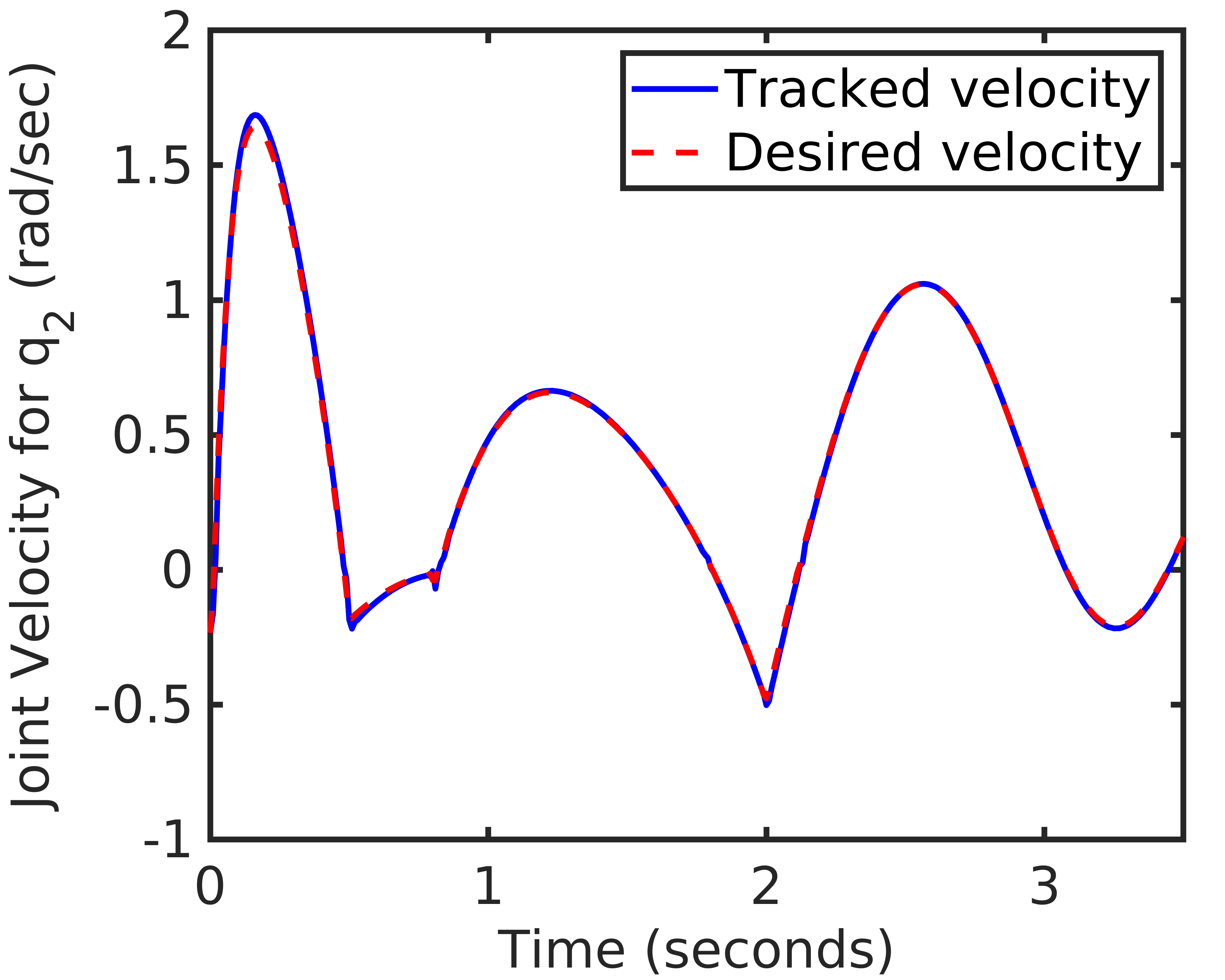}
  \caption{For Case 2}
\label{fig:case2s}
\end{subfigure}
\begin{subfigure}{.30\textwidth}
  \centering
  \includegraphics[width=\linewidth]{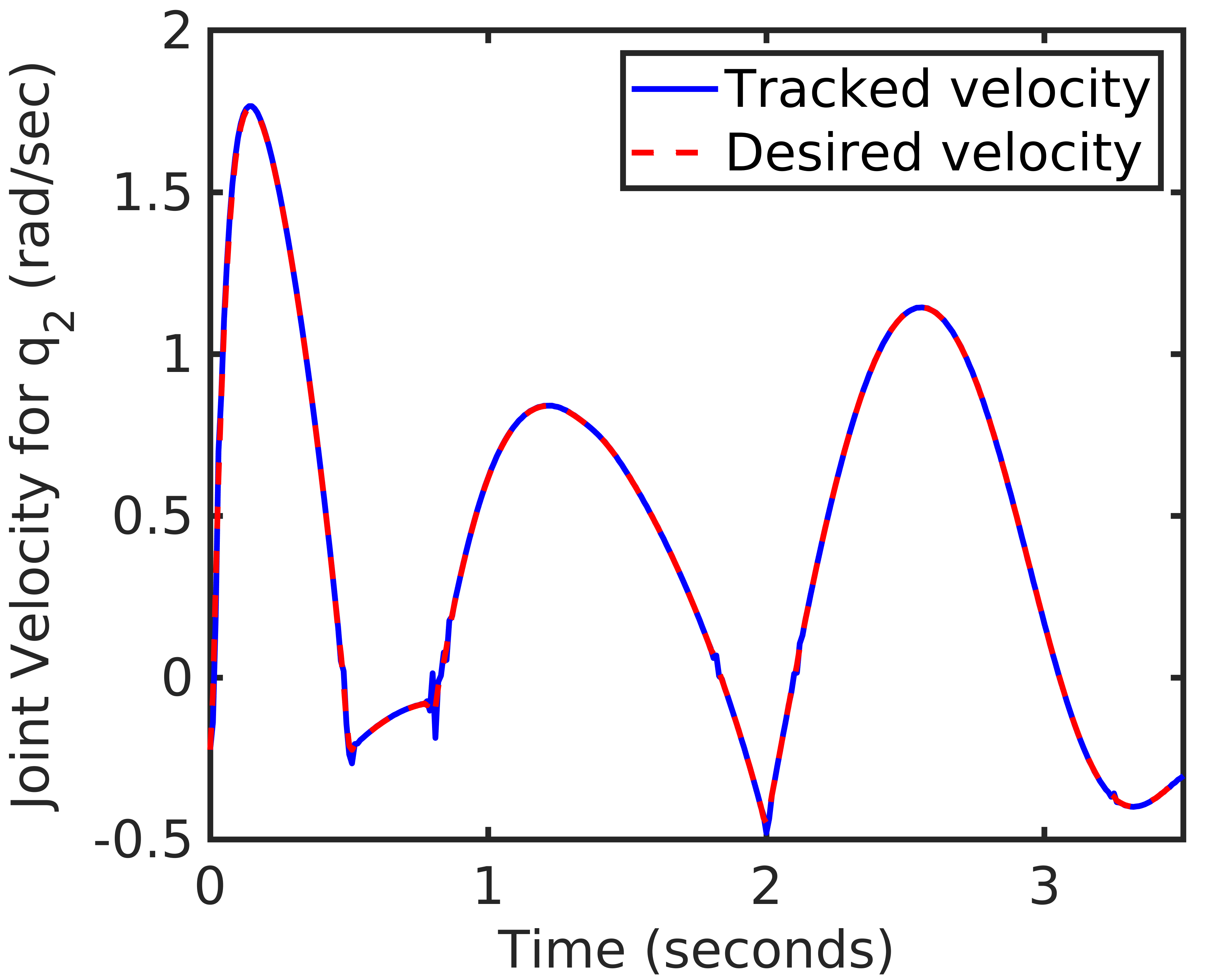}
  \caption{For case 3}
\label{fig:case3s}
\end{subfigure}
\caption{Tracking Performance $\dot{q}_2$}
\label{fig:dq2tp}
\vspace{-2mm}
\end{figure*}

\begin{figure*}[thpb]
\centering
\begin{subfigure}{.30\textwidth}
  \centering
  \includegraphics[width=\linewidth]{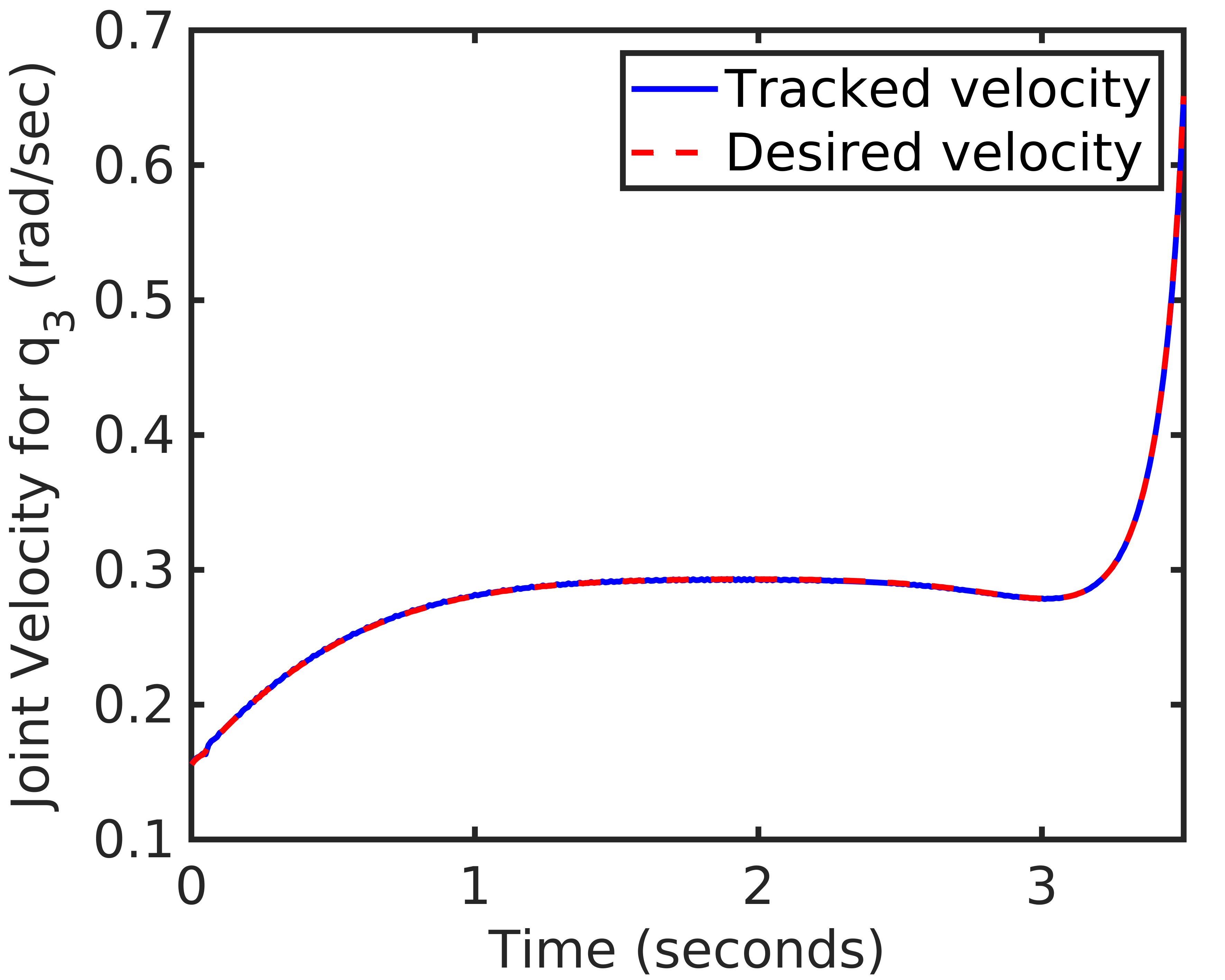}
  \caption{For Case 1}
\label{fig:case1s}
\end{subfigure}
\begin{subfigure}{.30\textwidth}
  \centering
  \includegraphics[width=\linewidth]{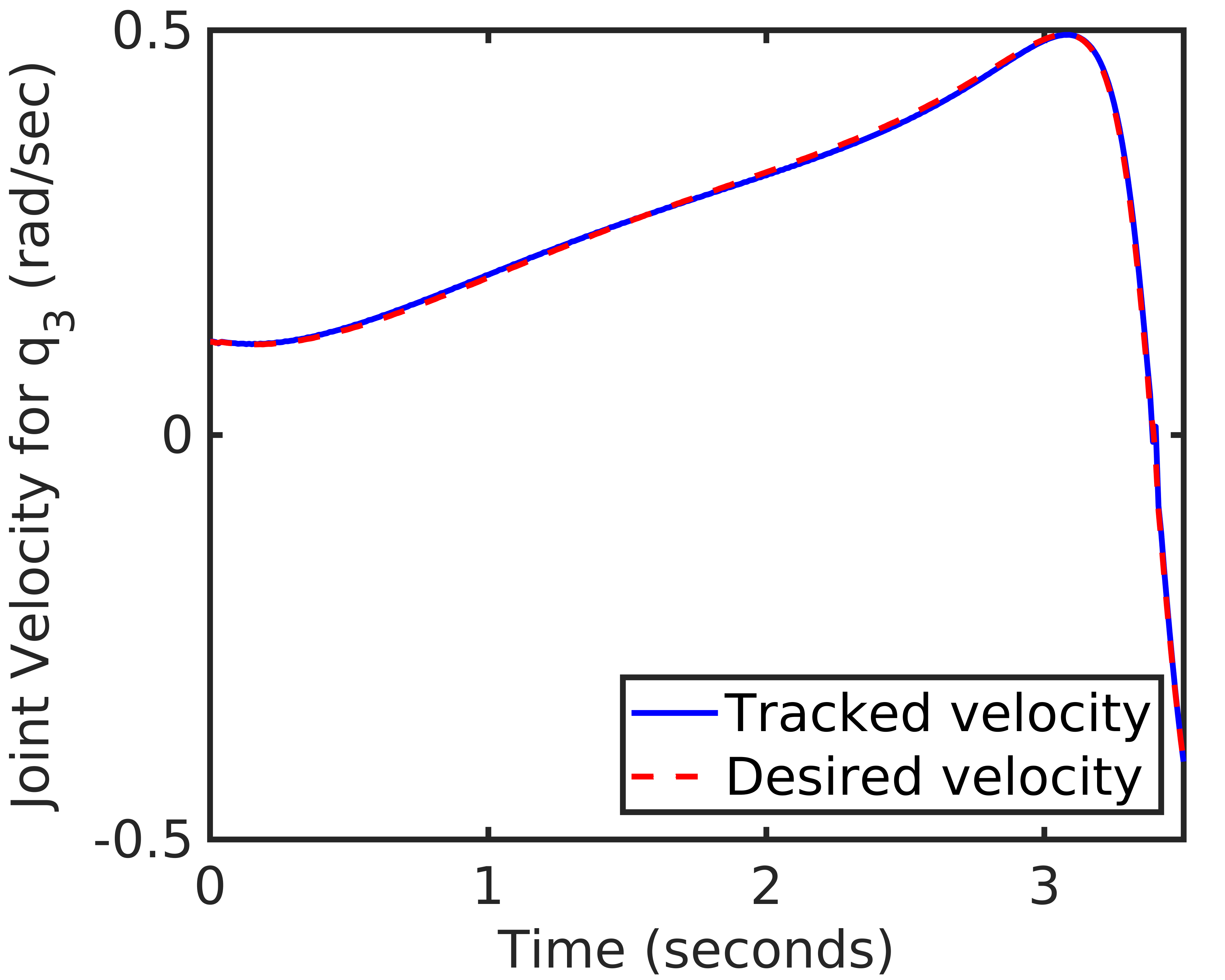}
  \caption{For Case 2}
\label{fig:case2s}
\end{subfigure}
\begin{subfigure}{.30\textwidth}
  \centering
  \includegraphics[width=\linewidth]{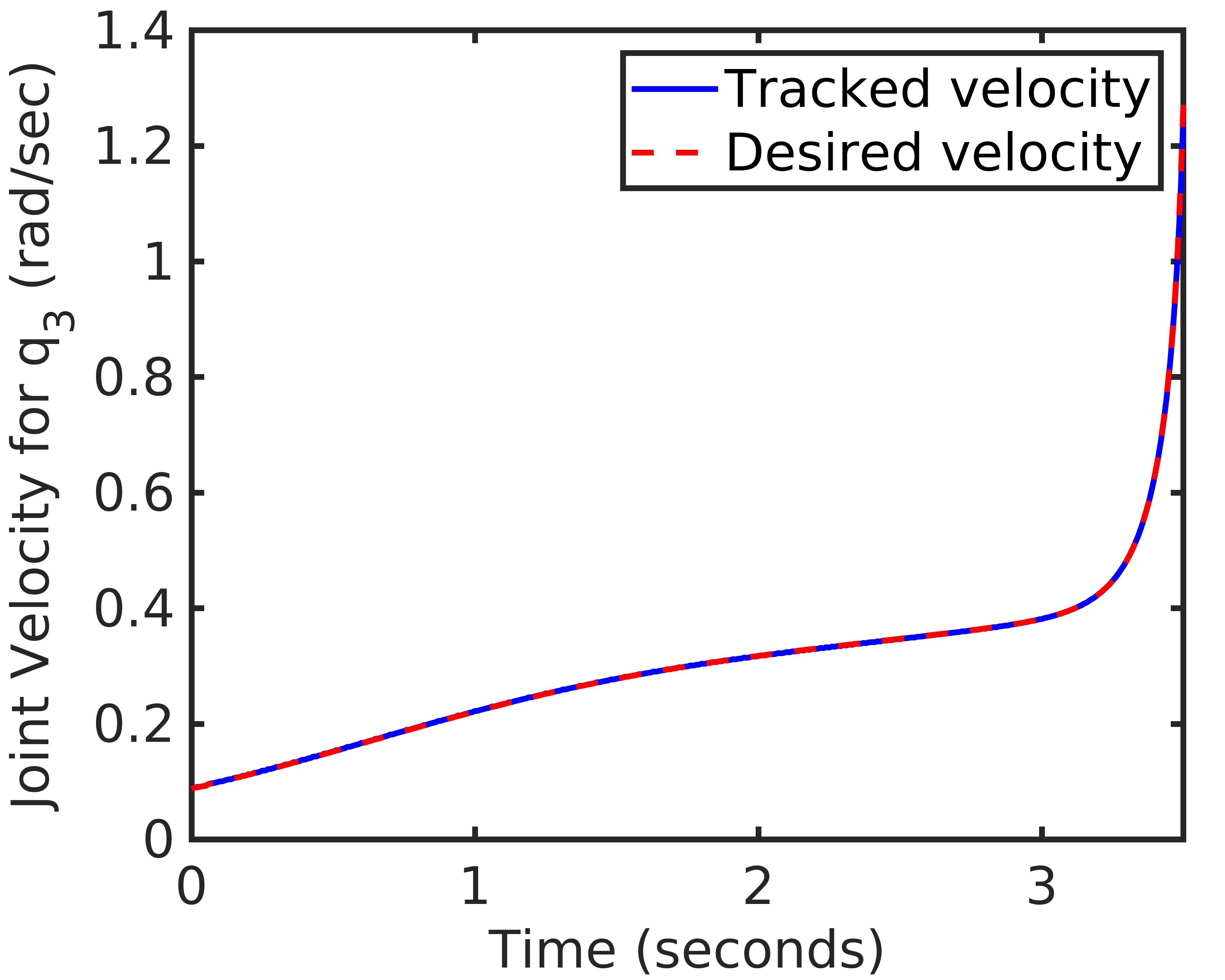}
  \caption{For case 3}
\label{fig:case3s}
\end{subfigure}
\caption{Tracking Performance $\dot{q}_3$}
\label{fig:dq3tp}
\vspace{-2mm}
\end{figure*}

\begin{figure*}[thpb]
\centering
\begin{subfigure}{.30\textwidth}
  \centering
  \includegraphics[width=\linewidth]{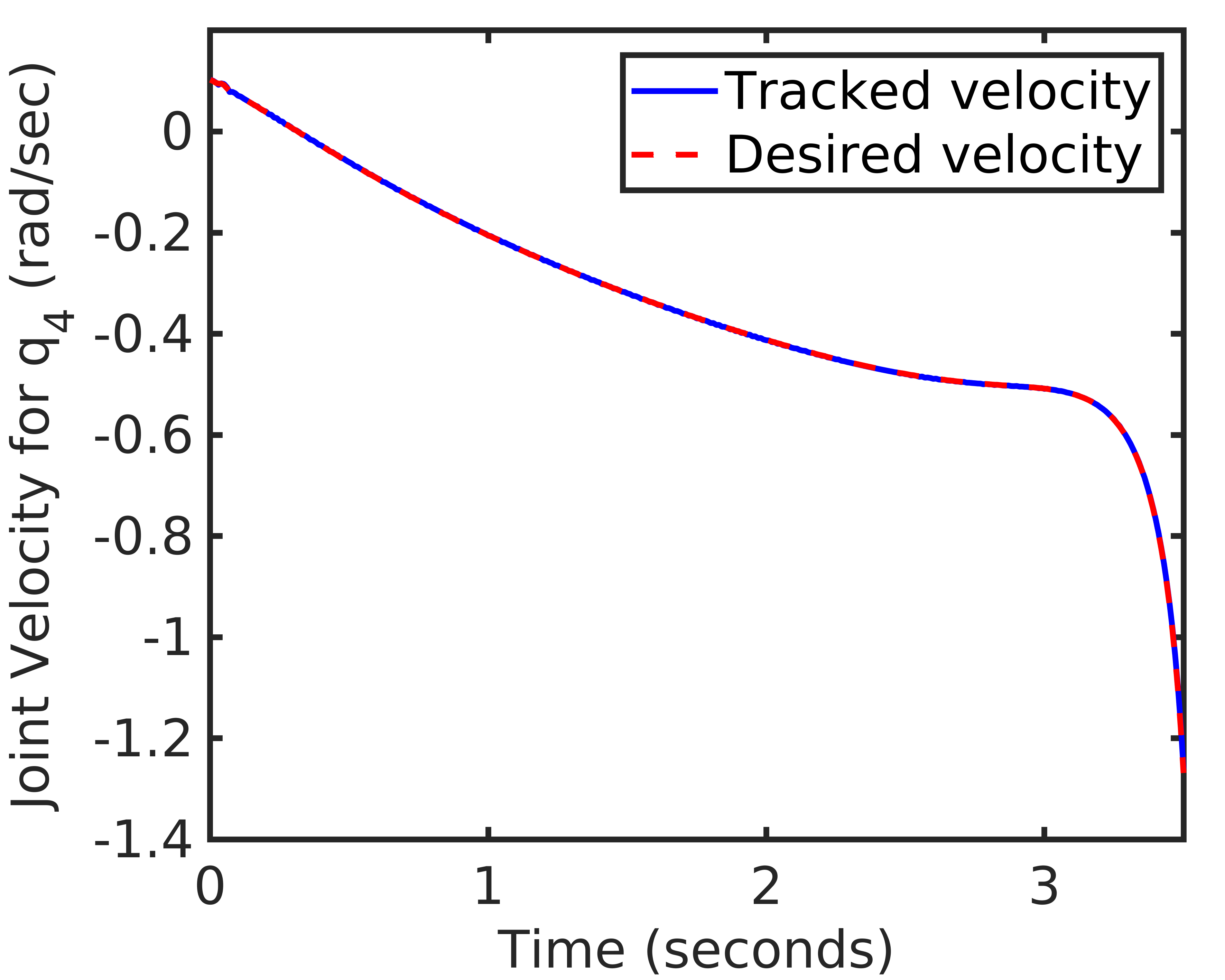}
  \caption{For Case 1}
\label{fig:case1s}
\end{subfigure}
\begin{subfigure}{.30\textwidth}
  \centering
  \includegraphics[width=\linewidth]{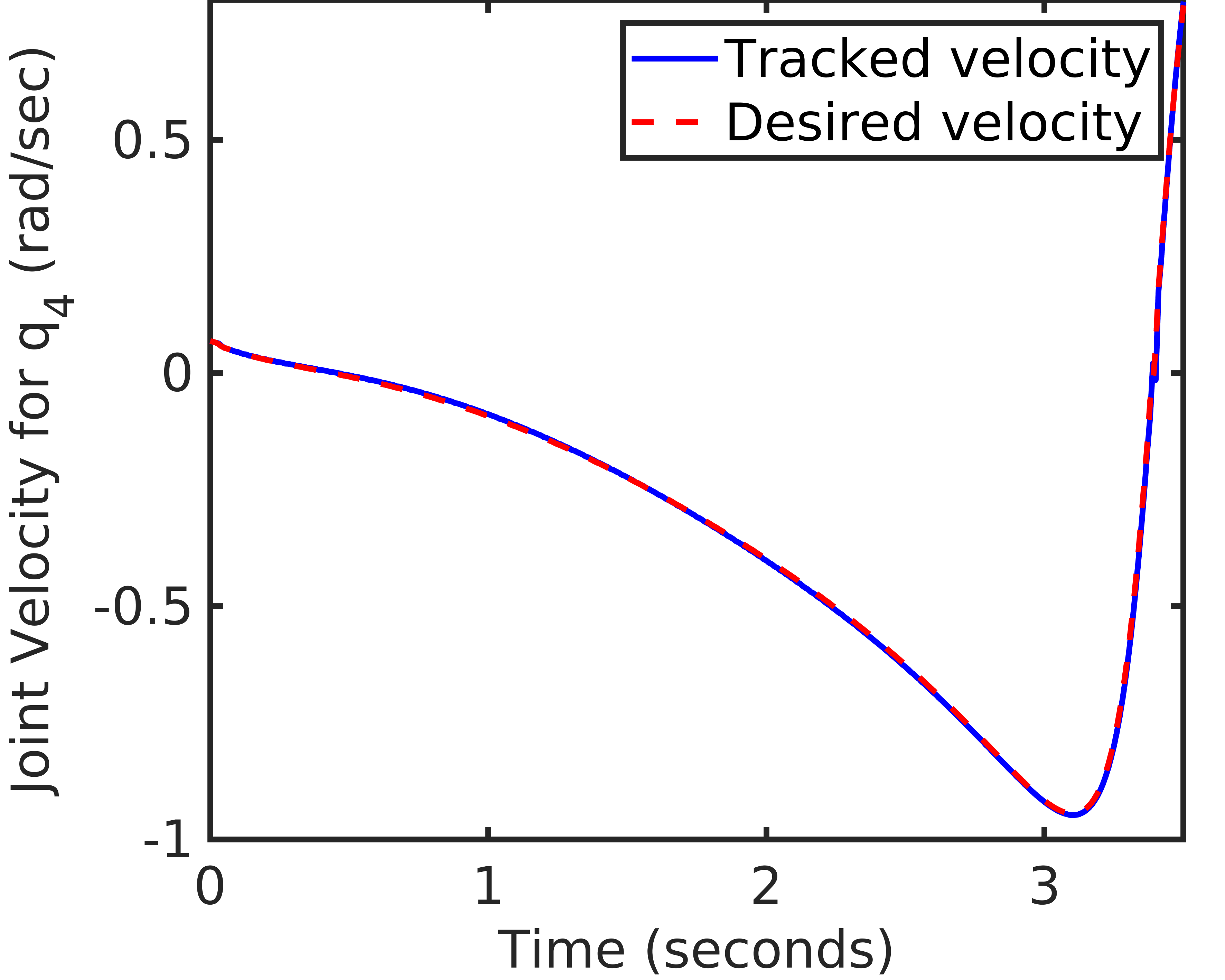}
  \caption{For Case 2}
\label{fig:case2s}
\end{subfigure}
\begin{subfigure}{.30\textwidth}
  \centering
  \includegraphics[width=\linewidth]{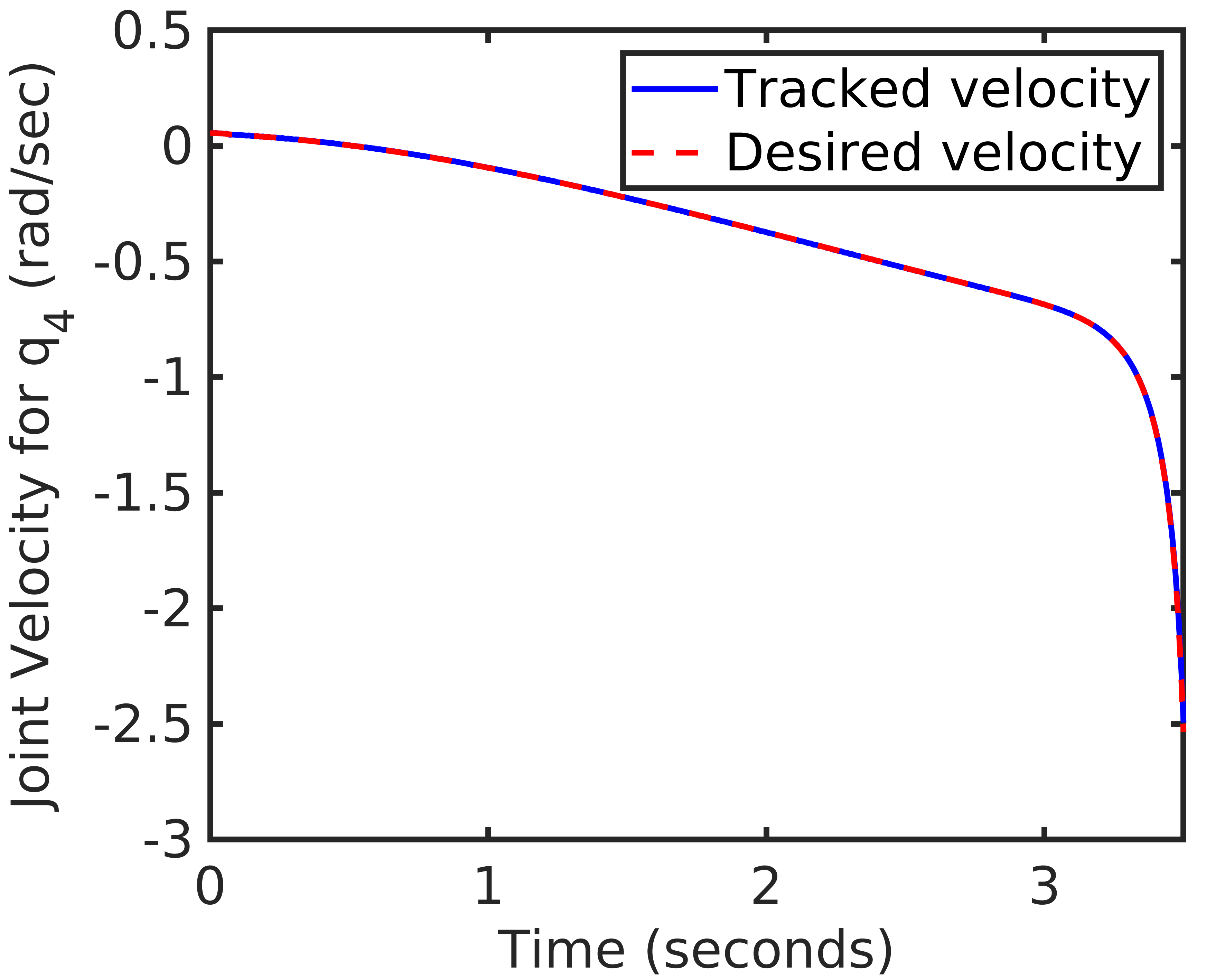}
  \caption{For case 3}
\label{fig:case3s}
\end{subfigure}
\caption{Tracking Performance $\dot{q}_4$}
\label{fig:dq4tp}
\vspace{-2mm}
\end{figure*}

\begin{figure*}[thpb]
\centering
\begin{subfigure}{.30\textwidth}
  \centering
  \includegraphics[width=\linewidth]{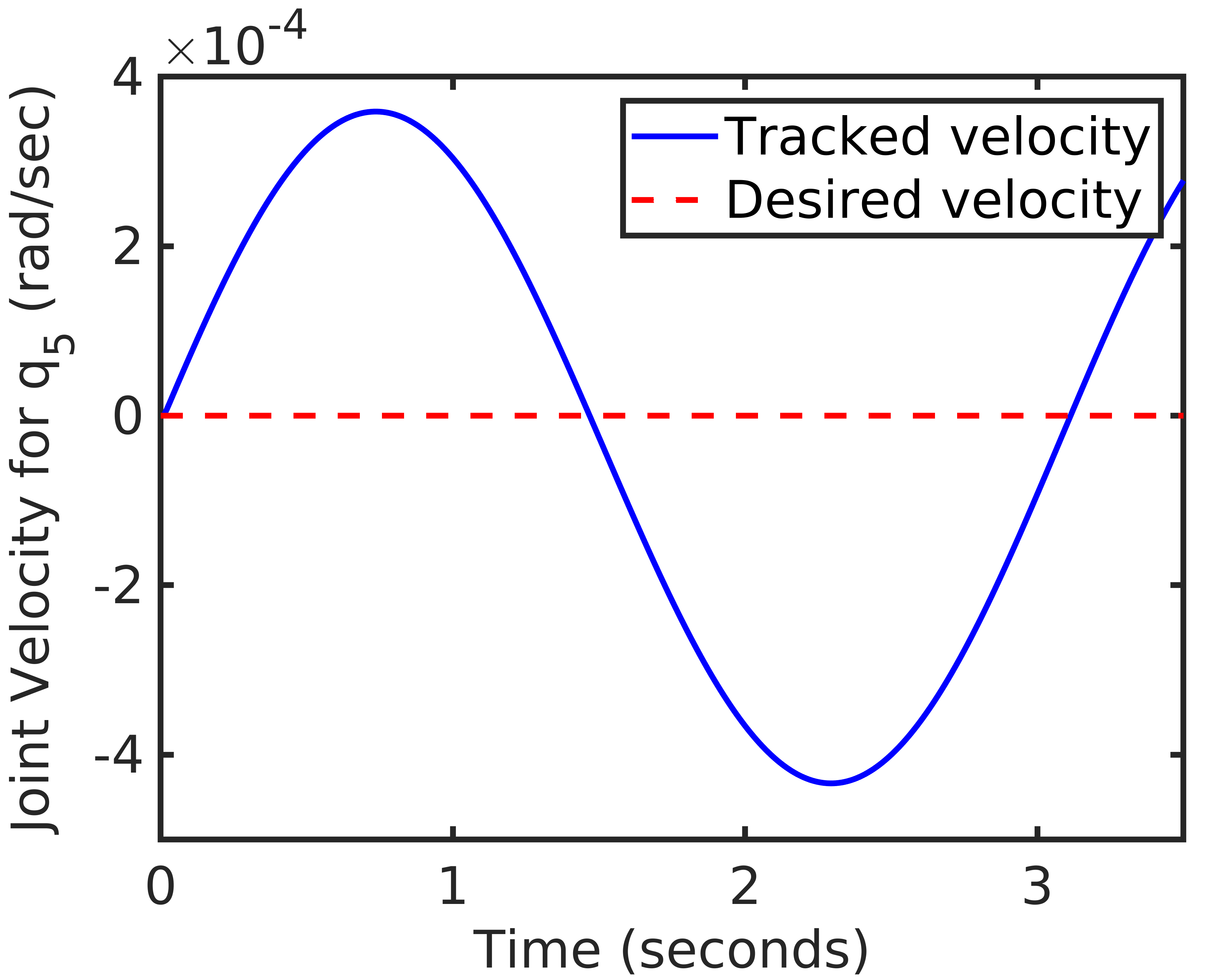}
  \caption{For Case 1}
\label{fig:case1s}
\end{subfigure}
\begin{subfigure}{.30\textwidth}
  \centering
  \includegraphics[width=\linewidth]{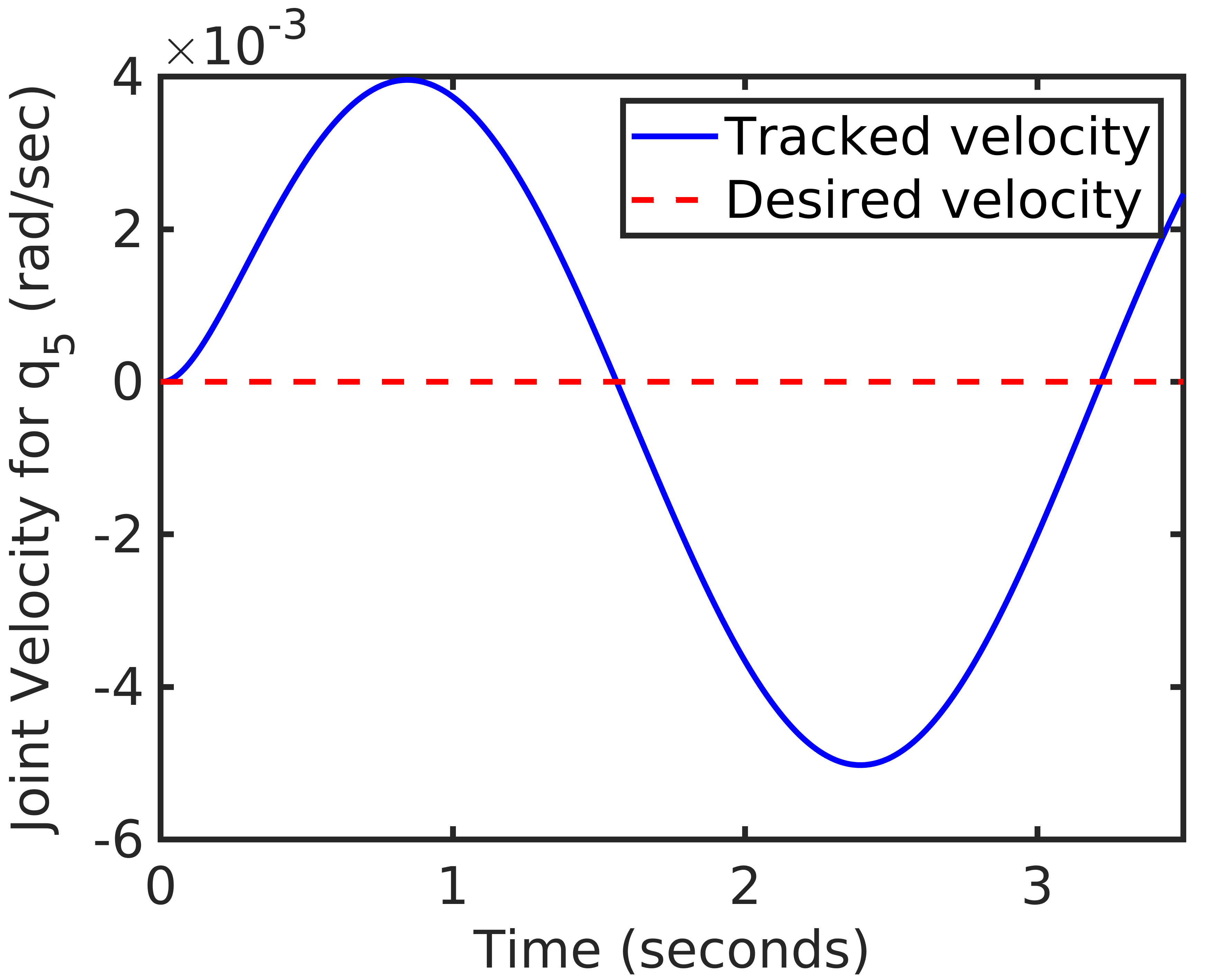}
  \caption{For Case 2}
\label{fig:case2s}
\end{subfigure}
\begin{subfigure}{.30\textwidth}
  \centering
  \includegraphics[width=\linewidth]{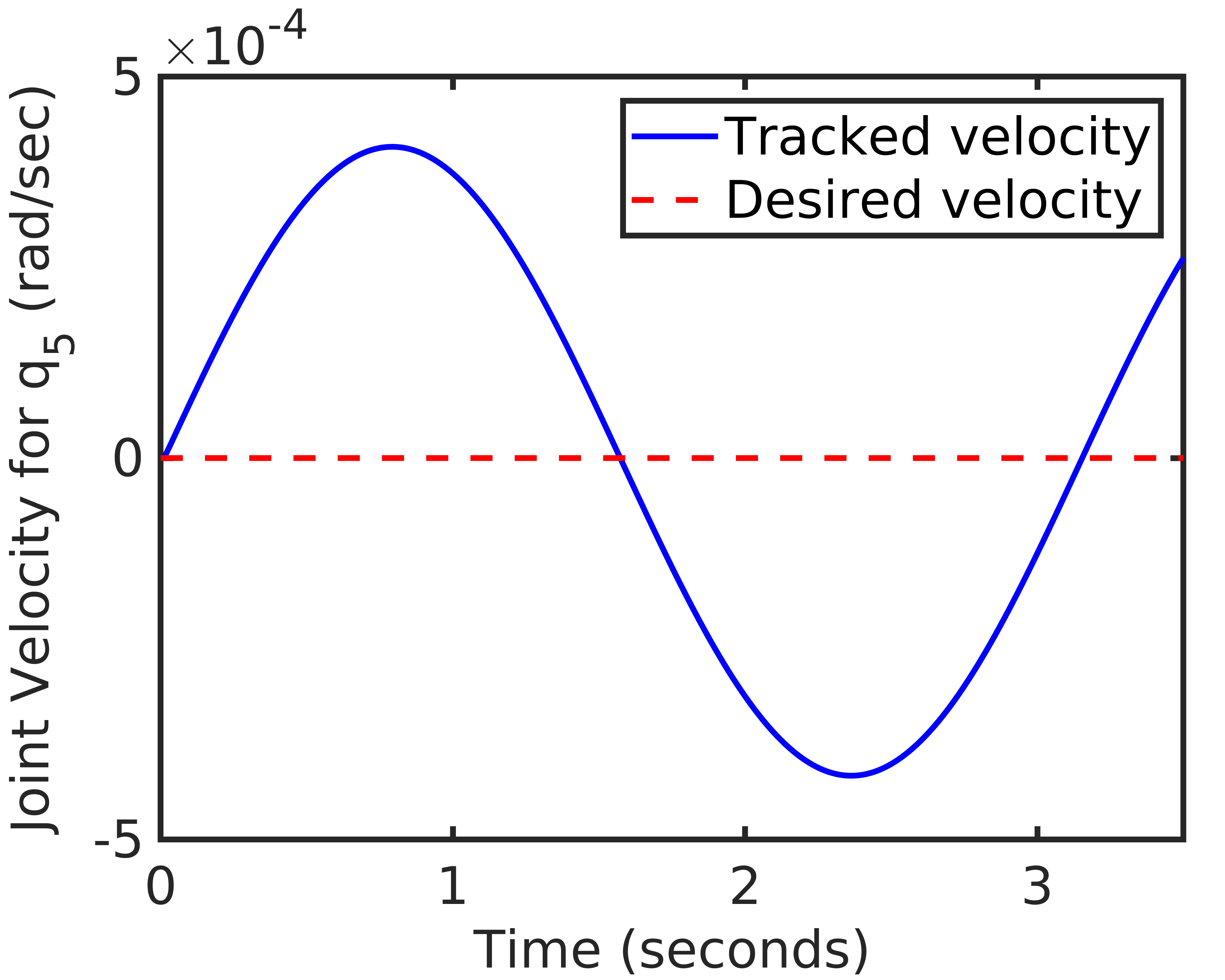}
  \caption{For case 3}
\label{fig:case3s}
\end{subfigure}
\caption{Tracking Performance $\dot{q}_5$}
\label{fig:dq5tp}
\vspace{-2mm}
\end{figure*}

\begin{figure*}[thpb]
\centering
\begin{subfigure}{.30\textwidth}
  \centering
  \includegraphics[width=\linewidth]{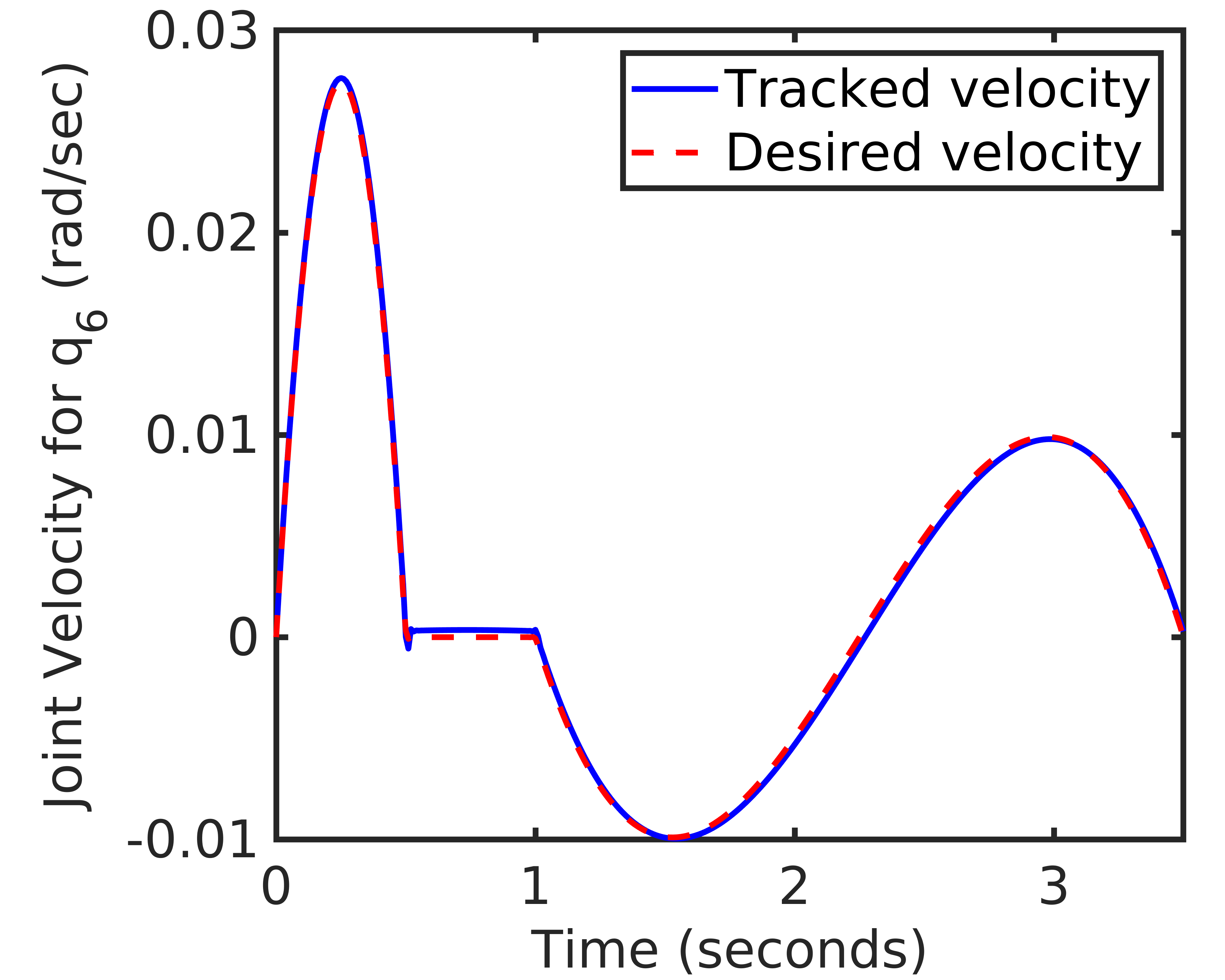}
  \caption{For Case 1}
\label{fig:case1s}
\end{subfigure}
\begin{subfigure}{.30\textwidth}
  \centering
  \includegraphics[width=\linewidth]{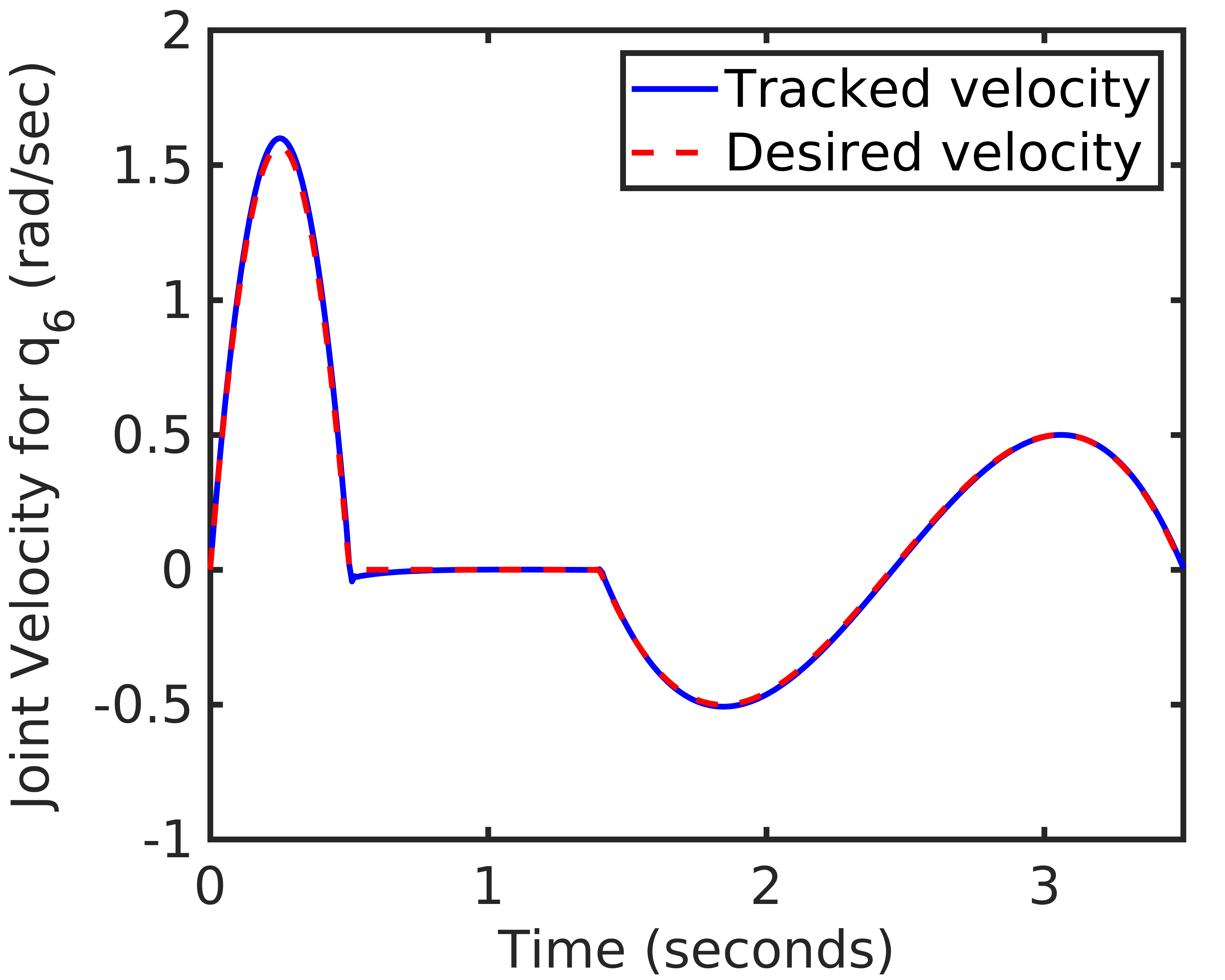}
  \caption{For Case 2}
\label{fig:case2s}
\end{subfigure}
\begin{subfigure}{.30\textwidth}
  \centering
  \includegraphics[width=\linewidth]{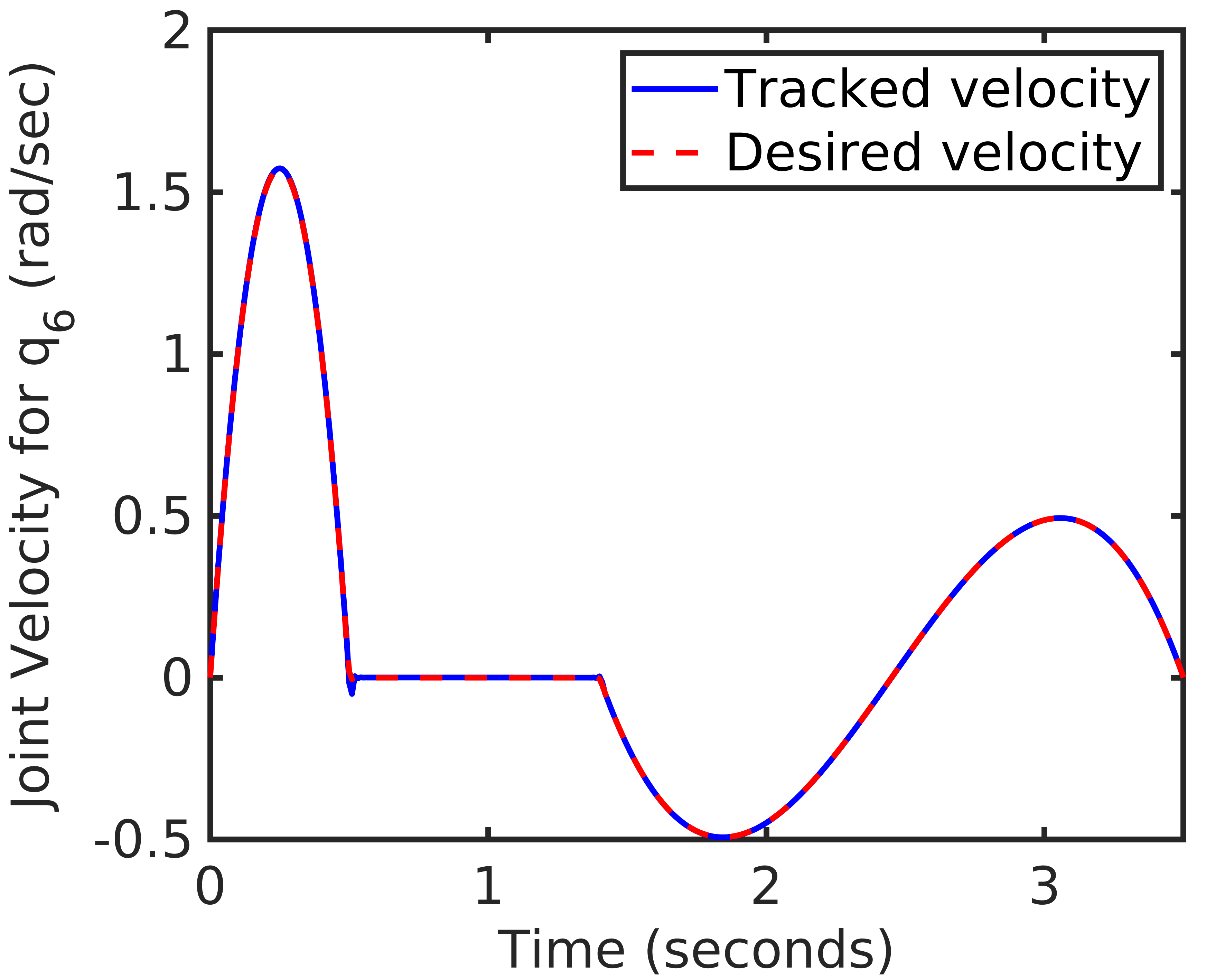}
  \caption{For case 3}
\label{fig:case3s}
\end{subfigure}
\caption{Tracking Performance $\dot{q}_6$}
\label{fig:dq6tp}
\vspace{-2mm}
\end{figure*}

\begin{figure*}[thpb]
\centering
\begin{subfigure}{.30\textwidth}
  \centering
  \includegraphics[width=\linewidth]{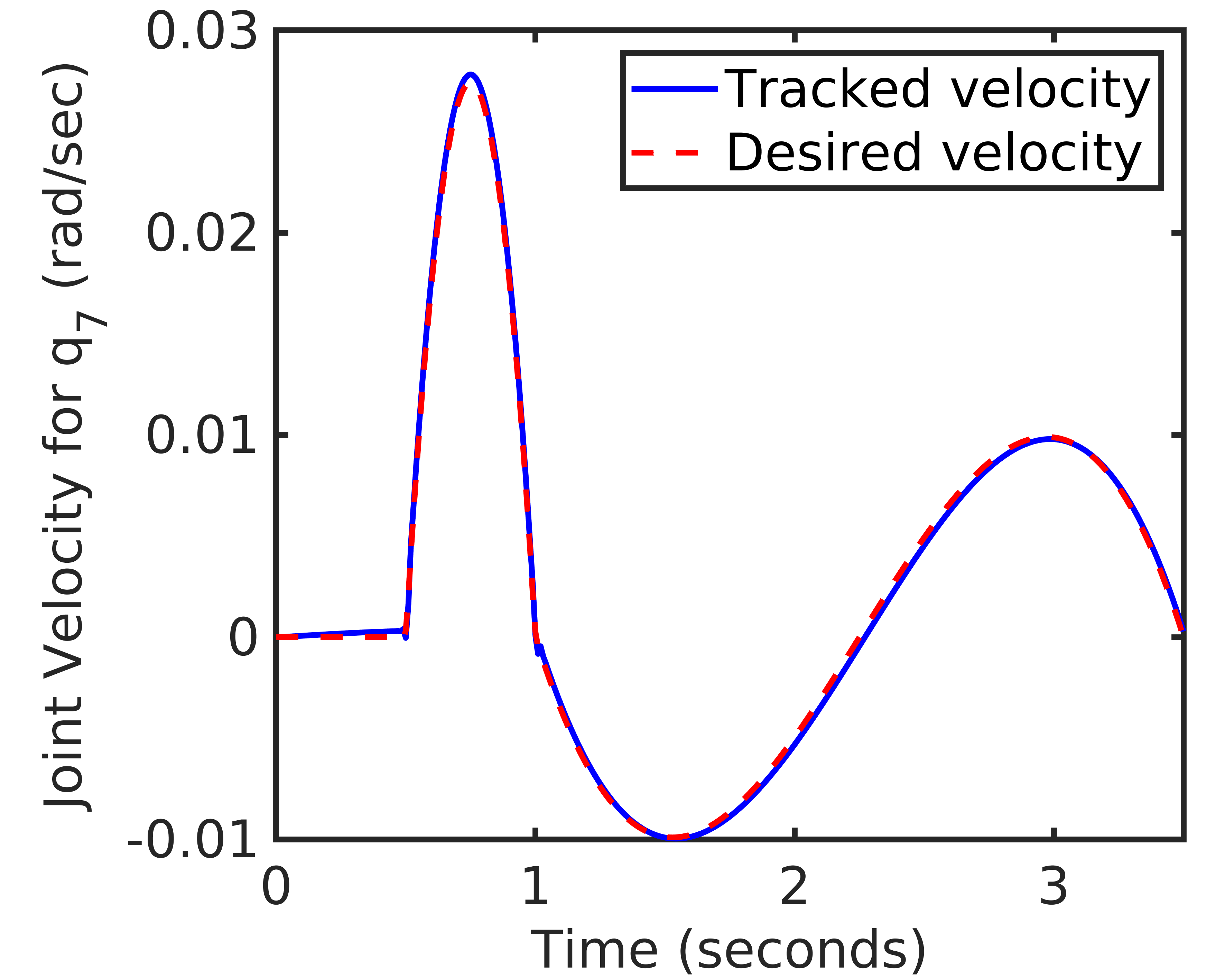}
  \caption{For Case 1}
\label{fig:case1s}
\end{subfigure}
\begin{subfigure}{.30\textwidth}
  \centering
  \includegraphics[width=\linewidth]{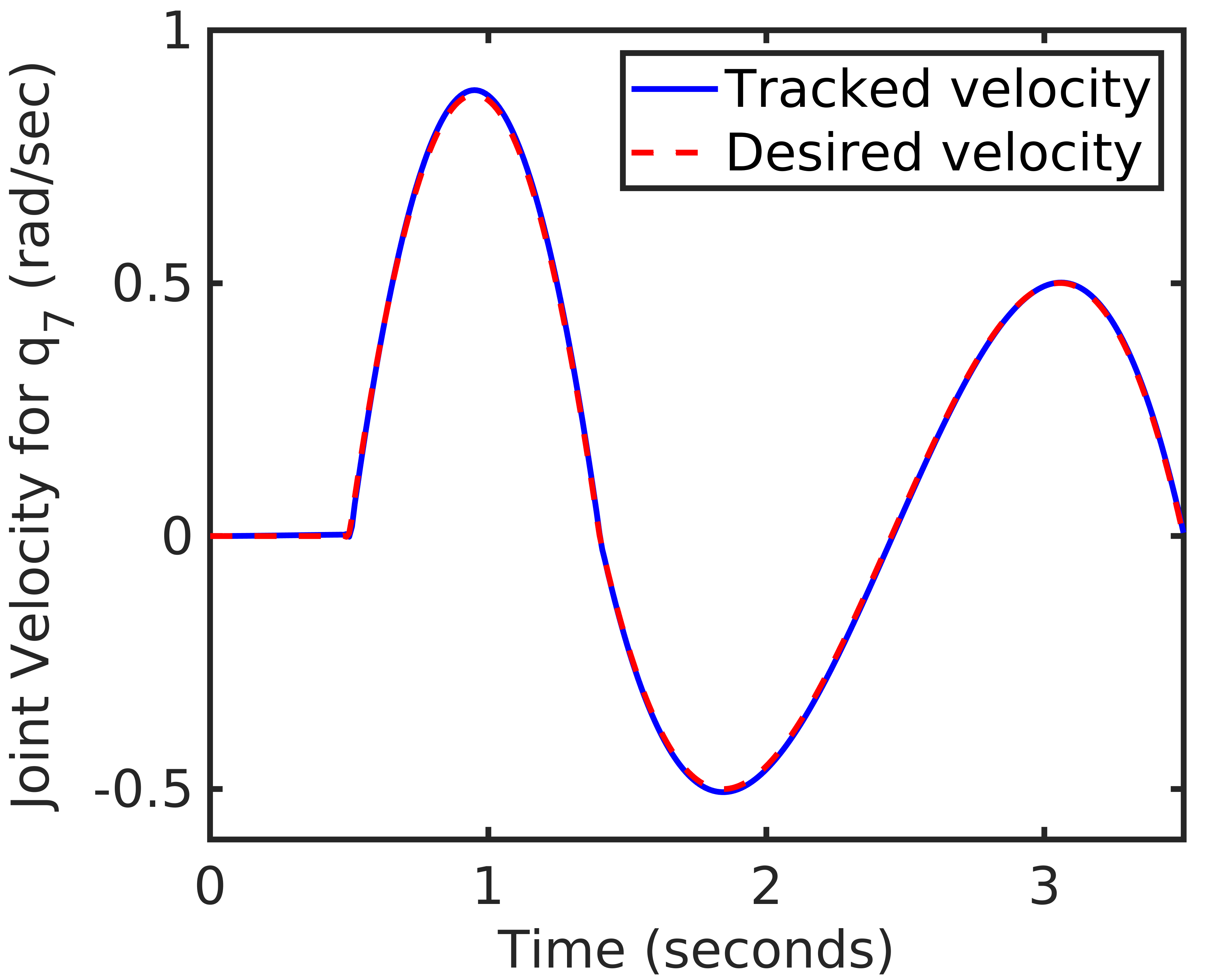}
  \caption{For Case 2}
\label{fig:case2s}
\end{subfigure}
\begin{subfigure}{.30\textwidth}
  \centering
  \includegraphics[width=\linewidth]{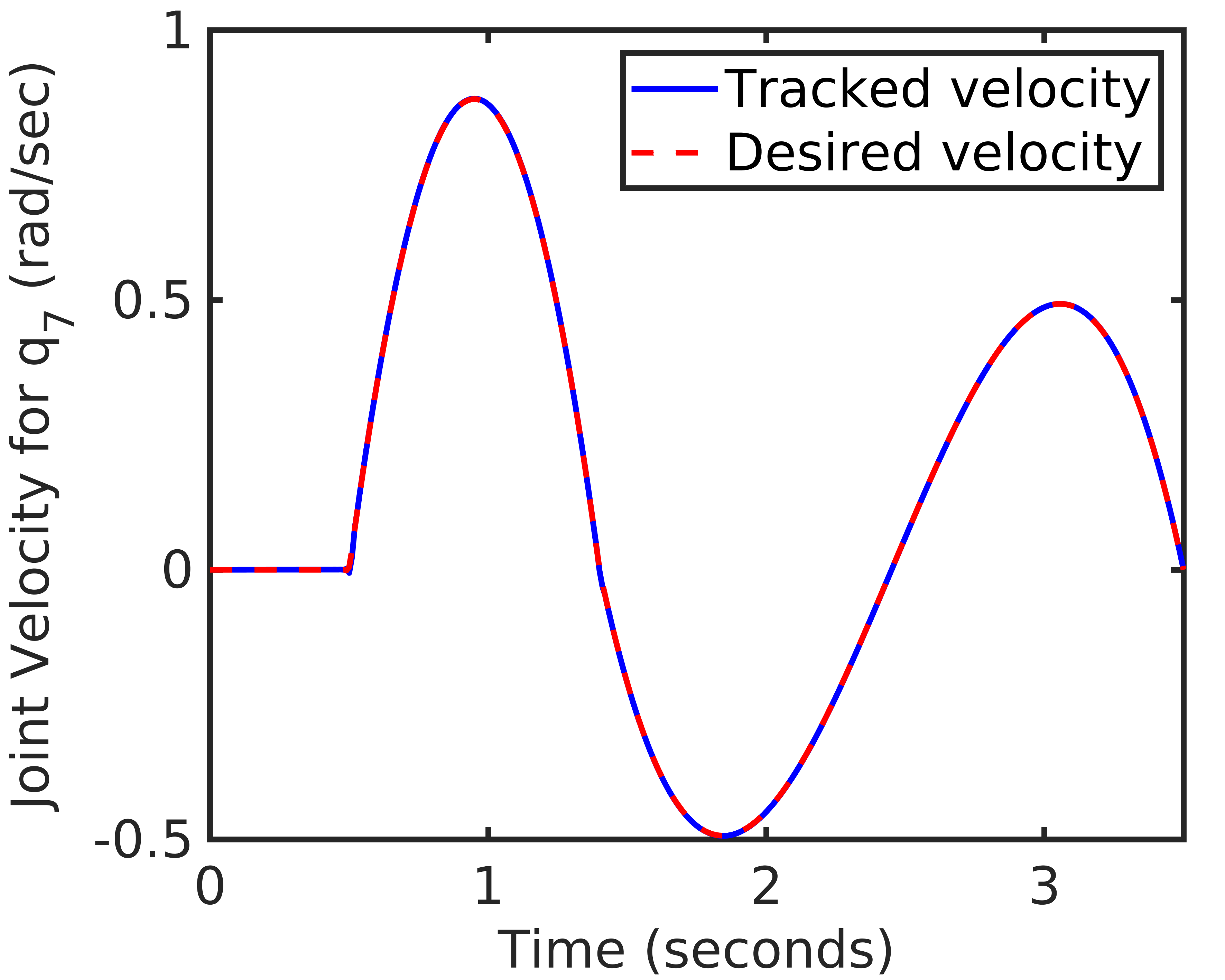}
  \caption{For case 3}
\label{fig:case3s}
\end{subfigure}
\caption{Tracking Performance $\dot{q}_7$}
\label{fig:dq7tp}
\vspace{-2mm}
\end{figure*}

\begin{figure*}[thpb]
\centering
\begin{subfigure}{.30\textwidth}
  \centering
  \includegraphics[width=\linewidth]{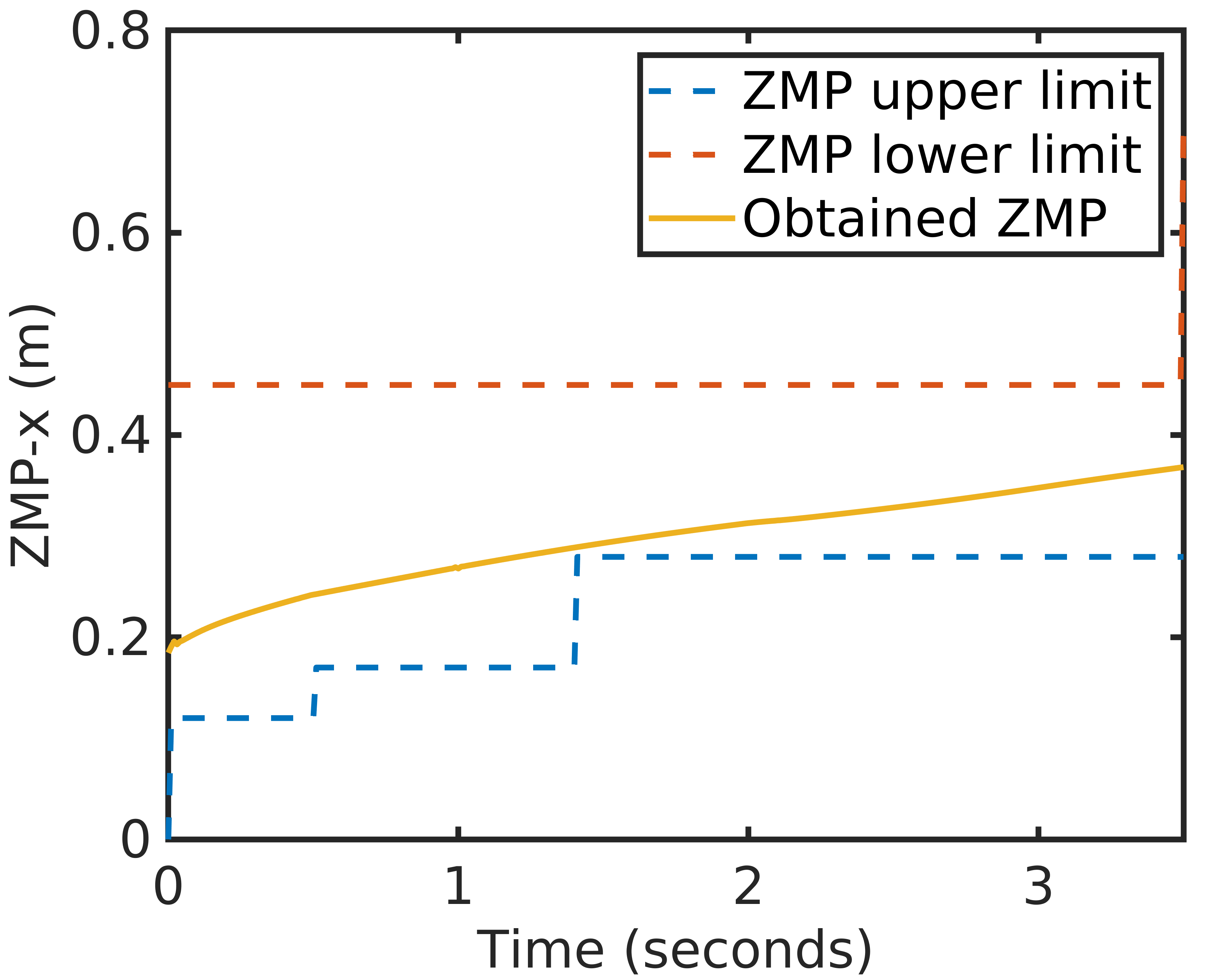}
  \caption{For Case 1}
\label{fig:case1s}
\end{subfigure}
\begin{subfigure}{.30\textwidth}
  \centering
  \includegraphics[width=\linewidth]{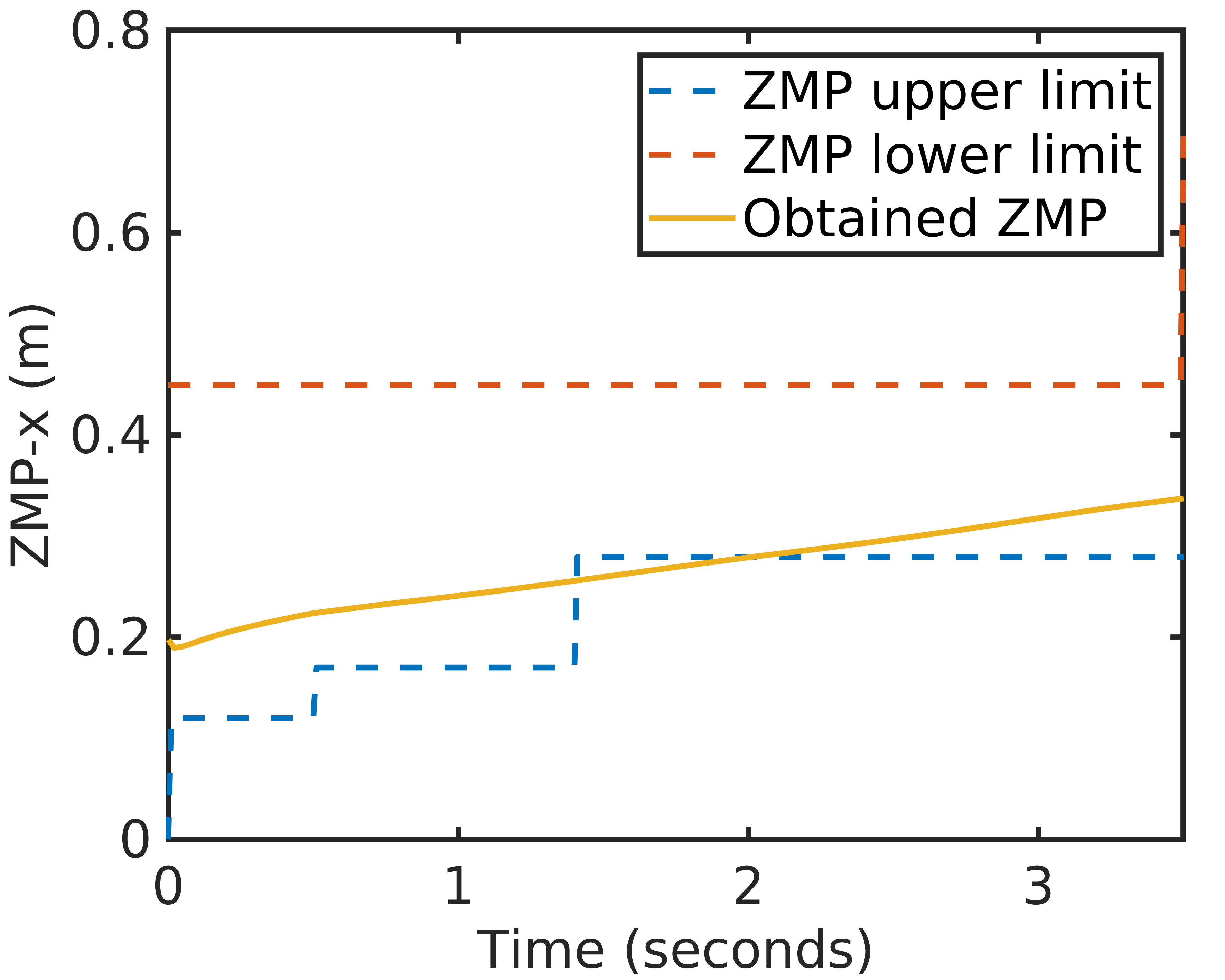}
  \caption{For Case 2}
\label{fig:case2s}
\end{subfigure}
\begin{subfigure}{.30\textwidth}
  \centering
  \includegraphics[width=\linewidth]{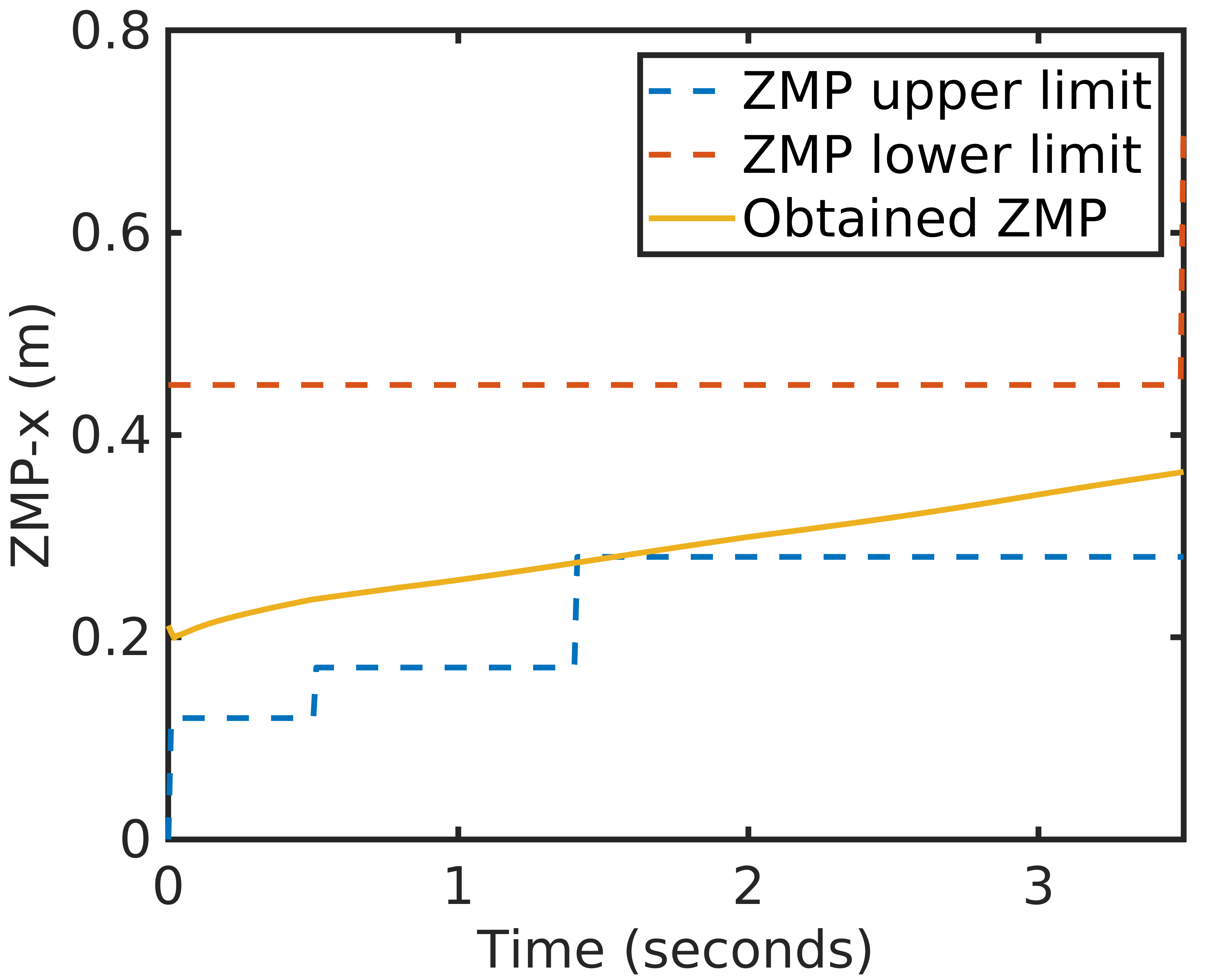}
  \caption{For case 3}
\label{fig:case3s}
\end{subfigure}
\caption{Obtained ZMP for all cases}
\label{fig:obzmp}
\vspace{-2mm}
\end{figure*}

\begin{figure*}[thpb]
\centering
\begin{subfigure}{.32\textwidth}
  \centering
  \includegraphics[width=\linewidth]{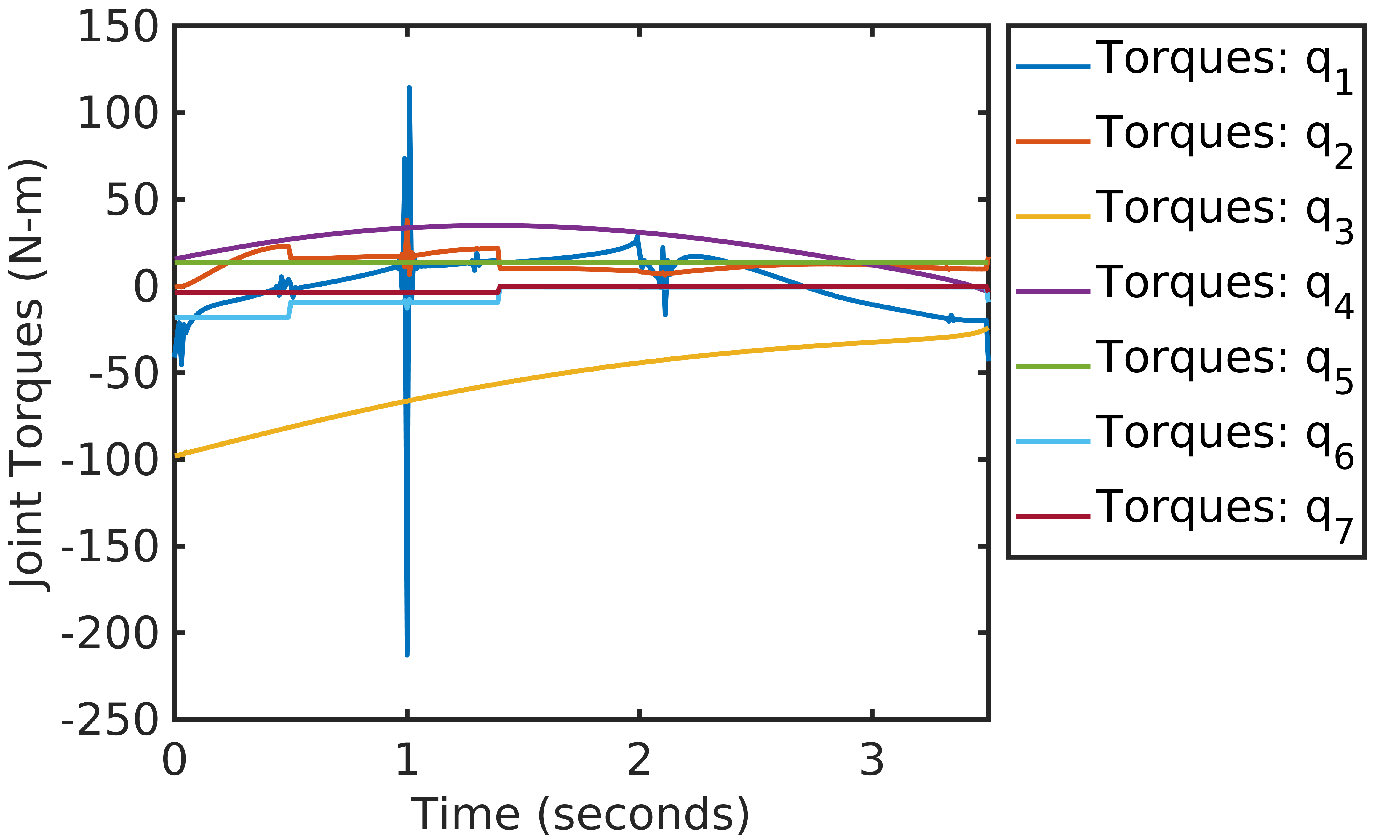}
  \caption{For Case 1}
\label{fig:case1s}
\end{subfigure}
\begin{subfigure}{.32\textwidth}
  \centering
  \includegraphics[width=\linewidth]{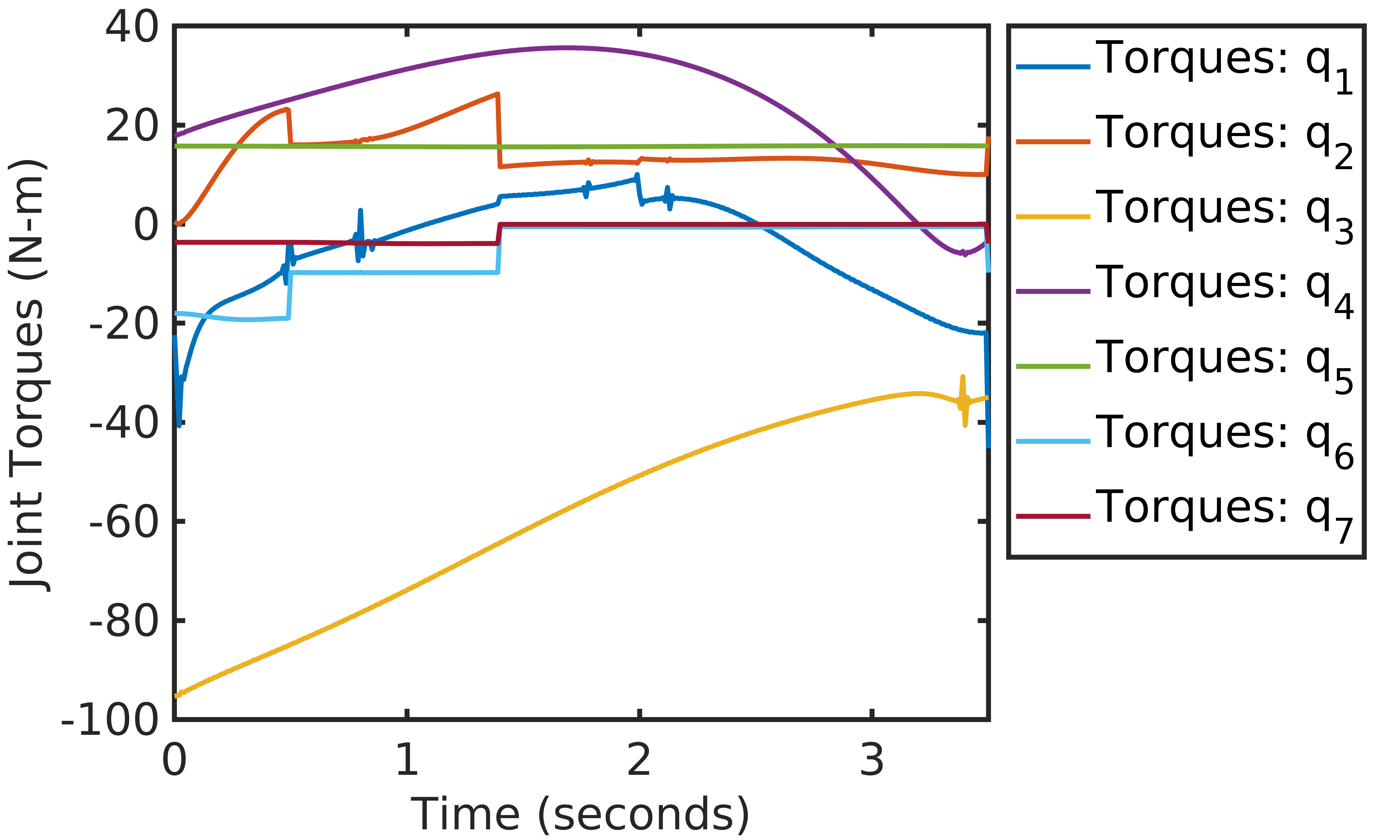}
  \caption{For Case 2}
\label{fig:case2s}
\end{subfigure}
\begin{subfigure}{.32\textwidth}
  \centering
  \includegraphics[width=\linewidth]{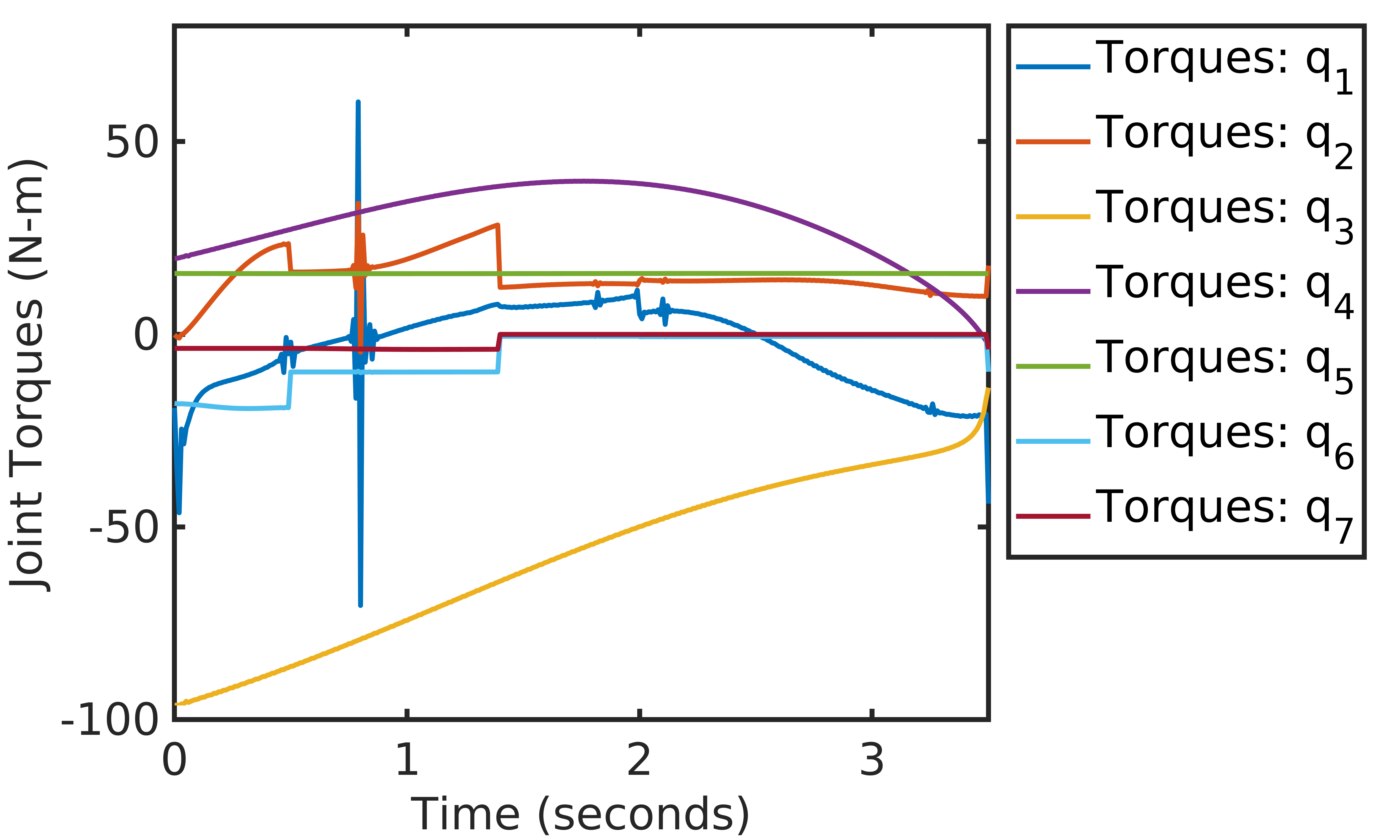}
  \caption{For case 3}
\label{fig:case3s}
\end{subfigure}
\caption{Joint Torques Obtained}
\label{fig:torques}
\vspace{-2mm}
\end{figure*}

Now, coming towards the joint space, we want joint angle trajectories to be as smooth as possible to avoid jerk during the joint motion. We have implemented unsupervised artificial neural network (UIKNN) to obtain inverse kinematics solution to get joint angle trajectories. Figures \ref{fig:q1tp} - \ref{fig:q7tp} and \ref{fig:dq1tp} - \ref{fig:dq7tp} shows joint space trajectory tracking performance for joint position and joint velocity vectors respectively in all the three cases. Figure \ref{fig:torques} shows obtained actual torque applied for all cases. Finally figure \ref{fig:obzmp} is concerned with stability which shows obtained ZMP trajectory for all three cases.Based on total energy consumption and minimum torque also, case 2 which in our case is 0.700 Rise/Run ratio was found to be most optimal with minimum cost based on Ant Colony Optimization. 

\clearpage
\section{CONCLUSIONS} \label{cncl}

The study presents a novel methodology for a cycloidal realization on upstairs climbing. Cycloidal trajectory has benefits of smooth velocity and acceleration profiles which is desired for serial manipulator movements.  We track this trajectory by a fast non-linear approximation of the inverse kinematics module using an unsupervised feed forward network. Coupled with the trajectory planning is the dynamic trajectory tracking which involves the novelty of using Time Quantized Lagrange Dynamics, an approach which tends to simplify the trajectory control torque calculations with small simulation step sizes. We have considered Ant colony Optimization to tune the PD Controller gains and torso pitch angle with combined objective to track joint space trajectories as well as instantaneous power resulting in a more energy efficient gait. Following that, We validate the adaptive capacity of our complete framework with 3 cases of different stair geometries with varying rise-run ratio.  The torques are calculated to give minimum cost function and produce joint accelerations capable enough to maintain ZMP-stability. The study concludes with a proper joint space trajectories analysis, ZMP-Stability analysis, calculation of instantaneous power produced by the actuators as well as overall energy consumption analysis.  Future investigations includes completion of various task objectives along with trajectory tracking along with the extension of our model movement into the frontal plane as well.


%
\section*{Conflict of interest}
The authors declare that they have no conflict of interest.



\end{document}